\documentclass[11pt,a4paper]{article}

\usepackage{polyu-vclab}

\usepackage{lipsum}                % placeholder text, remove for real paper
\usepackage{amsmath}
\usepackage[numbers,sort&compress]{natbib}
\usepackage[utf8]{inputenc} % allow utf-8 input
\usepackage[T1]{fontenc}    % use 8-bit T1 fonts
\usepackage{hyperref}       % hyperlinks
\usepackage{url}            % simple URL typesetting
\usepackage{booktabs}       % professional-quality tables
\usepackage{amsfonts}       % blackboard math symbols
\usepackage{nicefrac}       % compact symbols for 1/2, etc.
\usepackage{microtype}      % microtypography
\usepackage{multirow}
\usepackage{graphicx}
\usepackage[dvipsnames]{xcolor}
\usepackage{adjustbox}

\setEyebrow{PolyU VCLab\,\textbullet\,Preprint 2026}

\setReportTag{Visual Computing Lab\,\textperiodcentered\,The Hong Kong Polytechnic University}

\papertitle{GGT-100K: Generative Ground Truth for Generalizable Real-World Image Restoration}

\paperauthors{%
  Xiangtao Kong\equalmark\affilmark{1,2}\quad
  Jixin Zhao\equalmark\affilmark{1,2}\quad
  Lingchen Sun\affilmark{1,2}\quad
  Rongyuan Wu\affilmark{1,2}\quad
  Lei Zhang\correspondmark\affilmark{1,2}%
}

\paperaffil{%
  \affilmark{1}\,The Hong Kong Polytechnic University\quad
  \affilmark{2}\,OPPO Research Institute
}

\papernotes{%
  \equalmark\,Equal contribution.\quad
  \correspondmark\,Corresponding author (\href{mailto:cslzhang@comp.polyu.edu.hk}{cslzhang@comp.polyu.edu.hk}).%
}

\paperbadges{%
  \vclabbadge{Project Page}{https://polyu-vclab.github.io/GGT-100K/}\;%
  \vclabbadge{Code}{https://github.com/PolyU-VCLab/GGT-100K/}\;%
  \vclabbadge{Dataset}{https://huggingface.co/datasets/VCLab-PolyU/GGT-100K/tree/main}\;%
}

\begin{document}
%==========================================================================

\maketitleVCLab

%-----------------------------------------------------------
% Abstract -- bordered box with red left bar (VCLab mission style)
%-----------------------------------------------------------

\begin{figure*}[h]
    \centering
    \includegraphics[width=\linewidth]{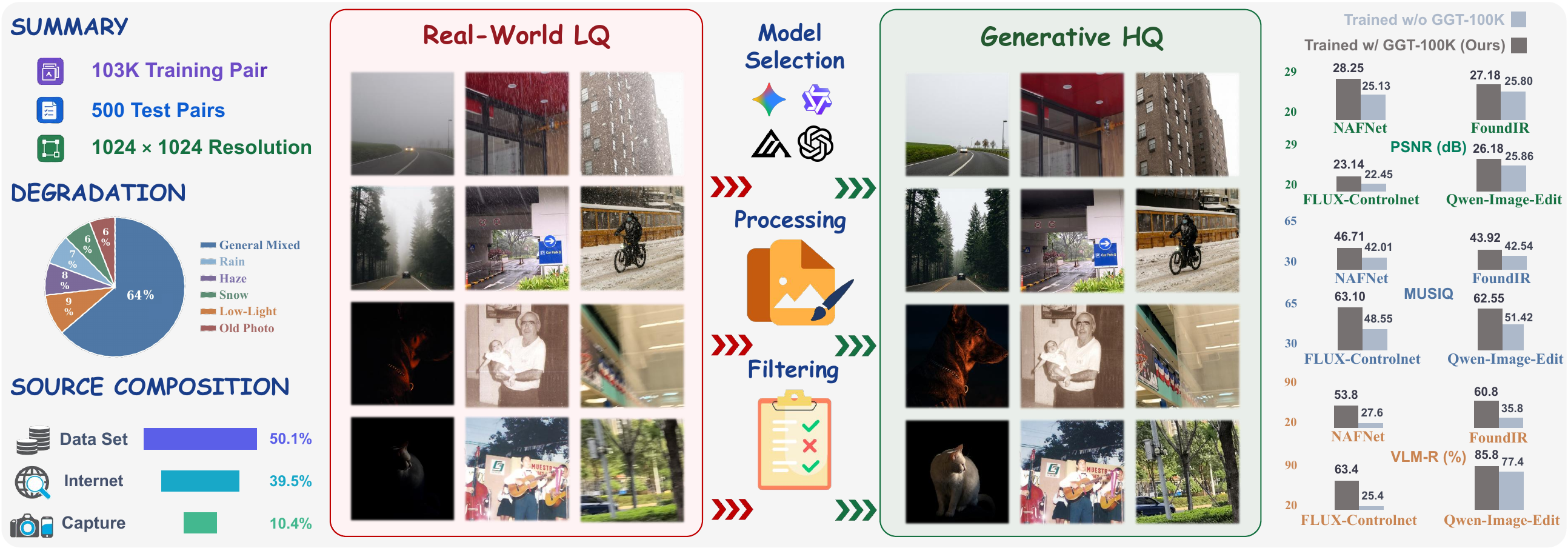}
    \caption{We construct GGT-100K through a carefully designed pipeline, aiming to improve the generalization performance of a wide range of image restoration models. }
    \label{fig:teaser}
\end{figure*}

\begin{vclabAbstract}
Real-world image restoration (IR) is bottlenecked by the scarcity of high-quality paired training data. Synthetic datasets are abundant but often fail to model real-world degradations, while real-world paired datasets are expensive and difficult to capture. As a result, IR models trained on these datasets show limited generalization in real-world scenarios. In this work, we propose Generative Ground Truth (GGT) by using generative multimodal foundation models (MFMs) to produce high-quality (HQ) targets from real-world low-quality (LQ) images. We first conduct a systematic evaluation of nine state-of-the-art MFMs, including Nano-Banana-2 and GPT-Image-2, on images of various scenes and degradation types. The results demonstrate that Nano-Banana-2 with VLM-based adaptive prompting shows the highest capability to synthesize perceptually realistic and content-faithful HQ targets, which can serve as the GGT for the LQ input. We then employ Nano-Banana-2 to build a GGT synthesis pipeline, which involves multi-stage quality control to ensure data reliability, and construct GGT-100K, an LQ-HQ paired dataset comprising 103,707 training pairs and covering diverse scenes and complex real-world degradations. A test set of 500 image pairs is also established. Extensive experiments show that GGT-100K consistently improves the real-world generalization of a wide range of IR models, with particularly strong benefits for finetuning generative models for IR tasks. Our results suggest that MFMs can serve as practical tools for restoration-oriented data generation, and GGT-100K is a useful resource to expand the generalization boundaries of real-world IR models. %Our dataset will be publicly available. 
\end{vclabAbstract}

\keywords{Generalizable image restoration; Generative ground truth; Multimodal foundation models}

\begin{figure*}[t]
    \centering
    \includegraphics[width=\linewidth]{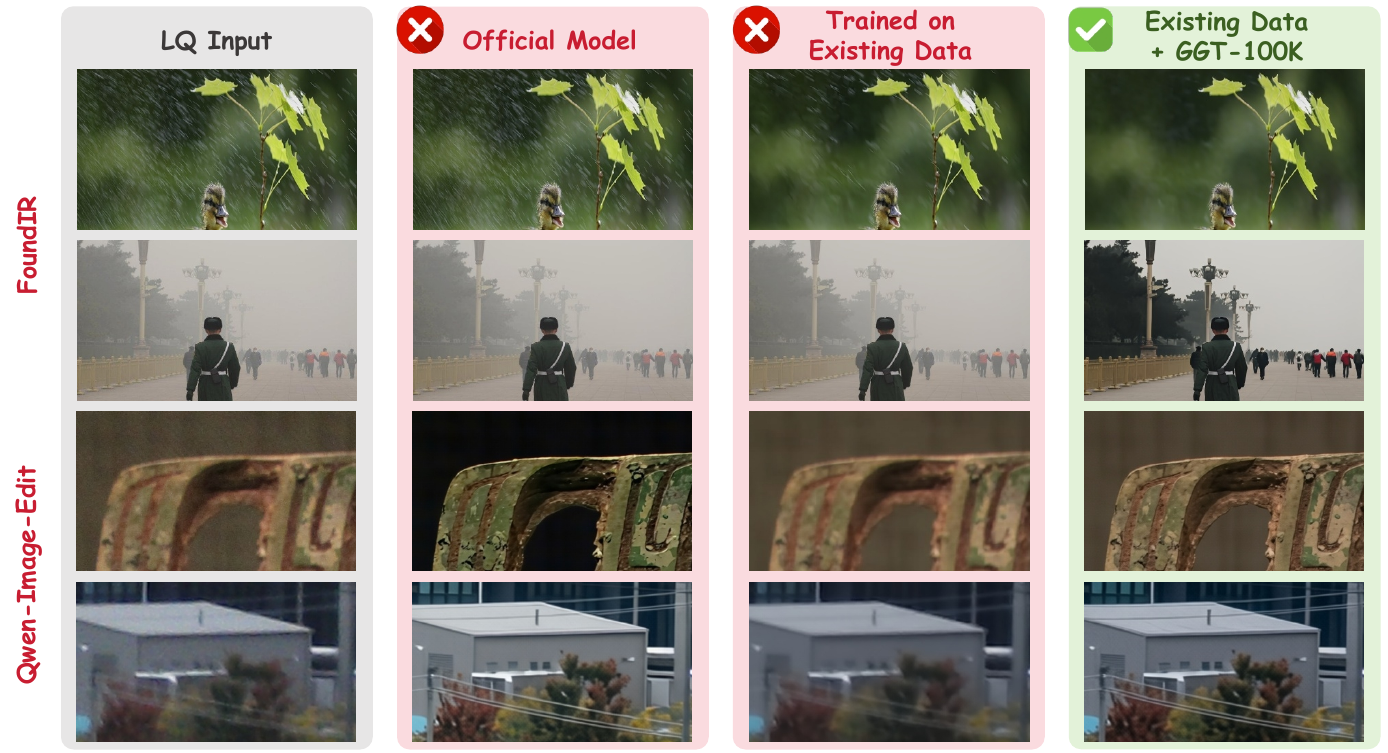}
    \vspace{-6mm}
    \caption{Qualitative comparison of FoundIR~\cite{li2025foundir} and Qwen-Image-Edit~\cite{Qwen-Image-Edit} trained/finetuned on different datasets. FoundIR shows limited generalization to real rain and haze, even after finetuned on existing restoration datasets. Qwen-Image-Edit produces sharper details with its strong generative prior but may introduce hallucinations or color shifts, while finetuning on existing datasets weakens its detail generation ability. GGT-100K significantly improves the generalization capability of them to real-world degradations, enhancing visual details while preserving scene fidelity.}
    \label{fig:intro}
    \vspace{-4mm}
\end{figure*}

\section{Introduction}

Classical image restoration (IR)~\cite{zhang2017beyond,dong2014learning,RESIDE,cho2021_deblur} typically focuses on a specific predefined degradation, such as denoising, deblurring, super-resolution or dehazing. In contrast, real-world IR~\cite{realsr,li2025foundir} aims to recover images affected by complex, mixed, and often unknown degradations in practical environments. Despite substantial progress in IR network architectures--from CNN and Transformer backbones~\cite{zhang2017beyond,liang2021swinir,zamir2021multi,chen2022simple,X-Restormer} to all-in-one models~\cite{AirNet,PromptIR,MoCE-IR,DA-CLIP,li2025foundir} and to recent generative frameworks~\cite{lin2023diffbir,yu2024scaling,wu2024seesr}--robust generalization to real-world scenarios remains far from solved.

The lack of paired training data is one of the key bottlenecks for generalizable real-world IR models. Existing paired data construction mainly follows two routes: synthetic generation and real-world acquisition. Synthetic data are scalable, but simulated degradations often fail to model the complexity of real-world image formation process, leading to a substantial domain gap~\cite{zhang2021designing,wang2021real,liu2022evaluating,liu2023evaluating,kong2024preliminary}. Physically collected real-world image pairs provide more realistic supervision, but they are expensive, difficult to scale, and often limited in scene diversity because high-quality (HQ) and well-aligned references are hard to obtain under transient conditions such as weather, motion, and illumination changes~\cite{realsr,drealsr,plotz2017benchmarking,RESIDE,li2025foundir}. As shown in Fig.~\ref{fig:intro}, the FoundIR model \cite{li2025foundir} trained on synthetic data or its officially released real-world dataset still produces noticeable artifacts for low-quality (LQ) input images. This persistent data bottleneck motivates us to develop a scalable paradigm to construct HQ restoration targets from diverse real-world LQ inputs.

Recent generative multimodal foundation models (MFMs)  ~\cite{team2023gemini,openai2025gptimage15,Qwen-Image-Edit} offer a promising opportunity to achieve our goal. Modern MFMs can take image and instructions as inputs to produce the desired output, suggesting that they may generate restoration-oriented HQ targets for real-world LQ images. However, this task is nontrivial: current MFMs may distort image structures, hallucinate details, or behave inconsistently across images and prompts. This raises a key question: \textit{can MFMs generate HQ targets with sufficient fidelity and stability for supervised real-world IR model training}?

In this work, we conduct a systematic investigation to answer this question. We first evaluate nine modern MFMs, including Nano-Banana-2~\cite{team2023gemini} and GPT-Image-2~\cite{openai2025gptimage2}, by prompting them to generate HQ counterparts of the input LQ images with various image scenes and degradation types. Fixed and VLM-based adaptive prompting strategies are employed. We evaluate these MFMs in terms of image content fidelity, perceptual quality, VLM-based evaluation, and human preference, and find that Nano-Banana-2 with adaptive prompting can generate the most reliable HQ target image, which can serve as the Generative Ground Truth (GGT) to supervise the real-world IR model training.

Based on the above finding, we employ Nano-Banana-2 to construct GGT-100K, an LQ-HQ paired real-world IR dataset with 103,707 training pairs. A test set of 500 image pairs is also carefully established. In particular, as shown in Fig.~\ref{fig:pipeline}, we collect real-world LQ images from existing datasets, Internet sources, and our own captures, covering diverse scenes and degradation types, including general mixed degradations, rain, haze, snow, low-light conditions, and old photos. It is worth mentioning that these categories should not be viewed as isolated single-degradation settings. For example, images captured under rain are not only rain-corrupted but also coupled with commonly encountered degradations in real photography, such as blur, noise, and compression artifacts. We then generate restoration targets using Nano-Banana-2 with a multi-stage quality control process, including automatic metric-based filtering, VLM-assisted screening, and manual verification. To validate the effectiveness of GGT-100K, we retrain representative CNN-based models~\cite{zamir2021multi,chen2022simple}, transformer-based models~\cite{liang2021swinir,X-Restormer}, all-in-one restoration models~\cite{PromptIR,MoCE-IR,DA-CLIP,li2025foundir}, T2I-~\cite{FLUX} and TI2I-~\cite{Qwen-Image-Edit} adapted generative restoration models with and without GGT-100K, and observe consistent gains in real-world generalization across multiple evaluation datasets and model families.

Our contributions are threefold. First, we propose GGT, a scalable paradigm for constructing real-world paired IR training data with MFMs, and systematically evaluate nine MFMs with multiple prompting strategies, providing practical insights for restoration-oriented GGT generation under diverse real-world degradations. Second, we build GGT-100K, a comprehensive paired dataset with 103K high-quality LQ-HQ pairs for real-world IR model training. Finally, we demonstrate that GGT-100K can consistently improve the generalization performance of different restoration model families, especially the modern generative models with strong priors and learning capacity. An overview of our GGT-100K dataset is illustrated in Fig.~\ref{fig:teaser}.

\section{Related Work}

\vspace{2mm}
\noindent\textbf{Real-world Image Restoration.}
Real-world IR aims to recover LQ images captured in complex real-world environments into expected HQ images. To achieve this goal, restoration methods have evolved from CNN-based ~\cite{zamir2021multi,chen2022simple} to transformer-based architectures~\cite{liang2021swinir,zamir2022restormer}, all-in-one frameworks~\cite{AirNet,PromptIR}, and more recent generative approaches~\cite{ho2020denoising,lin2023diffbir,wu2024seesr}. Lightweight CNN backbones remain attractive for their efficiency, while heavier transformer-based models often achieve stronger restoration performance~\cite{zamir2022restormer,liang2021swinir}. All-in-one IR methods improve applicability by handling multiple degradation types within a unified framework~\cite{AirNet,PromptIR}. More recently, generative methods~\cite{lin2023diffbir,wu2024seesr,wei2025perceive} have shown stronger restoration capability and improved generalization, but they may sacrifice fidelity and introduce hallucinated details. Overall, despite rapid progress in model design, robust generalization across diverse real-world degradations remains difficult, mainly due to the data bottleneck.

Most existing real-world IR methods rely on synthetic training data, where LQ inputs are generated from HQ images using hand-crafted degradation models~\cite{zhang2021designing,wang2021real}. Such synthetic data are easy to scale but hard to capture the complexity of real-world image degradation, leading to a substantial domain gap~\cite{liu2022evaluating,liu2023evaluating}. A few real-world paired datasets have been built through settings such as multiple acquisition, controlled imaging, etc., ~\cite{realsr,drealsr,plotz2017benchmarking,RESIDE}. These datasets provide more realistic supervision, but they are expensive and difficult to scale. Recent efforts such as FoundIR~\cite{li2025foundir} have advanced real-world data collection in this direction, but relying on real capture alone encounters practical difficulties since scenarios such as rain and haze still need synthetic data and the captured data are constrained in scene and degradation diversity. As a result, existing real-world paired datasets remain limited in scalability, scene diversity, and degradation coverage. %This limitation motivates exploring a different route: generating supervision directly for real degraded images with MFMs.

\vspace{2mm}
\noindent\textbf{Multimodal Foundation Models.}
Recent generative MFMs have rapidly advanced the tasks such as visual understanding, instruction-based editing, and image generation~\cite{openai2025gptimage15,team2023gemini,Qwen-Image-Edit}. Modern MFMs can produce content-aware HQ outputs from image and text inputs, making them a promising tool for generating HQ targets from LQ inputs. However, real-world IR poses stricter requirements than general image editing, as it demands both perceptual quality improvement and faithful content preservation. Current MFMs may still hallucinate details and behave inconsistently across prompts and degradation types. Moreover, their effectiveness in IR remains insufficiently studied. For example, recent works~\cite{sun2026can} evaluate MFMs only on limited real-world degradation categories, while RealRestorer~\cite{yang2026realrestorer} explores MFMs for restoration but relied only on simple fixed prompts without studying the image fidelity preservation. More importantly, existing efforts have not yet established a complete pipeline for using MFMs to build IR training data, including model selection, prompting design and data screening. In this work, we make a systematic evaluation of state-of-the-art MFMs and present a practical pipeline for dataset construction.

\section{GGT-100K: Dataset Construction}

\begin{figure*}[t]
    \centering
    \includegraphics[width=\linewidth]{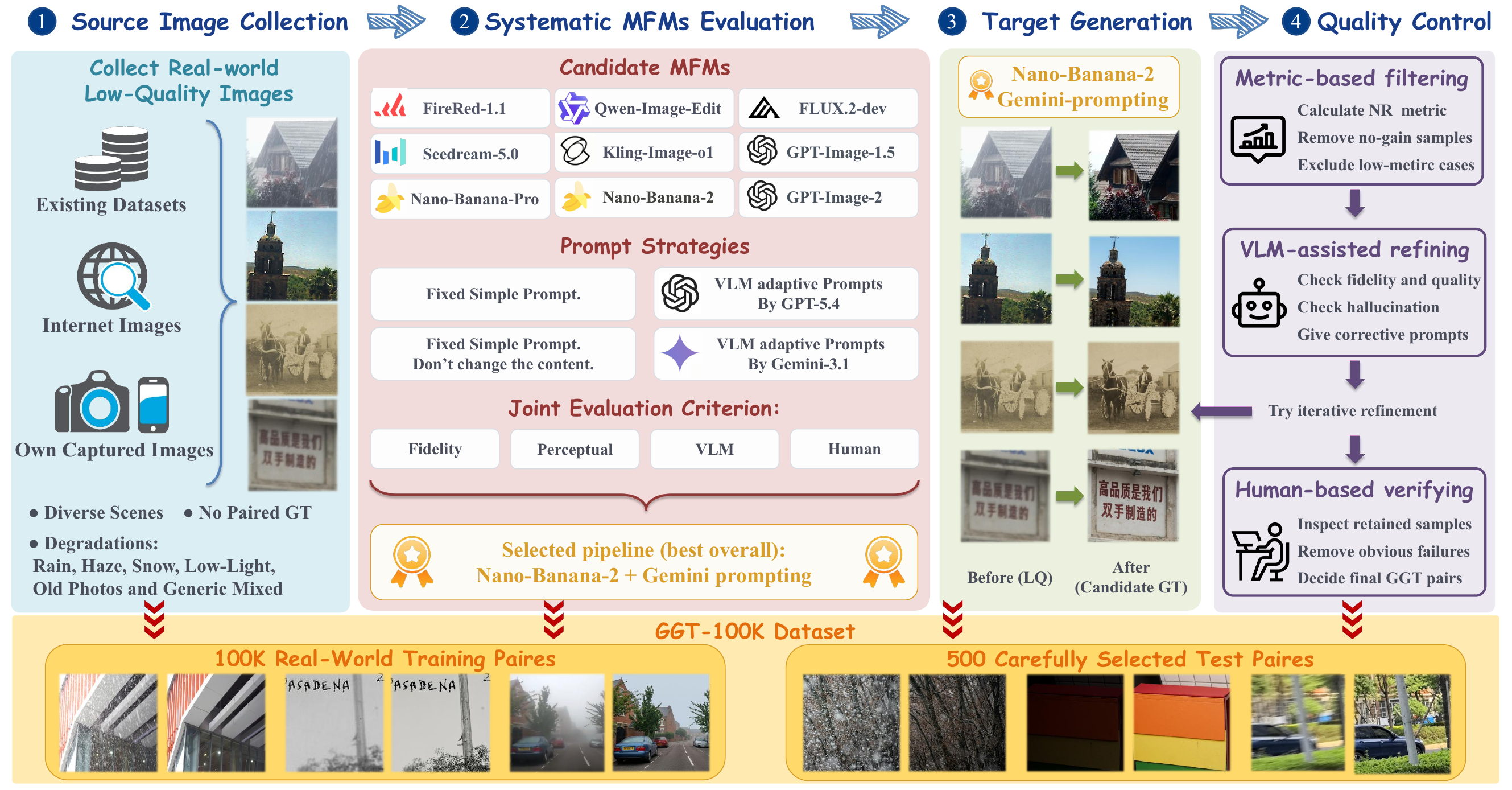}

    \caption{Overview of the GGT-100K construction pipeline. We collect diverse real-world LQ images, evaluate MFMs for HQ target generation, and apply multi-stage quality control to build the dataset.}
    \label{fig:pipeline}

\end{figure*}

This section presents our pipeline for constructing real-world LQ-HQ paired data using MFMs. %Our goal is not to treat MFMs as fully reliable restorers but to use them as controllable generators that can provide faithful HQ targets for real-world IR. 
As illustrated in Fig.~\ref{fig:pipeline}, we begin by collecting real-world LQ images from various sources. Given these inputs, we evaluate candidate MFMs and prompting strategies to identify the generation setting that best balances perceptual quality and fidelity preservation. We then use the selected MFM to generate candidate HQ targets at scale, followed by a multi-stage screening process to retain samples with both strong perceptual quality and high content fidelity, resulting in our GGT-100K.

\subsection{Source Image Collection}
\label{sec:Source Image Collection}

To expand the generalization boundary of real-world IR models, we strategically collect degraded images that lack HQ references and are not covered by existing paired datasets, mainly from three sources: existing datasets, Internet sources, and our own captures. The first source includes restoration datasets without ground-truth references~\cite{RESIDE, snow100k, chen2018robust}, as well as broader vision datasets that contain low-quality or bad-weather images~\cite{deng2009imagenet, Exdark, yang2020advancing, SDV21}. The second source consists of Internet sources collected through web crawling~\cite{unsplash, pexels, pixabay, flickr}, with proper usage rights ensured. The third source is our own captured data, obtained with different cameras and mobile phones across diverse scenes and conditions, covering blur, noise, low-light degradation, and related practical artifacts. After collection, we normalize all candidate images to $1024\times1024$. More details are provided in {\textbf{Appendix~\ref{app:a}}}.

\begin{table}[th]
\centering
\caption{Comparison of different MFMs and prompting strategies, including fidelity metrics, perceptual metrics, VLM-based success rate (VLM-R), average score (Avg.), and human preference. The \textbf{\textcolor{red}{best}}, \textbf{\textcolor{blue}{second-best}}, and \textbf{\textcolor{ForestGreen}{third-best}} results are highlighted in red, blue, and green, respectively.}
\label{tab:model_prompt_excl_realesrgan}

\resizebox{\textwidth}{!}{
\begin{tabular}{c|c|cccc|ccccc|cc|c}
\toprule
\multicolumn{1}{c}{\multirow{2}{*}{Model}} & \multicolumn{1}{|c|}{\multirow{2}{*}{Prompt}} & \multicolumn{4}{c|}{Fidelity results of DIV2K-Val} & \multicolumn{5}{c|}{Perceptual results of 200 Real LQ images} & \multicolumn{1}{c}{\multirow{2}{*}{VLM-R $\uparrow$}} & \multicolumn{1}{c|}{\multirow{2}{*}{Avg.$\uparrow$}} & \multirow{2}{*}{Human $\uparrow$} \\
\cmidrule(lr){3-6}\cmidrule(lr){7-11}
& & PSNR $\uparrow$ & SSIM $\uparrow$ & LPIPS $\downarrow$ & DISTS $\downarrow$ & NIQE $\downarrow$ & MUSIQ $\uparrow$ & MANIQA $\uparrow$ & TOPIQ $\uparrow$ & AFINE-NR $\downarrow$ &  &  \\
\midrule

\multirow{4}{*}{\shortstack{FireRed-1.1~\\(Open-source)\cite{team2026firered}}} 
& Fix & 23.9340 & 0.7520 & 0.2165 & 0.1089 & 4.8022 & 64.6538 & 0.5928 & 0.5013 & -0.9917 & 57.0\% & 0.7224 & \multirow{4}{*}{6.0\%} \\
& Fix-NC & 23.7276 & 0.7493 & 0.2189 & 0.1096 & 4.7554 & 65.0938 & 0.5936 & 0.5041 & \textbf{\textcolor{ForestGreen}{-0.9922}} & 58.0\% & 0.7297 &  \\
& GPT & 19.9648 & 0.6870 & 0.2730 & 0.1272 & \textbf{\textcolor{red}{4.0810}} & 68.2131 & 0.6058 & 0.5452 & \textbf{\textcolor{blue}{-0.9957}} & 23.0\% & 0.4787 &  \\
& Gemini & 21.2432 & 0.7233 & 0.2603 & 0.1237 & 4.5822 & 62.8360 & 0.5798 & 0.4802 & -0.9582 & 52.0\% & 0.6322 &  \\
\midrule
\multirow{4}{*}{\shortstack{Qwen-Image-Edit\\(Open-source)~\cite{Qwen-Image-Edit}}} 
& Fix & \textbf{\textcolor{red}{27.5471}} & \textbf{\textcolor{red}{0.8170}} & 0.1710 & 0.0876 & 5.8545 & 51.7165 & 0.5280 & 0.3966 & -0.9278 & 57.0\% & 0.6602 & \multirow{4}{*}{5.5\%} \\
& Fix-NC & \textbf{\textcolor{blue}{27.2219}} & \textbf{\textcolor{blue}{0.8101}} & 0.1761 & 0.0906 & 6.0303 & 51.4212 & 0.5263 & 0.3977 & -0.9285 & 56.5\% & 0.6433 &  \\
& GPT & 26.1490 & 0.7930 & 0.1830 & 0.0925 & 5.5224 & 54.9141 & 0.5400 & 0.4120 & -0.9250 & 58.5\% & 0.6755 &  \\
& Gemini & 26.1282 & 0.7816 & 0.2082 & 0.1056 & 5.9969 & 51.0975 & 0.5172 & 0.3861 & -0.9171 & 59.5\% & 0.6245 &  \\
\midrule
\multirow{4}{*}{\shortstack{FLUX.2-dev\\(Open-source)~\cite{flux-2-2025}}} 
& Fix & 19.9945 & 0.7314 & 0.2251 & 0.1324 & 5.4224 & 57.4157 & 0.5617 & 0.4777 & -0.8987 & 58.0\% & 0.6058 & \multirow{4}{*}{7.0\%} \\
& Fix-NC & 20.2668 & 0.7371 & 0.2173 & 0.1288 & 5.5124 & 56.4839 & 0.5574 & 0.4700 & -0.8894 & 57.5\% & 0.5994 &  \\
& GPT & 21.3079 & 0.7378 & 0.2029 & 0.1258 & 5.0138 & 60.5277 & 0.5867 & 0.4971 & -0.9105 & 68.0\% & 0.7285 &  \\
& Gemini & 21.1881 & 0.7054 & 0.2351 & 0.1384 & 5.4573 & 55.6259 & 0.5605 & 0.4491 & -0.8947 & 68.5\% & 0.6597 &  \\
\midrule
\multirow{4}{*}{\shortstack{Kling-Image-o1\\(Closed-source)~\cite{Kling-Image-O1}}} 
& Fix & 24.5655 & 0.7422 & 0.2239 & 0.1115 & 5.5488 & 44.0313 & 0.5097 & 0.3233 & -0.7539 & 55.5\% & 0.5187 & \multirow{4}{*}{5.5\%} \\
& Fix-NC & 24.3646 & 0.7334 & 0.2177 & 0.1104 & 5.6452 & 42.2917 & 0.5018 & 0.3114 & -0.7367 & 62.0\% & 0.5451 &  \\
& GPT & 25.9992 & 0.7765 & 0.1824 & 0.1026 & 5.9970 & 39.3803 & 0.4810 & 0.2928 & -0.7291 & 64.5\% & 0.5626 &  \\
& Gemini & 25.1805 & 0.7665 & 0.1880 & 0.1053 & 5.8738 & 39.4746 & 0.4876 & 0.2956 & -0.7172 & 63.5\% & 0.5501 &  \\
\midrule
\multirow{4}{*}{\shortstack{Seedream-5.0\\(Closed-source)~\cite{Seedream5}}} 
& Fix & 22.4237 & 0.7361 & 0.1889 & 0.1086 & 5.1858 & 54.5541 & 0.5339 & 0.4477 & -0.8574 & 69.0\% & 0.6977 & \multirow{4}{*}{7.0\%} \\
& Fix-NC & 22.7089 & 0.7362 & 0.1894 & 0.1089 & 5.2306 & 52.9141 & 0.5278 & 0.4344 & -0.8469 & 64.5\% & 0.6568 &  \\
& GPT & 23.3330 & 0.7515 & 0.1909 & 0.1088 & 5.0132 & 54.0857 & 0.5370 & 0.4597 & -0.8566 & 61.0\% & 0.6607 &  \\
& Gemini & 23.7293 & 0.7477 & 0.1856 & 0.1103 & 5.3304 & 53.0703 & 0.5231 & 0.4380 & -0.8510 & 68.0\% & 0.6855 &  \\
\midrule
\multirow{4}{*}{\shortstack{GPT-Image-1.5\\(Closed-source)~\cite{openai2025gptimage15}}} 
& Fix & 14.5390 & 0.3559 & 0.4902 & 0.1849 & 4.4747 & \textbf{\textcolor{blue}{69.6380}} & 0.6226 & \textbf{\textcolor{blue}{0.5842}} & -0.9675 & 31.0\% & 0.3412 & \multirow{4}{*}{0\%} \\
& Fix-NC & 14.6410 & 0.3567 & 0.4854 & 0.1822 & 4.6371 & 67.6234 & 0.6114 & 0.5549 & -0.9427 & 41.0\% & 0.3860 &  \\
& GPT & 15.0862 & 0.3665 & 0.4706 & 0.1801 & 4.8494 & 66.1326 & 0.6083 & 0.5449 & -0.9112 & 39.5\% & 0.3646 &  \\
& Gemini & 14.8436 & 0.3560 & 0.4719 & 0.1814 & 4.7243 & 66.4074 & 0.6151 & 0.5519 & -0.9216 & 34.0\% & 0.3344 &  \\
\midrule
\multirow{4}{*}{\shortstack{GPT-Image-2\\(Closed-source)~\cite{openai2025gptimage2}}} 
& Fix & 16.7439 & 0.4410 & 0.3374 & 0.1381 & 4.4268 & \textbf{\textcolor{red}{70.8689}} & \textbf{\textcolor{red}{0.6651}} & \textbf{\textcolor{red}{0.6166}} & \textbf{\textcolor{red}{-0.9984}} & 57.0\% & 0.6479 & \multirow{4}{*}{\textbf{\textcolor{blue}{21.5\%}}} \\
& Fix-NC & 18.8703 & 0.5157 & 0.2503 & 0.1149 & 4.5807 & \textbf{\textcolor{ForestGreen}{68.3071}} & \textbf{\textcolor{blue}{0.6503}} & \textbf{\textcolor{ForestGreen}{0.5741}} & -0.9672 & 65.0\% & 0.7336 &  \\
& GPT & 17.8234 & 0.4830 & 0.2977 & 0.1143 & 4.9140 & 63.9640 & 0.6275 & 0.5020 & -0.9304 & \textbf{\textcolor{red}{73.0\%}} & 0.7115 &  \\
& Gemini & 18.2098 & 0.4995 & 0.2722 & 0.1137 & 4.7421 & 67.1697 & \textbf{\textcolor{ForestGreen}{0.6429}} & 0.5558 & -0.9532 & 67.5\% & 0.7213 &  \\
\midrule
\multirow{4}{*}{\shortstack{Nano-Banana-Pro\\(Closed-source)~\cite{team2023gemini}}} 
& Fix & 25.5366 & 0.7671 & 0.1588 & 0.0845 & 4.8273 & 57.1012 & 0.5739 & 0.4373 & -0.8739 & 64.5\% & 0.7522 & \multirow{4}{*}{\textbf{\textcolor{ForestGreen}{15.0\%}}} \\
& Fix-NC & 26.5977 & 0.7865 & 0.1375 & 0.0780 & 5.0341 & 54.0572 & 0.5573 & 0.4063 & -0.8473 & 68.5\% & 0.7667 &  \\
& GPT & 27.0899 & 0.7917 & 0.1327 & 0.0753 & 5.3172 & 49.5661 & 0.5361 & 0.3681 & -0.8234 & \textbf{\textcolor{blue}{72.0\%}} & 0.7569 &  \\
& Gemini & 26.8784 & 0.7816 & 0.1429 & 0.0786 & 5.2612 & 51.0284 & 0.5440 & 0.3793 & -0.8324 & \textbf{\textcolor{ForestGreen}{70.0\%}} & 0.7479 &  \\
\midrule
\multirow{4}{*}{\shortstack{Nano-Banana-2\\(Closed-source)~\cite{team2023gemini}}} 
& Fix & 25.6879 & 0.7758 & 0.1557 & 0.0856 & \textbf{\textcolor{blue}{4.2132}} & 64.4820 & 0.6185 & 0.5189 & -0.9393 & 52.0\% & 0.7564 & \multirow{4}{*}{\textbf{\textcolor{Red}{32.5\%}}} \\
& Fix-NC & 26.1795 & 0.7824 & \textbf{\textcolor{ForestGreen}{0.1418}} & \textbf{\textcolor{ForestGreen}{0.0814}} & \textbf{\textcolor{ForestGreen}{4.4061}} & 62.7827 & 0.6089 & 0.4958 & -0.9186 & 63.0\% & \textbf{\textcolor{blue}{0.8171}} &  \\
& GPT & 26.7364 & 0.7923 & \textbf{\textcolor{blue}{0.1300}} & \textbf{\textcolor{blue}{0.0757}} & 4.8221 & 56.6527 & 0.5765 & 0.4272 & -0.8825 & 69.0\% & \textbf{\textcolor{ForestGreen}{0.8078}} &  \\
& Gemini & \textbf{\textcolor{ForestGreen}{27.1701}} & \textbf{\textcolor{ForestGreen}{0.7949}} & \textbf{\textcolor{red}{0.1280}} & \textbf{\textcolor{red}{0.0738}} & 4.6881 & 59.2831 & 0.5872 & 0.4536 & -0.8980 & \textbf{\textcolor{ForestGreen}{70.0\%}} & \textbf{\textcolor{Red}{0.8427}} &  \\
\bottomrule
\end{tabular}}

\end{table}

\subsection{Systematic Evaluation of MFMs}
\label{sec:Systematic MFMs Evaluation}

\vspace{2mm}
\noindent\textbf{Candidate MFMs and Prompts.} We evaluate models and prompts jointly because they are tightly coupled. For models, we evaluate 3 open-source MFMs (FireRed-1.1~\cite{team2026firered}, Qwen-Image-Edit-2511~\cite{Qwen-Image-Edit} and FLUX.2-dev~\cite{flux-2-2025}) and 6 closed-source MFMs (Kling-Image-O1~\cite{Kling-Image-O1}, Seedream-5.0~\cite{Seedream5}, GPT-Image-1.5~\cite{openai2025gptimage15}, GPT-Image-2~\cite{openai2025gptimage2}, Nano-Banana-Pro~\cite{team2023gemini} and Nano-Banana-2~\cite{team2023gemini}). For prompts, we consider both fixed and adaptive ones. Fixed prompts share a restoration instruction for each input category, while a prompt variant (denoted as ``Fix-NC'' in Tab.~\ref{tab:model_prompt_excl_realesrgan}) explicitly requires the image content to remain unchanged. For adaptive prompts, we first use a VLM (GPT-5.4-Pro~\cite{gpt5.4} or Gemini-3.1-Pro~\cite{team2023gemini}) to analyze the input image and then generate an image-specific instruction based on its content and degradation. Detailed prompt designs are provided in {\textbf{Appendix~\ref{app:b}}}.

\vspace{2mm}
\noindent\textbf{Evaluation Criterion.} We consider four complementary aspects in evaluation: \emph{fidelity preservation}, \emph{perceptual quality}, \emph{VLM-based assessment}, and \emph{human preference}. As shown in Tab.~\ref{tab:model_prompt_excl_realesrgan}, for fidelity evaluation, we use 100 DIV2K validation images~\cite{div2k}, degrade them with the Real-ESRGAN pipeline~\cite{wang2021real}, and compare the generated results with the original clean images using four full-reference metrics. We exclude tasks such as low-light enhancement and dehazing from this part because their acceptable outputs may differ in global brightness or illumination, which can bias full-reference evaluation. For perceptual evaluation, we use 200 collected real-world images, including 20 images for each of the rain, haze, snow, low-light, and old photo categories, together with 100 images with general mixed degradations, and assess the outputs with five no-reference perceptual metrics. To evaluate the problems that cannot be well reflected by IQA metrics, such as hallucinated objects or unreasonable edits, we further introduce a VLM-based assessment and report the VLM-estimated restoration success rate (VLM-R). {Details of VLM-R are provided in \textbf{Appendix~\ref{app:c}}}. 

We first convert all metrics to the same direction so that higher values indicate better performance. Denote by ${m}_{i,j}$ the value of metric $j$ for the $i$-th model-prompt. We first apply min--max normalization to obtain the normalized metric $\tilde{m}_{i,j}$ of ${m}_{i,j}$. Then we average the $\tilde{m}_{i,j}$ values along each aspect to obtain aspect-level scores $s_i^{a}$, where aspect $a\in\{\mathrm{fid},\mathrm{per},\mathrm{vlm}\}$. Finally, we average the three aspect-level scores to obtain $\mathrm{Avg.}_i$, the final overall score of the $i$-th model-prompt setting. The whole calculation process can be summarized as follows:
\begin{equation}
\small
\tilde{m}_{i,j}=\frac{m_{i,j}-m_j^{\min}}{m_j^{\max}-m_j^{\min}},\qquad
s_i^{a}=\frac{1}{|\mathcal{M}_a|}\sum_{j\in\mathcal{M}_a} \tilde{m}_{i,j},\qquad
\mathrm{Avg.}_i=\frac{1}{3}\left(s_i^{\mathrm{fid}}+s_i^{\mathrm{per}}+s_i^{\mathrm{vlm}}\right).
\end{equation}

We also conduct a user study {(details are in \textbf{Appendix~\ref{app:user}})} on these 200 real-world samples: for each of the nine MFMs, we select its best-performing prompt setting, present the LQ input together with nine anonymous results, and record how often each method is chosen as the best (``Human'' in Tab.~\ref{tab:model_prompt_excl_realesrgan}). 

\vspace{2mm}
\noindent\textbf{Evaluation Results.} The results are reported in Tab.~\ref{tab:model_prompt_excl_realesrgan}. First, we see that both model choice and prompting strategy have a substantial impact on restoration behavior, and the performance gap across different model--prompt combinations can be large. Even for the same MFM, different prompts may lead to noticeably different outcomes. For example, Nano-Banana-2 improves from an Avg. score of 0.76 under fixed prompting to 0.84 under Gemini-based adaptive prompting. This indicates that prompting sensitivity should be taken seriously when evaluating and using MFMs for restoration. %More detailed analyses are provided in the \textbf{appendix}.

\begin{figure*}[t]
    \centering
    \includegraphics[width=\linewidth]{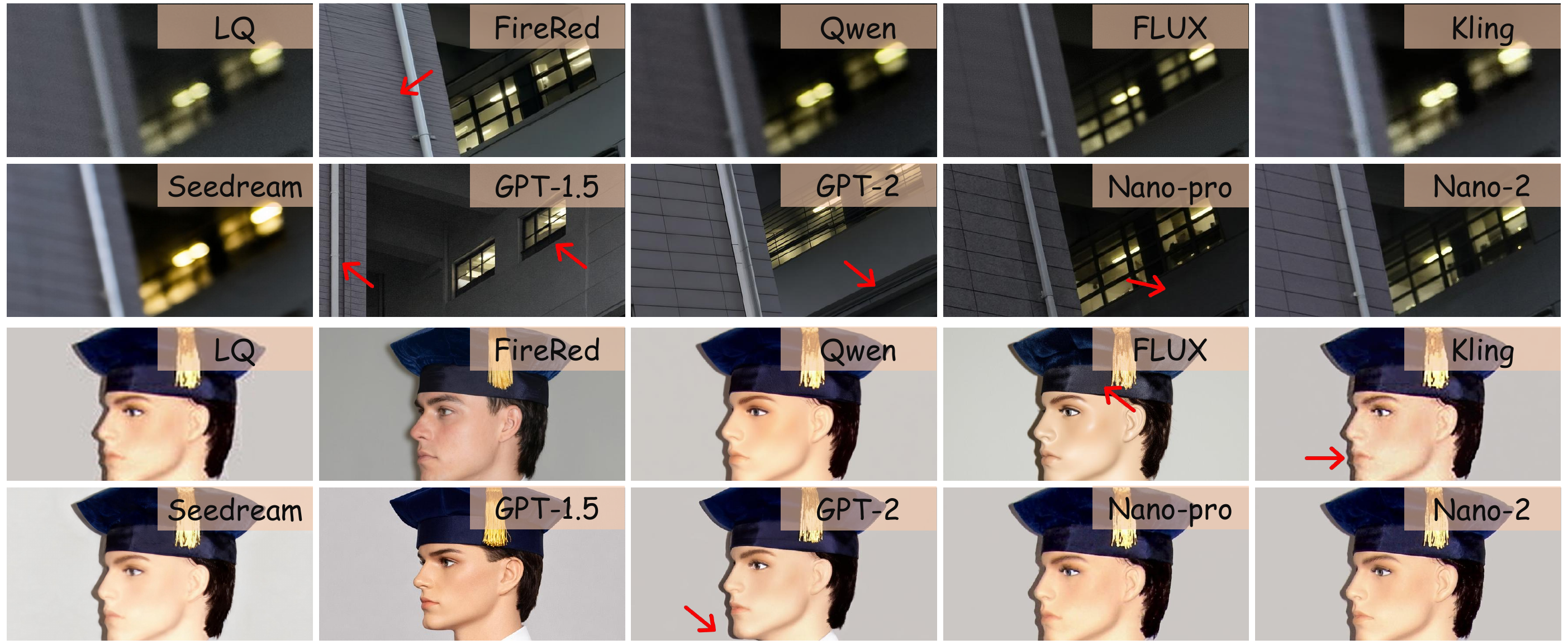}

    \caption{Typical outputs of the MFMs on real-world LQ inputs. For each model, we show its best-performing prompt setting. The examples highlight the different restoration preferences of these MFMs, especially the trade-off between perceptual enhancement and fidelity preservation.}
    \label{fig:mfmvis}

\end{figure*}

Second, different MFMs exhibit clear preference biases. As shown in Fig.~\ref{fig:mfmvis} and Tab.~\ref{tab:model_prompt_excl_realesrgan}, some models preserve input content better but produce conservative restorations, while others generate stronger visual enhancement at the cost of content inconsistency. For example, Qwen-Image-Edit and Kling-Image-o1 achieve high fidelity scores, but their perceptual scores and visual results are relatively modest. In contrast, GPT-Image family and the FireRed-1.1 obtain better perceptual metrics, but their fidelity is much weaker and they often significantly change the content of the image. 

Overall, \textbf{Nano-Banana-2 with Gemini-based adaptive prompting} is the most well-balanced setting among all candidates. It achieves the best Avg. score of 0.84 and the highest human preference of 32.5\%. %Under full-reference fidelity evaluation, it delivers the strongest overall performance. 
Although it does not achieve the best score on every perceptual metric, it achieves highly competitive perceptual results. %Its VLM-based success rate reaches 70.0\%, ranking among the top settings. Overall, 
Rather than excelling in only one aspect, it performs strongly and consistently across all the four aspects, maintaining a favorable balance between perceptual quality and content faithfulness. We therefore choose it as the MFM for GGT-100K construction.

\subsection{Multi-stage Quality Control}
\label{sec:Multi-stage Quality Control}

Even with the best-performing MFM and adaptive prompting strategy, we cannot guarantee that the generated outputs are always ideal HQ targets. Therefore, we further apply a multi-stage quality control process to improve the reliability of the generated HQ targets.%, including \textit{metric-based filtering}, \textit{VLM-assisted refinement}, and \textit{final manual verification}.

\vspace{2mm}
\noindent\textbf{Metric-based Filtering.} We first conduct coarse automatic filtering by no-reference perceptual quality metrics. Specifically, for each LQ image and its  HQ counterpart generated by Nano-Banana-2, we compare their metric scores. If the score shows little improvement or even worse, we regard this sample as a failure case and exclude it from the dataset. %This stage therefore acts as an efficient first-pass filter that removes low-value samples and reduces the burden on subsequent screening stages. Since the current no-reference metrics are not specifically designed for different weather conditions, we are mainly taking this step for the general mixed images.

% \noindent\textbf{VLM-assisted Refinement.} Metric-based filtering is useful for coarse selection, but it cannot reliably capture more subtle MFM failures such as local structure distortion, semantic inconsistency, or unreasonable edits. To address this limitation, we introduce a VLM-assisted refinement stage using Gemini-3.1-Pro~\cite{team2023gemini}. For each generated result, the VLM assesses five aspects that are important for HQ target quality: \emph{restoration quality}, \emph{object consistency}, \emph{geometry alignment}, \emph{content reasonableness}, and \emph{color consistency}. These dimensions reflect whether the generated image not only looks better, but also remains faithful to the original input. For samples judged unsatisfactory, we do not discard them directly; instead, we use the VLM feedback to perform regeneration by appending additional prompt instructions tailored to the identified issues. The regenerated candidate is then re-evaluated. In this way, the VLM serves both as a quality controller and as a feedback source for iterative refinement.

\vspace{2mm}
\noindent\textbf{VLM-assisted Refinement.} Metric-based filtering is useful for coarse selection, but it cannot reliably capture more subtle MFM failures such as local structure distortion, semantic inconsistency, or unreasonable edits. We thus introduce a VLM-assisted refinement step. For each generated result, the VLM~\cite{team2023gemini} assesses five aspects that are important for HQ target: \emph{restoration quality}, \emph{object consistency}, \emph{geometry alignment}, \emph{content reasonableness}, and \emph{color consistency}. For samples judged unsatisfactory, we do not discard them directly; instead, we use the VLM feedback to perform regeneration by appending additional prompt instructions tailored to the identified issues. The regenerated sample is then re-evaluated. In this way, the VLM serves as both a quality controller and a feedback source for iterative refinement. {Details are provided in \textbf{Appendix~\ref{app:c}}}. 

\vspace{2mm} \noindent\textbf{Manual Verification.} Finally, we perform manual verification as the last safeguard of the pipeline. Human annotators review the automatically retained samples and remove those that still contain noticeable artifacts, structural inconsistency or unreasonable content changes. 

By combining metric filtering, VLM-assisted refinement, and manual verification, we obtain a more reliable set of restoration-oriented targets to build GGT-100K. As shown in Fig.~\ref{fig:teaser}, GGT-100K contains 103K training pairs and 500 test pairs at a unified resolution of $1024\times1024$. In particular, the 500 test pairs are jointly selected through careful manual review by multiple researchers to ensure high fidelity, strong restoration quality, and no obvious hallucinated content. %This pipeline processes real-world LQ images into generated HQ targets through MFMs, which avoids the main constraints of paired real-image capture. 
As a result, GGT-100K contains diverse and real degradations arising naturally in real photography, storage, and transmission, making it particularly suitable for improving the generalization performance of real-world IR models. % More details of the datasets can be found in the \textbf{appendix}.

\begin{table*}[t]
\centering
\caption{Comparison of representative restoration models ``w/o'' and ``w/'' GGT-100K on our test set. For some models whose official releases can also handle real-world degradations, we additionally report their \textbf{official results}. ``Improvement'' indicates the performance gain brought by GGT-100K. The positive improvements are highlighted in \textcolor{red}{\textbf{red}}.}
\label{tab:main_ours}
\vspace{1mm}
\resizebox{\textwidth}{!}{
\begin{tabular}{c|c|cccc|ccccc|c}
\toprule
\multirow{2}{*}{Model} & \multirow{2}{*}{GGT-100K} & \multicolumn{4}{c|}{Full-reference fidelity metrics} & \multicolumn{5}{c|}{No-reference perceptual metrics} & \multirow{2}{*}{VLM-R $\uparrow$} \\
\cmidrule(lr){3-6}\cmidrule(lr){7-11}
& & PSNR $\uparrow$ & SSIM $\uparrow$ & LPIPS $\downarrow$ & DISTS $\downarrow$ & NIQE $\downarrow$ & MUSIQ $\uparrow$ & MANIQA $\uparrow$ & TOPIQ $\uparrow$ & AFINE-NR $\downarrow$ &  \\
\midrule
\multirow{3}{*}{MPRNet~\cite{zamir2021multi}}
& w/o & 24.7919 & 0.7637 & 0.3779 & 0.2390 & 6.9774 & 42.1039 & 0.4641 & 0.3120 & -0.7140 & 22.2\% \\
& w/ & 27.3044 & 0.8189 & 0.3377 & 0.2212 & 7.0035 & 44.9597 & 0.4431 & 0.3075 & -0.7794 & 33.2\% \\
& Improvement & \textcolor{red}{\textbf{+2.5125}} & \textcolor{red}{\textbf{+0.0552}} & \textcolor{red}{\textbf{-0.0402}} & \textcolor{red}{\textbf{-0.0178}} & +0.0261 & \textcolor{red}{\textbf{+2.8558}} & -0.0210 & -0.0045 & \textcolor{red}{\textbf{-0.0654}} & \textcolor{red}{\textbf{+11.0\%}} \\
\midrule
\multirow{3}{*}{NAFNet~\cite{chen2022simple}}
& w/o & 25.1255 & 0.7708 & 0.3653 & 0.2298 & 6.6654 & 42.0124 & 0.4635 & 0.3118 & -0.7012 & 27.6\% \\
& w/ & 28.2461 & 0.8349 & 0.3110 & 0.2043 & 6.7983 & 46.7094 & 0.4330 & 0.3097 & -0.7881 & 53.8\% \\
& Improvement & \textcolor{red}{\textbf{+3.1206}} & \textcolor{red}{\textbf{+0.0641}} & \textcolor{red}{\textbf{-0.0543}} & \textcolor{red}{\textbf{-0.0255}} & +0.1329 & \textcolor{red}{\textbf{+4.6970}} & -0.0305 & -0.0021 & \textcolor{red}{\textbf{-0.0869}} & \textcolor{red}{\textbf{+26.2\%}} \\
\midrule
\multirow{3}{*}{SwinIR~\cite{liang2021swinir}}
& w/o & 23.9225 & 0.7590 & 0.3878 & 0.2343 & 6.4979 & 41.0176 & 0.4488 & 0.3082 & -0.6940 & 18.6\% \\
& w/ & 27.0781 & 0.8150 & 0.3369 & 0.2131 & 6.7366 & 43.6689 & 0.4308 & 0.2966 & -0.7437 & 37.6\% \\
& Improvement & \textcolor{red}{\textbf{+3.1556}} & \textcolor{red}{\textbf{+0.0560}} & \textcolor{red}{\textbf{-0.0509}} & \textcolor{red}{\textbf{-0.0212}} & +0.2387 & \textcolor{red}{\textbf{+2.6513}} & -0.0180 & -0.0116 & \textcolor{red}{\textbf{-0.0497}} & \textcolor{red}{\textbf{+19.0\%}} \\
\midrule
\multirow{3}{*}{X-Restormer~\cite{X-Restormer}}
& w/o & 24.6901 & 0.7705 & 0.3587 & 0.2234 & 6.6564 & 43.1437 & 0.4643 & 0.3192 & -0.7185 & 30.4\% \\
& w/ & 28.2298 & 0.8362 & 0.3130 & 0.2065 & 6.9218 & 46.8069 & 0.4295 & 0.3109 & -0.8126 & 54.6\% \\
& Improvement & \textcolor{red}{\textbf{+3.5397}} & \textcolor{red}{\textbf{+0.0657}} & \textcolor{red}{\textbf{-0.0457}} & \textcolor{red}{\textbf{-0.0169}} & +0.2654 & \textcolor{red}{\textbf{+3.6632}} & -0.0348 & -0.0083 & \textcolor{red}{\textbf{-0.0941}} & \textcolor{red}{\textbf{+24.2\%}} \\
\midrule
\multirow{3}{*}{PromptIR~\cite{PromptIR}}
& w/o & 24.5733 & 0.7564 & 0.3630 & 0.2312 & 6.5942 & 43.2318 & 0.4695 & 0.3202 & -0.7176 & 24.8\% \\
& w/ & 28.1775 & 0.8344 & 0.3113 & 0.2051 & 6.8029 & 46.7198 & 0.4369 & 0.3124 & -0.7968 & 49.6\% \\
& Improvement & \textcolor{red}{\textbf{+3.6042}} & \textcolor{red}{\textbf{+0.0780}} & \textcolor{red}{\textbf{-0.0517}} & \textcolor{red}{\textbf{-0.0261}} & +0.2087 & \textcolor{red}{\textbf{+3.4880}} & -0.0326 & -0.0078 & \textcolor{red}{\textbf{-0.0792}} & \textcolor{red}{\textbf{+24.8\%}} \\
\midrule
\multirow{3}{*}{MoCE-IR~\cite{MoCE-IR}}
& w/o & 24.8641 & 0.7575 & 0.3654 & 0.2291 & 6.6248 & 43.0279 & 0.4671 & 0.3200 & -0.7116 & 25.4\% \\
& w/ & 28.2140 & 0.8392 & 0.3097 & 0.2093 & 6.9402 & 48.4846 & 0.4344 & 0.3242 & -0.8270 & 55.2\% \\
& Improvement & \textcolor{red}{\textbf{+3.3499}} & \textcolor{red}{\textbf{+0.0817}} & \textcolor{red}{\textbf{-0.0557}} & \textcolor{red}{\textbf{-0.0198}} & +0.3154 & \textcolor{red}{\textbf{+5.4567}} & -0.0327 & \textcolor{red}{\textbf{+0.0042}} & \textcolor{red}{\textbf{-0.1154}} & \textcolor{red}{\textbf{+29.8\%}} \\
\midrule
\multirow{3}{*}{DA-CLIP~\cite{DA-CLIP}}
& w/o & 25.8809 & 0.7726 & 0.3427 & 0.2180 & 6.3011 & 39.3370 & 0.4846 & 0.3080 & -0.6853 & 31.6\% \\
& w/ & 26.6914 & 0.7910 & 0.2954 & 0.1938 & 6.3039 & 41.8839 & 0.4834 & 0.3073 & -0.7277 & 51.0\% \\
& Improvement & \textcolor{red}{\textbf{+0.8105}} & \textcolor{red}{\textbf{+0.0184}} & \textcolor{red}{\textbf{-0.0473}} & \textcolor{red}{\textbf{-0.0242}} & +0.0028 & \textcolor{red}{\textbf{+2.5469}} & -0.0012 & -0.0007 & \textcolor{red}{\textbf{-0.0424}} & \textcolor{red}{\textbf{+19.4\%}} \\
\midrule
\multirow{4}{*}{FoundIR~\cite{li2025foundir}}
& \textbf{official} & 26.0398 & 0.7866 & 0.3486 & 0.2199 & 6.5311 & 39.3646 & 0.4793 & 0.3048 & -0.6971 & 28.8\% \\
& w/o & 25.8048 & 0.7844 & 0.3508 & 0.2220 & 6.9023 & 42.5388 & 0.4701 & 0.3153 & -0.7365 & 35.8\% \\
& w/ & 27.1777 & 0.8213 & 0.3351 & 0.2209 & 7.4717 & 43.9238 & 0.4402 & 0.3019 & -0.8087 & 60.8\% \\
& Improvement & \textcolor{red}{\textbf{+1.3729}} & \textcolor{red}{\textbf{+0.0369}} & \textcolor{red}{\textbf{-0.0157}} & \textcolor{red}{\textbf{-0.0011}} & +0.5694 & \textcolor{red}{\textbf{+1.3850}} & -0.0299 & -0.0134 & \textcolor{red}{\textbf{-0.0722}} & \textcolor{red}{\textbf{+25.0\%}} \\
\midrule
\multirow{3}{*}{FLUX-Controlnet~\cite{FLUX}}
& w/o & 22.4486 & 0.6901 & 0.3773 & 0.2129 & 5.5636 & 48.5454 & 0.5157 & 0.3798 & -0.6912 & 25.4\% \\
& w/ & 23.1413 & 0.7325 & 0.2625 & 0.1520 & 4.8504 & 63.0910 & 0.5854 & 0.5013 & -0.9280 & 63.4\% \\
& Improvement & \textcolor{red}{\textbf{+0.6927}} & \textcolor{red}{\textbf{+0.0424}} & \textcolor{red}{\textbf{-0.1148}} & \textcolor{red}{\textbf{-0.0609}} & \textcolor{red}{\textbf{-0.7132}} & \textcolor{red}{\textbf{+14.5456}} & \textcolor{red}{\textbf{+0.0697}} & \textcolor{red}{\textbf{+0.1215}} & \textcolor{red}{\textbf{-0.2368}} & \textcolor{red}{\textbf{+38.0\%}} \\
\midrule
\multirow{4}{*}{Qwen-Image-Edit~\cite{Qwen-Image-Edit}}
& \textbf{official} & 22.3141 & 0.7479 & 0.3042 & 0.1693 & 5.4776 & 60.8628 & 0.5576 & 0.4709 & -0.9565 & 68.0\% \\
& w/o & 25.8559 & 0.7787 & 0.2813 & 0.1625 & 6.0198 & 51.4215 & 0.5333 & 0.3765 & -0.8401 & 77.4\% \\
& w/ & 26.1811 & 0.7828 & 0.2155 & 0.1183 & 5.4648 & 62.5519 & 0.5811 & 0.4707 & -0.9611 & 87.6\% \\
& Improvement & \textcolor{red}{\textbf{+0.3252}} & \textcolor{red}{\textbf{+0.0041}} & \textcolor{red}{\textbf{-0.0658}} & \textcolor{red}{\textbf{-0.0442}} & \textcolor{red}{\textbf{-0.5550}} & \textcolor{red}{\textbf{+11.1304}} & \textcolor{red}{\textbf{+0.0478}} & \textcolor{red}{\textbf{+0.0942}} & \textcolor{red}{\textbf{-0.1210}} & \textcolor{red}{\textbf{+10.2\%}} \\
\bottomrule
\end{tabular}}
\end{table*}

\section{Expanding Real-World IR Boundaries using GGT-100K}

In this section, we demonstrate how GGT-100K can improve the real-world generalization ability of a variety of IR models. Specifically, we train a set of representative IR models under different data settings and compare their performance on our test set and existing real-world test sets.

\subsection{Experimental Settings}
\label{sec:Experimental Settings}

\vspace{2mm} \noindent\textbf{Representative IR Models.} We employ a series of representative restoration models, including $L_1$-loss optimized CNN and transformer backbones (MPRNet~\cite{zamir2021multi}, NAFNet~\cite{chen2022simple}, SwinIR~\cite{liang2021swinir}, X-Restormer~\cite{X-Restormer}) and all-in-one models (PromptIR~\cite{PromptIR}, MoCE-IR~\cite{MoCE-IR}, DA-CLIP~\cite{DA-CLIP}, FoundIR~\cite{li2025foundir}). In addition, since finetuning large-scale generative models has shown impressive real-world IR performance~\cite{wu2024seesr,yu2024scaling}, we also finetune the T2I model FLUX.1-dev~\cite{FLUX} and the TI2I model Qwen-Image-Edit-2511~\cite{Qwen-Image-Edit} on GGT-100K in the experiments. % Since real-world restoration has broad practical relevance, models of different scales and paradigms are all meaningful to study, even CNN-based methods remain important in resource-constrained applications. Together, these methods allow us to verify that GGT-100K can enhance the generalization ability of diverse mainstream restoration methods.

%\vspace{1mm}
\vspace{2mm} \noindent\textbf{Dataset Settings.} To validate the effectiveness of GGT-100K for real-world IR, we train IR models under two settings. \textbf{(1) Existing training data (``w/o'' in Tab.~\ref{tab:main_ours})}. {This setting contains 200K training pairs from 15 existing synthetic and real-world datasets. We control the category composition to ensure a fair comparison: the general mixed category contains 120K pairs, roughly matching its proportion in GGT-100K, while the remaining pairs cover low-light, haze, rain, and snow. For most datasets, we keep the original image resolution unchanged. For very high-resolution datasets such as RealSR~\cite{realsr} and SIDD~\cite{sidd}, we first crop images into $512\times512$ patches and then sample from the resulting pool. More details are in \textbf{Appendix~\ref{app:e}}.} \textbf{(2) Existing training data + GGT-100K (``w/'' in Tab.~\ref{tab:main_ours})}, which adds GGT-100K to the training pool with a \textbf{1:1 sampling ratio}. This reflects the intended use of GGT-100K in practice:  \textit{as a complementary real-world data source to expand the generalization boundary of IR models rather than a replacement for existing datasets}. For testing sets, we use two benchmark groups: our GGT-100K test set with paired GGT references, and public real-world test sets without GT (social media and old photo of RealDeg~\cite{chen2024faithdiff}, and OpenReal80K~\cite{jarvisir2025} subsets covering haze, rain, snow, and night scenarios).

%\vspace{1mm}
\vspace{2mm} \noindent\textbf{Evaluation.} Similar to Sec.~\ref{sec:Systematic MFMs Evaluation}, we use full-reference fidelity metrics, including PSNR, SSIM~\cite{ssim}, LPIPS~\cite{lpips}, and DISTS~\cite{dists}, to measure content fidelity, and use no-reference metrics, including NIQE~\cite{niqe}, MUSIQ~\cite{musiq}, MANIQA~\cite{maniqa}, TOPIQ~\cite{chen2024topiq}, and AFINE-NR~\cite{afine}, to assess perceptual restoration quality. We further report the VLM-based success rate (VLM-R) to evaluate restoration success from a multimodal semantic perspective like that in Sec.~\ref{sec:Systematic MFMs Evaluation}.

%\vspace{1mm}
\vspace{2mm} \noindent\textbf{Implementation Details.} For models trained from scratch, we follow their original experimental settings. %, keeping the patch size, batch size, and learning rate unchanged from the corresponding official implementations. 
For FLUX.1-dev~\cite{FLUX} and Qwen-Image-Edit-2511~\cite{Qwen-Image-Edit}, we use a simple prompt during finetuning and set the input resolution to $512\times512$ by randomly cropping images from GGT-100K. We finetune FLUX.1-dev with a ControlNet-style setup using 4 double blocks and 0 single block, with a batch size of 16, a learning rate of $5\times10^{-5}$, for 200K iterations. We finetune Qwen-Image-Edit-2511 with LoRA~\cite{lora} of rank 8, a batch size of 8, a learning rate of $1\times10^{-4}$, for 50K iterations. All experiments are implemented in PyTorch~\cite{paszke2019pytorch} on 32 NVIDIA A800 GPUs.

\begin{table*}[t]
\centering
\caption{Comparison of representative restoration models ``w/o'' and ``w/'' GGT-100K on public RealLQ test sets without GT, including social media~\cite{chen2024faithdiff}, old photos~\cite{chen2024faithdiff}, rain~\cite{jarvisir2025}, haze~\cite{jarvisir2025}, snow~\cite{jarvisir2025}, and low light~\cite{jarvisir2025}. The average results over the six test sets are reported. ``Improvement'' indicates the performance gain brought by using GGT-100K. Positive improvements are highlighted in \textcolor{red}{\textbf{red}}.}
\label{tab:public_reallq}
\vspace{1mm}
\resizebox{\textwidth}{!}{
\begin{tabular}{c|c|cc|cc|cc|cc|cc|cc}
\toprule
\multirow{2}{*}{Model} & \multirow{2}{*}{GGT-100K} & \multicolumn{2}{c|}{NIQE $\downarrow$} & \multicolumn{2}{c|}{MUSIQ $\uparrow$} & \multicolumn{2}{c|}{MANIQA $\uparrow$} & \multicolumn{2}{c|}{TOPIQ $\uparrow$} & \multicolumn{2}{c|}{AFINE-NR $\downarrow$} & \multicolumn{2}{c}{VLM-R $\uparrow$}\\
\cmidrule(lr){3-4}\cmidrule(lr){5-6}\cmidrule(lr){7-8}\cmidrule(lr){9-10}\cmidrule(lr){11-12}\cmidrule(lr){13-14}
& & Value & Imp. & Value & Imp. & Value & Imp. & Value & Imp. & Value & Imp. & Value & Imp. \\
\midrule
\multirow{2}{*}{MPRNet~\cite{zamir2021multi}}
& w/o & 5.1502 & \multirow{2}{*}{\textcolor{red}{\textbf{-0.0529}}} & 56.2493 & \multirow{2}{*}{\textcolor{red}{\textbf{+2.3796}}} & 0.5928 & \multirow{2}{*}{\textcolor{red}{\textbf{+0.0009}}} & 0.4427 & \multirow{2}{*}{\textcolor{red}{\textbf{+0.0113}}} & -0.8439 & \multirow{2}{*}{\textcolor{red}{\textbf{-0.0323}}} & 38.0\% & \multirow{2}{*}{\textcolor{red}{\textbf{+7.0\%}}} \\
& w/ & 5.0973 &  & 58.6289 &  & 0.5937 &  & 0.4540 &  & -0.8762 &  & 45.0\% &  \\
\midrule
\multirow{2}{*}{NAFNet~\cite{chen2022simple}}
& w/o & 4.9061 & \multirow{2}{*}{+0.1442} & 57.9680 & \multirow{2}{*}{-0.4102} & 0.5986 & \multirow{2}{*}{-0.0145} & 0.4576 & \multirow{2}{*}{-0.0228} & -0.8559 & \multirow{2}{*}{\textcolor{red}{\textbf{-0.0106}}} & 36.0\% & \multirow{2}{*}{\textcolor{red}{\textbf{+9.0\%}}} \\
& w/ & 5.0503 &  & 57.5578 &  & 0.5841 &  & 0.4348 &  & -0.8665 &  & 45.0\% &  \\
\midrule
\multirow{2}{*}{SwinIR~\cite{liang2021swinir}}
& w/o & 4.8474 & \multirow{2}{*}{+0.2517} & 56.3785 & \multirow{2}{*}{\textcolor{red}{\textbf{+1.1097}}} & 0.5791 & \multirow{2}{*}{\textcolor{red}{\textbf{+0.0041}}} & 0.4500 & \multirow{2}{*}{-0.0027} & -0.8176 & \multirow{2}{*}{\textcolor{red}{\textbf{-0.0363}}} & 28.0\% & \multirow{2}{*}{\textcolor{red}{\textbf{+10.7\%}}} \\
& w/ & 5.0991 &  & 57.4882 &  & 0.5832 &  & 0.4473 &  & -0.8539 &  & 38.7\% &  \\
\midrule
\multirow{2}{*}{X-Restormer~\cite{X-Restormer}}
& w/o & 4.9505 & \multirow{2}{*}{+0.1197} & 57.2285 & \multirow{2}{*}{\textcolor{red}{\textbf{+0.7960}}} & 0.6030 & \multirow{2}{*}{-0.0202} & 0.4589 & \multirow{2}{*}{-0.0178} & -0.8625 & \multirow{2}{*}{\textcolor{red}{\textbf{-0.0176}}} & 43.3\% & \multirow{2}{*}{\textcolor{red}{\textbf{+8.0\%}}} \\
& w/ & 5.0702 &  & 58.0245 &  & 0.5828 &  & 0.4411 &  & -0.8801 &  & 51.3\% &  \\
\midrule
\multirow{2}{*}{PromptIR~\cite{PromptIR}}
& w/o & 4.9273 & \multirow{2}{*}{+0.1247} & 57.1754 & \multirow{2}{*}{\textcolor{red}{\textbf{+0.9759}}} & 0.6029 & \multirow{2}{*}{-0.0128} & 0.4556 & \multirow{2}{*}{-0.0147} & -0.8507 & \multirow{2}{*}{\textcolor{red}{\textbf{-0.0246}}} & 41.3\% & \multirow{2}{*}{\textcolor{red}{\textbf{+7.0\%}}} \\
& w/ & 5.0520 &  & 58.1513 &  & 0.5901 &  & 0.4409 &  & -0.8753 &  & 48.3\% &  \\
\midrule
\multirow{2}{*}{MoCE-IR~\cite{MoCE-IR}}
& w/o & 4.9853 & \multirow{2}{*}{+0.2161} & 57.9966 & \multirow{2}{*}{\textcolor{red}{\textbf{+0.3790}}} & 0.6053 & \multirow{2}{*}{-0.0216} & 0.4611 & \multirow{2}{*}{-0.0200} & -0.8622 & \multirow{2}{*}{\textcolor{red}{\textbf{-0.0223}}} & 39.0\% & \multirow{2}{*}{\textcolor{red}{\textbf{+6.7\%}}} \\
& w/ & 5.2014 &  & 58.3756 &  & 0.5837 &  & 0.4411 &  & -0.8845 &  & 45.7\% &  \\
\midrule
\multirow{2}{*}{DA-CLIP~\cite{DA-CLIP}}
& w/o & 4.7681 & \multirow{2}{*}{+0.0406} & 57.3662 & \multirow{2}{*}{\textcolor{red}{\textbf{+0.7357}}} & 0.6067 & \multirow{2}{*}{-0.0035} & 0.4637 & \multirow{2}{*}{-0.0069} & -0.8592 & \multirow{2}{*}{\textcolor{red}{\textbf{-0.0212}}} & 46.7\% & \multirow{2}{*}{\textcolor{red}{\textbf{+8.3\%}}} \\
& w/ & 4.8087 &  & 58.1019 &  & 0.6032 &  & 0.4568 &  & -0.8804 &  & 55.0\% &  \\
\midrule
\multirow{3}{*}{FoundIR~\cite{li2025foundir}}
& \textbf{official} & 5.1570 &  & 58.0616 &  & 0.6131 &  & 0.4556 &  & -0.8610 &  & 52.3\% &  \\
& w/o & 5.0551 & +0.4790 & 59.2164 & -1.1196 & 0.6143 & -0.0237 & 0.4803 & -0.0390 & -0.8837 & \textcolor{red}{\textbf{-0.0079}} & 48.7\% & \textcolor{red}{\textbf{+7.7\%}} \\
& w/ & 5.5341 &  & 58.0968 &  & 0.5906 &  & 0.4413 &  & -0.8916 &  & 56.3\% &  \\
\midrule
\multirow{2}{*}{FLUX-Controlnet~\cite{FLUX}}
& w/o & 4.4951 & \multirow{2}{*}{\textcolor{red}{\textbf{-0.4267}}} & 60.9375 & \multirow{2}{*}{\textcolor{red}{\textbf{+6.6225}}} & 0.6431 & \multirow{2}{*}{\textcolor{red}{\textbf{+0.0397}}} & 0.4795 & \multirow{2}{*}{\textcolor{red}{\textbf{+0.1053}}} & -0.8779 & \multirow{2}{*}{\textcolor{red}{\textbf{-0.2005}}} & 47.7\% & \multirow{2}{*}{\textcolor{red}{\textbf{+11.3\%}}} \\
& w/ & 4.0684 &  & 67.5600 &  & 0.6828 &  & 0.5848 &  & -1.0784 &  & 59.0\% &  \\
\midrule
\multirow{3}{*}{Qwen-Image-Edit~\cite{Qwen-Image-Edit}}
& \textbf{official} & 4.3440 &  & 67.6115 &  & 0.6647 &  & 0.6069 &  & -1.2154 &  & 56.7\% &  \\
& w/o & 4.6176 & \textcolor{red}{\textbf{-0.6793}} & 63.1324 & \textcolor{red}{\textbf{+7.2735}} & 0.6524 & \textcolor{red}{\textbf{+0.0383}} & 0.4945 & \textcolor{red}{\textbf{+0.1340}} & -1.0204 & \textcolor{red}{\textbf{-0.1825}} & 60.7\% & \textcolor{red}{\textbf{+10.0\%}} \\
& w/ & 3.9383 &  & 70.4059 &  & 0.6907 &  & 0.6285 &  & -1.2029 &  & 70.7\% &  \\
\bottomrule
\end{tabular}}

\end{table*}

\begin{figure*}[t]
    \centering
    \includegraphics[width=\linewidth]{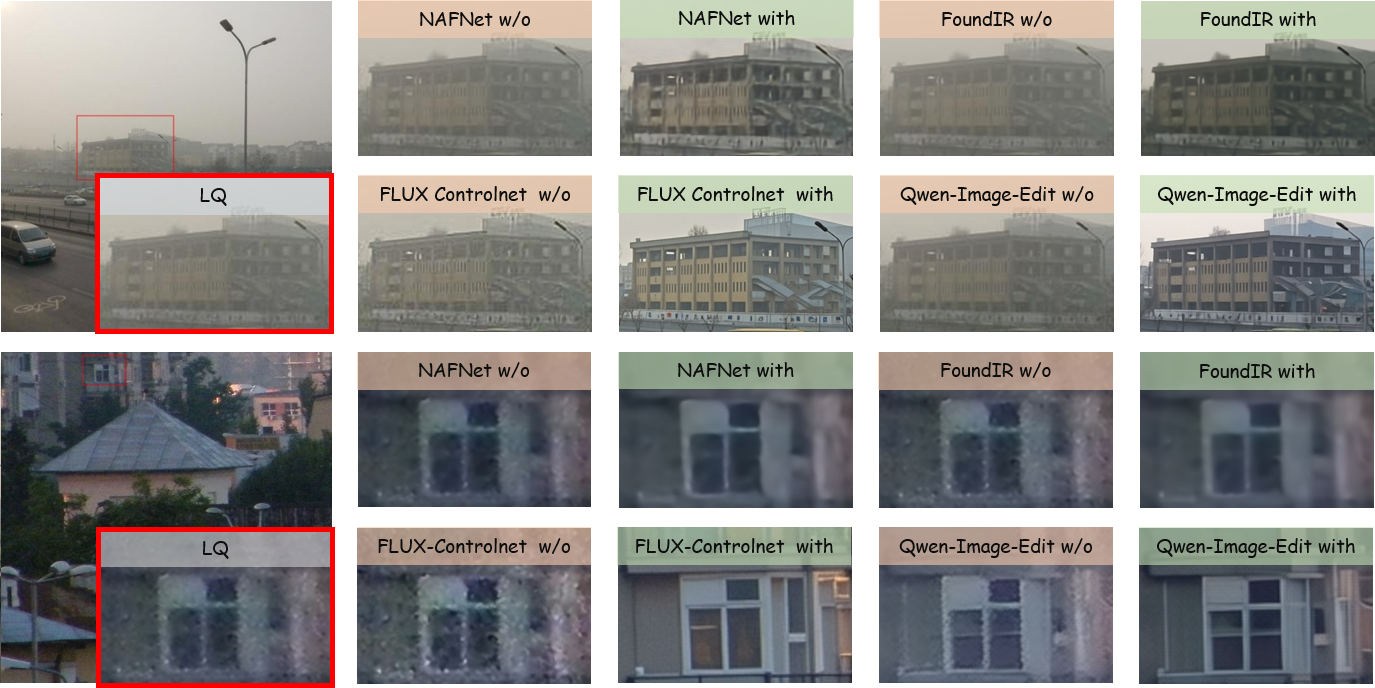}
    \caption{Qualitative comparison of restoration methods trained without and with GGT-100K. Training with GGT-100K yields clearer details and more faithful structures under real-world degradations.}
    \label{fig:visual_main}
\end{figure*}

\begin{figure*}[!t]
    \centering
    \includegraphics[width=1\linewidth]{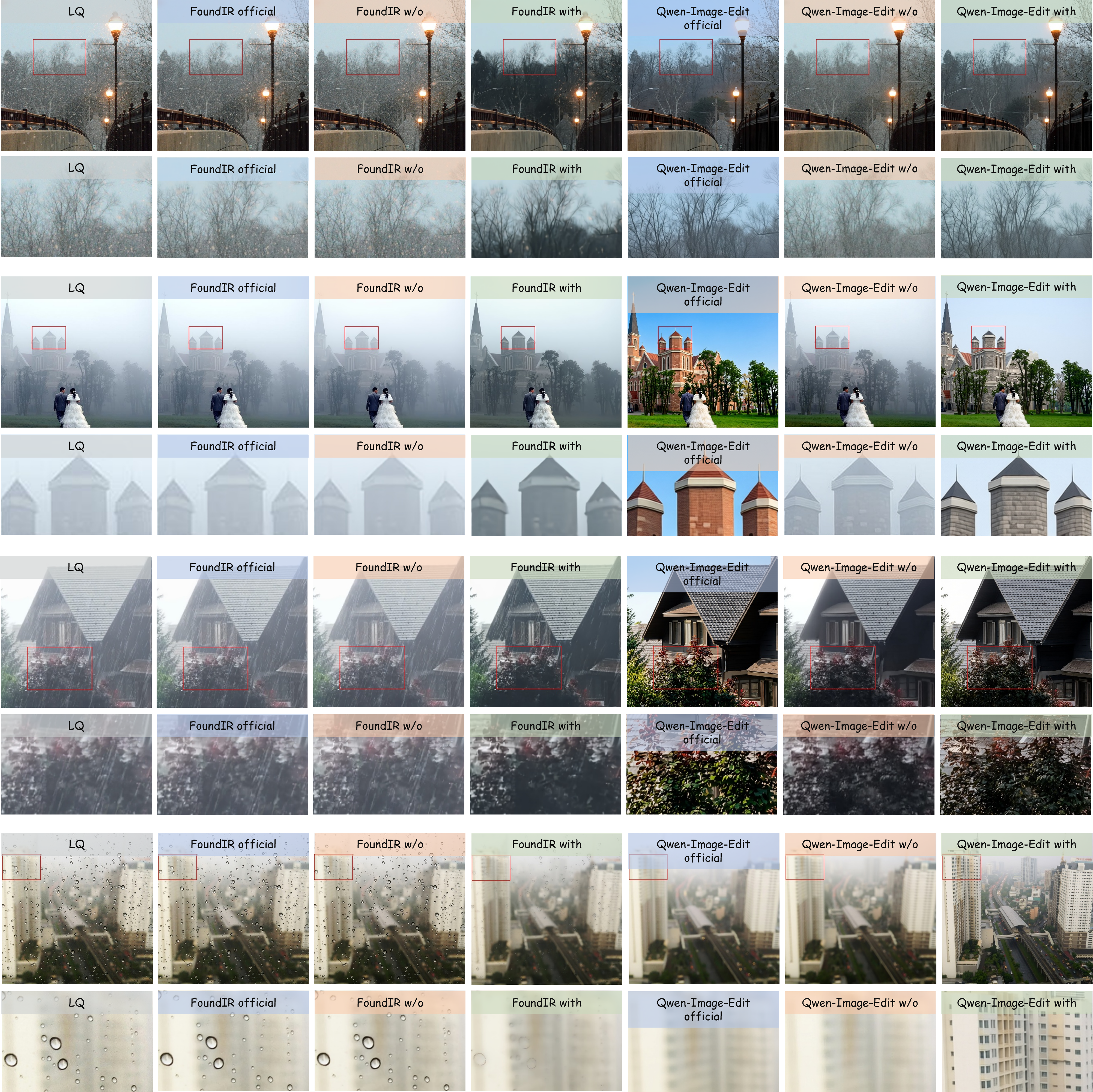}
    \caption{Qualitative comparison of FoundIR~\cite{li2025foundir} and Qwen-Image-Edit~\cite{Qwen-Image-Edit} on four real-world degradations: desnowing, dehazing, rain-streak removal, and rain-drop removal. We additionally report their \textbf{official} releases for reference. GGT-100K enables FoundIR to generalize to real-world degradations, and helps Qwen-Image-Edit achieve both strong generation ability and high content fidelity. Zoom in for better view.}
    \label{fig:app-withofficial}
\end{figure*}

\begin{figure*}[!t]
    \centering
    \includegraphics[width=0.94\linewidth]{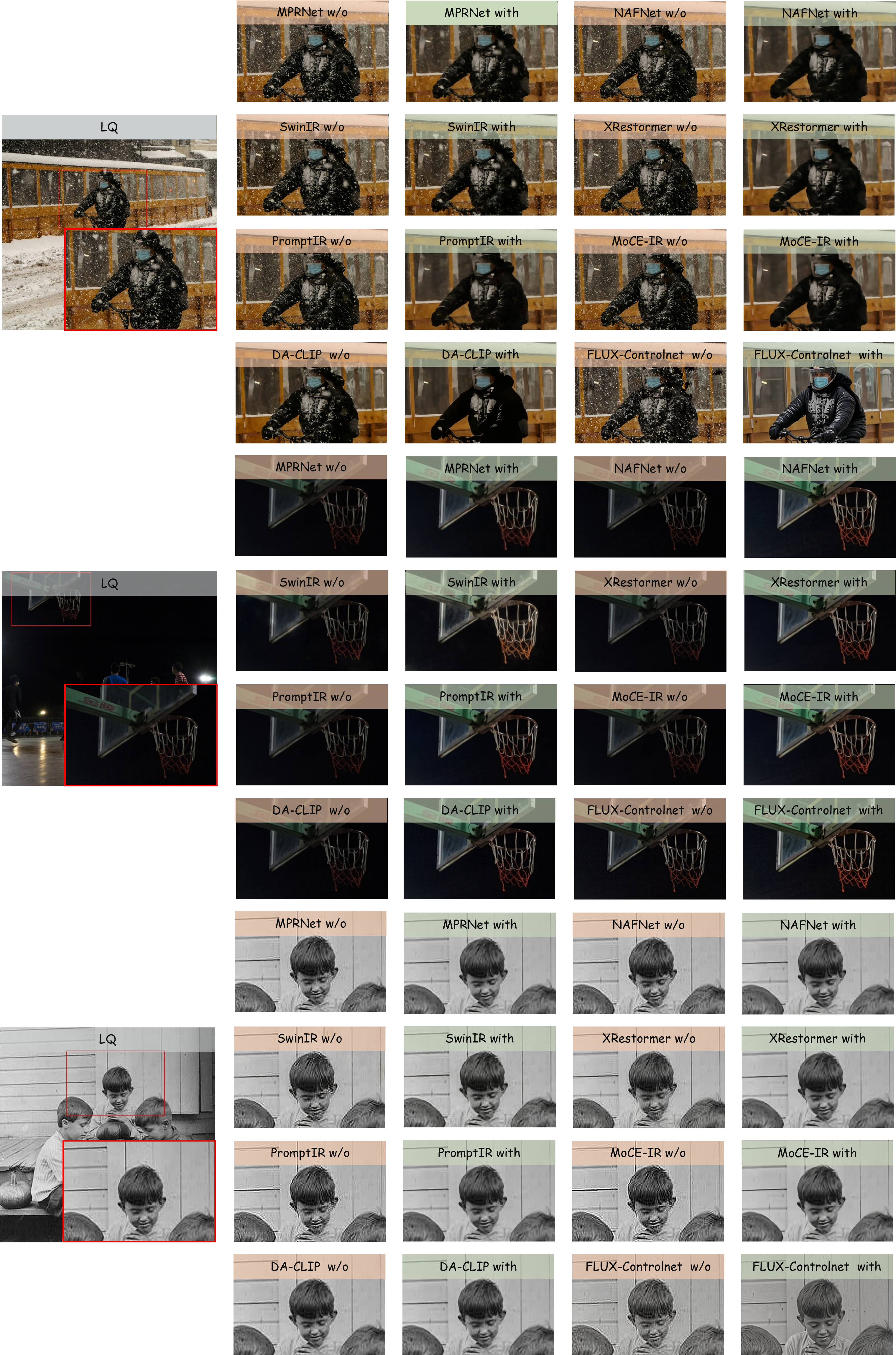}
    \caption{Qualitative comparison of eight representative restoration methods trained without (``w/o'') and with (``w/'') GGT-100K on three typical real-world degradations: desnowing, low-light enhancement, and old-photo restoration. Training with GGT-100K consistently strengthens their generalization to real-world degradations. Zoom in for better view.}
    \label{fig:app-woofficial}
\end{figure*}

\subsection{Experimental Results}
\label{sec:Experimental Results}

\vspace{2mm} \noindent\textbf{Results on GGT-100K Test set.}
As shown in Tab.~\ref{tab:main_ours}, incorporating GGT-100K generally benefits real-world restoration across diverse model families. Overall, training with GGT-100K improves all fidelity metrics, as well as MUSIQ, AFINE-NR and VLM-R, for all evaluated IR models. Notably, AFINE-NR is a more recent no-reference metric designed for the era of generative models, making it better aligned with human preference. Since many no-reference metrics are not well-suited for weather-related scenarios such as deraining and desnowing, VLM-R provides an important complementary indicator to this aspect. As shown in the table, adding GGT-100K in training consistently yields clear VLM-R gains across different models.

A closer look at different model types reveals some important findings. First, the fidelity gains are particularly large for fidelity-oriented methods. For example, X-Restormer~\cite{X-Restormer} and the all-in-one model PromptIR~\cite{PromptIR} both achieve PSNR gains of more than 3.5 dB, along with clear improvements in SSIM, LPIPS, and DISTS. This suggests that GGT-100K provides effective supervision for faithful reconstruction. In contrast, for generative models, such as FLUX-Controlnet~\cite{FLUX} and Qwen-Image-Edit~\cite{Qwen-Image-Edit}, GGT-100K substantially improves all metrics, with much larger gains than those of conventional restoration backbones on perceptual metrics. This indicates that GGT-100K is particularly suitable for modern generative models with strong priors and learning capacity.

The comparison with officially released real-world IR models further supports this conclusion. For FoundIR and Qwen-Image-Edit, training with GGT-100K brings broad improvements over their official versions. Qwen-Image-Edit is particularly representative: while its official version shows strong generative ability, it often suffers from limited fidelity and over-generation. Training only on existing synthetic and real paired data improves fidelity but degrades perceptual quality, whereas incorporating GGT-100K improves the both, yielding results that are both more faithful and visually more appealing, which can be clearly seen in the visual comparisons (See Fig.~\ref{fig:app-withofficial}). We also provide detailed results of each degradation category in \textbf{Appendix~\ref{app:h}}.

\vspace{2mm} \noindent\textbf{Results on Public RealLQ Test sets.} The average results on the existing RealLQ test sets are listed in Tab.~\ref{tab:public_reallq}, which shows a similar trend to Tab.~\ref{tab:main_ours}. Adding GGT-100K consistently improves AFINE-NR and VLM-R for all evaluated IR models, and MUSIQ for most models, with especially notable gains for generative IR models. Similar to the observations above, the finetuned pretrained models FLUX-Controlnet~\cite{FLUX} and Qwen-Image-Edit~\cite{Qwen-Image-Edit} also exhibit more visible improvements than the other baselines. The VLM-R improvements are smaller than that on GGT-100K test set, probably because these test sets contain samples with only mild or light degradations so that models trained without GGT-100K can also work. %做更多一些的分析

\vspace{2mm} \noindent\textbf{Visual Results.} Fig.~\ref{fig:visual_main} compares four representative methods covering distinct paradigms: pixel-wise $L_1$-loss-based NAFNet~\cite{chen2022simple}, train-from-scratch diffusion model FoundIR~\cite{li2025foundir}, pretrained T2I-based FLUX-Controlnet~\cite{FLUX}, and pretrained editing-model-based Qwen-Image-Edit~\cite{Qwen-Image-Edit}. In the first case with real haze, all four models trained only on existing datasets show limited generalization, leaving noticeable residual haze. After augmenting the training data with GGT-100K, all four models achieve clearly better haze removal. The slight color differences among the restored results are expected, since the true scene colors under real haze are inherently unknown and each method makes a reasonable estimation based on its own learned prior. Nonetheless, NAFNet and FoundIR, which lack strong generative priors, struggle to recover fine building details, whereas FLUX-Controlnet and Qwen-Image-Edit reconstruct sharper and more faithful textures. The second case with real noise exhibits a similar trend: without GGT-100K, all four models leave residual noise of unknown real-world distribution, and FLUX-Controlnet even erroneously amplifies the input noise; once GGT-100K is added to the training pool, their denoising capability is substantially improved.

{Fig.~\ref{fig:app-withofficial} compares FoundIR~\cite{li2025foundir} and Qwen-Image-Edit~\cite{Qwen-Image-Edit} on four real-world degradations, including desnowing, dehazing, rain-streak removal, and rain-drop removal, together with their official releases. For FoundIR, both the official model and the variant retrained on existing restoration datasets show limited generalization to real degradations, whereas incorporating GGT-100K enables it to effectively handle these cases. For Qwen-Image-Edit, its official release already exhibits decent restoration ability due to its strong generative prior, but the outputs tend to show oversaturated colors and limited fidelity. Finetuning on existing datasets improves fidelity but weakens degradation-removal capability, while finetuning with GGT-100K better preserves its generation ability and improves content fidelity, yielding the best balance among the three variants.}

{Fig.~\ref{fig:app-woofficial} further presents the results of more methods, including MPRNet~\cite{zamir2021multi}, NAFNet~\cite{chen2022simple}, SwinIR~\cite{liang2021swinir}, X-Restormer~\cite{X-Restormer}, PromptIR~\cite{PromptIR}, MoCE-IR~\cite{MoCE-IR}, DA-CLIP~\cite{DA-CLIP}, and FLUX-Controlnet~\cite{FLUX}, on real-world desnowing, low-light enhancement, and old-photo restoration. When trained only on existing datasets, most methods show limited generalization to real degradations. After adding GGT-100K in training, all eight methods achieve noticeably stronger restoration performance. Pixel-loss-based models such as MPRNet and SwinIR may still leave minor residual degradations, but their results are substantially improved over the corresponding baselines. In addition, although PromptIR and DA-CLIP benefit clearly from GGT-100K, their visual quality is still inferior to that of GGT-100K finetuned FLUX-Controlnet, whose strong generative prior enables sharper and more realistic reconstructions.}

\begin{table*}[t]
\centering
\caption{Comparison of representative restoration models under three settings: \textbf{baseline}, trained on the existing data setting only; \textbf{w/o-QC}, trained by directly adding the generated GGT data without quality control; and \textbf{w/-QC}, trained by adding the final GGT-100K after quality control. ``Improvement'' indicates the performance difference between \textbf{w/-QC} and \textbf{w/o-QC}, thus reflecting the gain brought by multi-stage quality control. The baseline results are shown in \textcolor{gray}{gray} for reference, and the positive improvements are highlighted in \textcolor{red}{\textbf{red}}.}
\vspace{1mm}
\label{tab:abl_qc}
\resizebox{\textwidth}{!}{
\begin{tabular}{c|c|cccc|ccccc|c}
\toprule
\multirow{2}{*}{Model} & \multirow{2}{*}{Quality control} & \multicolumn{4}{c|}{Full-reference fidelity metrics} & \multicolumn{5}{c|}{No-reference perceptual metrics} & \multirow{2}{*}{VLM-R $\uparrow$} \\
\cmidrule(lr){3-6}\cmidrule(lr){7-11}
& & PSNR $\uparrow$ & SSIM $\uparrow$ & LPIPS $\downarrow$ & DISTS $\downarrow$ & NIQE $\downarrow$ & MUSIQ $\uparrow$ & MANIQA $\uparrow$ & TOPIQ $\uparrow$ & AFINE-NR $\downarrow$ &  \\
\midrule
\multirow{4}{*}{MPRNet~\cite{zamir2021multi}}
& \textcolor{gray}{baseline} & \textcolor{gray}{24.7919} & \textcolor{gray}{0.7637} & \textcolor{gray}{0.3779} & \textcolor{gray}{0.2390} & \textcolor{gray}{6.9774} & \textcolor{gray}{42.1039} & \textcolor{gray}{0.4641} & \textcolor{gray}{0.3120} & \textcolor{gray}{-0.7140} & \textcolor{gray}{22.2\%} \\
& w/o-QC & 26.1638 & 0.8140 & 0.3422 & 0.2190 & 6.8109 & 44.1973 & 0.4369 & 0.3043 & -0.7540 & 33.8\% \\
& w/-QC & 27.3044 & 0.8189 & 0.3377 & 0.2212 & 7.0035 & 44.9597 & 0.4431 & 0.3075 & -0.7794 & 33.2\% \\
& Improvement & \textcolor{red}{\textbf{+1.1406}} & \textcolor{red}{\textbf{+0.0049}} & \textcolor{red}{\textbf{-0.0045}} & +0.0022 & +0.1926 & \textcolor{red}{\textbf{+0.7624}} & \textcolor{red}{\textbf{+0.0062}} & \textcolor{red}{\textbf{+0.0032}} & \textcolor{red}{\textbf{-0.0254}} & -0.6\% \\
\midrule
\multirow{4}{*}{NAFNet~\cite{chen2022simple}}
& \textcolor{gray}{baseline} & \textcolor{gray}{25.1255} & \textcolor{gray}{0.7708} & \textcolor{gray}{0.3653} & \textcolor{gray}{0.2298} & \textcolor{gray}{6.6654} & \textcolor{gray}{42.0124} & \textcolor{gray}{0.4635} & \textcolor{gray}{0.3118} & \textcolor{gray}{-0.7012} & \textcolor{gray}{27.6\%} \\
& w/o-QC & 27.9172 & 0.8312 & 0.3129 & 0.2071 & 6.7970 & 46.3143 & 0.4387 & 0.3090 & -0.7833 & 53.4\% \\
& w/-QC & 28.2461 & 0.8349 & 0.3110 & 0.2043 & 6.7983 & 46.7094 & 0.4330 & 0.3097 & -0.7881 & 53.8\% \\
& Improvement & \textcolor{red}{\textbf{+0.3289}} & \textcolor{red}{\textbf{+0.0037}} & \textcolor{red}{\textbf{-0.0019}} & \textcolor{red}{\textbf{-0.0028}} & +0.0013 & \textcolor{red}{\textbf{+0.3951}} & -0.0057 & \textcolor{red}{\textbf{+0.0007}} & \textcolor{red}{\textbf{-0.0048}} & \textcolor{red}{\textbf{+0.4\%}} \\
\midrule
\multirow{4}{*}{SwinIR~\cite{liang2021swinir}}
& \textcolor{gray}{baseline} & \textcolor{gray}{23.9225} & \textcolor{gray}{0.7590} & \textcolor{gray}{0.3878} & \textcolor{gray}{0.2343} & \textcolor{gray}{6.4979} & \textcolor{gray}{41.0176} & \textcolor{gray}{0.4488} & \textcolor{gray}{0.3082} & \textcolor{gray}{-0.6940} & \textcolor{gray}{18.6\%} \\
& w/o-QC & 26.6511 & 0.8055 & 0.3420 & 0.2160 & 6.6699 & 42.3312 & 0.4411 & 0.2956 & -0.7262 & 32.2\% \\
& w/-QC & 27.0781 & 0.8150 & 0.3369 & 0.2131 & 6.7366 & 43.6689 & 0.4308 & 0.2966 & -0.7437 & 37.6\% \\
& Improvement & \textcolor{red}{\textbf{+0.4270}} & \textcolor{red}{\textbf{+0.0095}} & \textcolor{red}{\textbf{-0.0051}} & \textcolor{red}{\textbf{-0.0029}} & +0.0667 & \textcolor{red}{\textbf{+1.3377}} & -0.0103 & \textcolor{red}{\textbf{+0.0010}} & \textcolor{red}{\textbf{-0.0175}} & \textcolor{red}{\textbf{+5.4\%}} \\
\midrule
\multirow{4}{*}{X-Restormer~\cite{X-Restormer}}
& \textcolor{gray}{baseline} & \textcolor{gray}{24.6901} & \textcolor{gray}{0.7705} & \textcolor{gray}{0.3587} & \textcolor{gray}{0.2234} & \textcolor{gray}{6.6564} & \textcolor{gray}{43.1437} & \textcolor{gray}{0.4643} & \textcolor{gray}{0.3192} & \textcolor{gray}{-0.7185} & \textcolor{gray}{30.4\%} \\
& w/o-QC & 28.0860 & 0.8348 & 0.3140 & 0.2090 & 6.9132 & 47.2496 & 0.4355 & 0.3153 & -0.8124 & 52.6\% \\
& w/-QC & 28.2298 & 0.8362 & 0.3130 & 0.2065 & 6.9218 & 46.8069 & 0.4295 & 0.3109 & -0.8126 & 54.6\% \\
& Improvement & \textcolor{red}{\textbf{+0.1438}} & \textcolor{red}{\textbf{+0.0014}} & \textcolor{red}{\textbf{-0.0010}} & \textcolor{red}{\textbf{-0.0025}} & +0.0086 & -0.4427 & -0.0060 & -0.0044 & \textcolor{red}{\textbf{-0.0002}} & \textcolor{red}{\textbf{+2.0\%}} \\
\midrule
\multirow{4}{*}{PromptIR~\cite{PromptIR}}
& \textcolor{gray}{baseline} & \textcolor{gray}{24.5733} & \textcolor{gray}{0.7564} & \textcolor{gray}{0.3630} & \textcolor{gray}{0.2312} & \textcolor{gray}{6.5942} & \textcolor{gray}{43.2318} & \textcolor{gray}{0.4695} & \textcolor{gray}{0.3202} & \textcolor{gray}{-0.7176} & \textcolor{gray}{24.8\%} \\
& w/o-QC & 27.9471 & 0.8324 & 0.3147 & 0.2094 & 6.8882 & 46.5024 & 0.4412 & 0.3127 & -0.7987 & 50.6\% \\
& w/-QC & 28.1775 & 0.8344 & 0.3113 & 0.2051 & 6.8029 & 46.7198 & 0.4369 & 0.3124 & -0.7968 & 49.6\% \\
& Improvement & \textcolor{red}{\textbf{+0.2304}} & \textcolor{red}{\textbf{+0.0020}} & \textcolor{red}{\textbf{-0.0034}} & \textcolor{red}{\textbf{-0.0043}} & \textcolor{red}{\textbf{-0.0853}} & \textcolor{red}{\textbf{+0.2174}} & -0.0043 & -0.0003 & +0.0019 & -1.0\% \\
\midrule
\multirow{4}{*}{MoCE-IR~\cite{MoCE-IR}}
& \textcolor{gray}{baseline} & \textcolor{gray}{24.8641} & \textcolor{gray}{0.7575} & \textcolor{gray}{0.3654} & \textcolor{gray}{0.2291} & \textcolor{gray}{6.6248} & \textcolor{gray}{43.0279} & \textcolor{gray}{0.4671} & \textcolor{gray}{0.3200} & \textcolor{gray}{-0.7116} & \textcolor{gray}{25.4\%} \\
& w/o-QC & 28.0974 & 0.8359 & 0.3140 & 0.2118 & 6.9658 & 47.9244 & 0.4368 & 0.3212 & -0.8171 & 53.4\% \\
& w/-QC & 28.2140 & 0.8392 & 0.3097 & 0.2093 & 6.9402 & 48.4846 & 0.4344 & 0.3242 & -0.8270 & 55.2\% \\
& Improvement & \textcolor{red}{\textbf{+0.1166}} & \textcolor{red}{\textbf{+0.0033}} & \textcolor{red}{\textbf{-0.0043}} & \textcolor{red}{\textbf{-0.0025}} & \textcolor{red}{\textbf{-0.0256}} & \textcolor{red}{\textbf{+0.5602}} & -0.0024 & \textcolor{red}{\textbf{+0.0030}} & \textcolor{red}{\textbf{-0.0099}} & \textcolor{red}{\textbf{+1.8\%}} \\
\midrule
\multirow{4}{*}{DA-CLIP~\cite{DA-CLIP}}
& \textcolor{gray}{baseline} & \textcolor{gray}{25.8809} & \textcolor{gray}{0.7726} & \textcolor{gray}{0.3427} & \textcolor{gray}{0.2180} & \textcolor{gray}{6.3011} & \textcolor{gray}{39.3370} & \textcolor{gray}{0.4846} & \textcolor{gray}{0.3080} & \textcolor{gray}{-0.6853} & \textcolor{gray}{31.6\%} \\
& w/o-QC & 26.6978 & 0.7920 & 0.2903 & 0.1923 & 6.3872 & 42.3654 & 0.4805 & 0.3090 & -0.7367 & 48.2\% \\
& w/-QC & 26.6914 & 0.7910 & 0.2954 & 0.1938 & 6.3039 & 41.8839 & 0.4834 & 0.3073 & -0.7277 & 51.0\% \\
& Improvement & -0.0064 & -0.0010 & +0.0051 & +0.0015 & \textcolor{red}{\textbf{-0.0833}} & -0.4815 & \textcolor{red}{\textbf{+0.0029}} & -0.0017 & +0.0090 & \textcolor{red}{\textbf{+2.8\%}} \\
\midrule
\multirow{4}{*}{FoundIR~\cite{li2025foundir}}
& \textcolor{gray}{baseline} & \textcolor{gray}{25.8048} & \textcolor{gray}{0.7844} & \textcolor{gray}{0.3508} & \textcolor{gray}{0.2220} & \textcolor{gray}{6.9023} & \textcolor{gray}{42.5388} & \textcolor{gray}{0.4701} & \textcolor{gray}{0.3153} & \textcolor{gray}{-0.7365} & \textcolor{gray}{35.8\%} \\
& w/o-QC & 26.0990 & 0.8068 & 0.3403 & 0.2284 & 7.3712 & 44.7388 & 0.4552 & 0.3123 & -0.8147 & 50.2\% \\
& w/-QC & 27.1777 & 0.8213 & 0.3351 & 0.2209 & 7.4717 & 43.9238 & 0.4402 & 0.3019 & -0.8087 & 60.8\% \\
& Improvement & \textcolor{red}{\textbf{+1.0787}} & \textcolor{red}{\textbf{+0.0145}} & \textcolor{red}{\textbf{-0.0052}} & \textcolor{red}{\textbf{-0.0075}} & +0.1005 & -0.8150 & -0.0150 & -0.0104 & \textcolor{red}{\textbf{+0.0060}} & \textcolor{red}{\textbf{+10.6\%}} \\
\midrule
\multirow{4}{*}{FLUX-Controlnet~\cite{FLUX}}
& \textcolor{gray}{baseline} & \textcolor{gray}{22.4486} & \textcolor{gray}{0.6901} & \textcolor{gray}{0.3773} & \textcolor{gray}{0.2129} & \textcolor{gray}{5.5636} & \textcolor{gray}{48.5454} & \textcolor{gray}{0.5157} & \textcolor{gray}{0.3798} & \textcolor{gray}{-0.6912} & \textcolor{gray}{25.4\%} \\
& w/o-QC & 19.4203 & 0.6613 & 0.3619 & 0.1979 & 4.7290 & 66.1931 & 0.5988 & 0.5519 & -0.9639 & 45.8\% \\
& w/-QC & 23.1413 & 0.7325 & 0.2625 & 0.1520 & 4.8504 & 63.0910 & 0.5854 & 0.5013 & -0.9280 & 63.4\% \\
& Improvement & \textcolor{red}{\textbf{+3.7210}} & \textcolor{red}{\textbf{+0.0712}} & \textcolor{red}{\textbf{-0.0994}} & \textcolor{red}{\textbf{-0.0459}} & +0.1214 & -3.1021 & -0.0134 & -0.0506 & +0.0359 & \textcolor{red}{\textbf{+17.6\%}} \\
\midrule
\multirow{4}{*}{Qwen-Image-Edit~\cite{Qwen-Image-Edit}}
& \textcolor{gray}{baseline} & \textcolor{gray}{25.8559} & \textcolor{gray}{0.7787} & \textcolor{gray}{0.2813} & \textcolor{gray}{0.1625} & \textcolor{gray}{6.0198} & \textcolor{gray}{51.4215} & \textcolor{gray}{0.5333} & \textcolor{gray}{0.3765} & \textcolor{gray}{-0.8401} & \textcolor{gray}{77.4\%} \\
& w/o-QC & 25.1892 & 0.7771 & 0.2220 & 0.1231 & 5.5180 & 62.1015 & 0.5768 & 0.4654 & -0.9588 & 85.2\% \\
& w/-QC & 26.1811 & 0.7828 & 0.2155 & 0.1183 & 5.4648 & 62.5519 & 0.5811 & 0.4707 & -0.9611 & 87.6\% \\
& Improvement & \textcolor{red}{\textbf{+0.9919}} & \textcolor{red}{\textbf{+0.0057}} & \textcolor{red}{\textbf{-0.0065}} & \textcolor{red}{\textbf{-0.0048}} & \textcolor{red}{\textbf{-0.0532}} & \textcolor{red}{\textbf{+0.4504}} & \textcolor{red}{\textbf{+0.0043}} & \textcolor{red}{\textbf{+0.0053}} & \textcolor{red}{\textbf{-0.0023}} & \textcolor{red}{\textbf{+2.4\%}} \\
\bottomrule
\end{tabular}}
\end{table*}

\subsection{Ablation Study of Multi-Stage Quality Control}

{We further conduct an ablation study to evaluate the effect of our multi-stage quality control. As shown in Tab.~\ref{tab:abl_qc}, we compare three settings: \textbf{baseline} (i.e., trained on existing data only); \textbf{w/o-QC}, trained by directly adding the generated Nano-Banana-2 data without further screening; and \textbf{w/-QC}, trained by adding the final GGT-100K after quality control.}

{We see that even the unscreened generated data (\textbf{w/o-QC}) already improve many models over the baseline. This is expected, since these samples are produced by the best-performing model-prompt combination selected through our systematic evaluation, and are therefore of reasonably good quality. However, these gains are not uniformly stable. For some models, the improvements over baseline are limited (e.g., FoundIR), and for a few cases, certain metrics even degrade (e.g., FLUX-Controlnet). This suggests that the generated data still contain imperfect samples, such as hallucinated details or semantic inconsistencies, which can weaken supervision.}

{Multi-stage quality control therefore plays an important role. As shown in Tab.~\ref{tab:abl_qc}, the final GGT-100K further improves fidelity-oriented metrics over the \textbf{w/o-QC} setting for most models, especially PSNR, SSIM, LPIPS, and DISTS, indicating that quality control makes the supervision more faithful and reliable. This effect is particularly clear for FLUX-Controlnet~\cite{FLUX}. Without quality control, the generated data even hurt PSNR and SSIM relative to the baseline, despite producing strong perceptual scores. After quality control, however, FLUX-Controlnet shows large gains in both fidelity and VLM-R. We attribute this to its strong learning capacity and the relative instability of ControlNet-style finetuning for IR tasks, which makes it more sensitive to hallucinated or inconsistent training pairs.}

%Overall, these results verify that multi-stage quality control is a key component of the GGT-100K pipeline, especially for improving the reliability of supervision for modern generative restoration models.

% \vspace{2mm} \noindent\textbf{Limitations and Discussion}.

\section{Limitation and Discussion}

First, GGT-100K should be regarded as a scalable and high-quality approximation of physically captured ground truth, rather than a perfect substitute for real captured references. Despite the multi-stage quality control, some images in GGT-100K can still contain subtle imperfections, such as minor artifacts or hallucinated details introduced by MFMs. This limitation is difficult to completely avoid when using generative models to synthesize restoration targets. Nevertheless, compared with synthetic degradation pipelines or costly real-world paired acquisition, our approach offers a much more practical way to build diverse real-world LQ-HQ pairs, and our experiments show that such data are already highly effective for improving model generalization.

Second, although GGT-100K broadens the generalization boundary of current restoration models, it cannot cover the full space of real-world degradations. Real-world image degradation is highly diverse, mixed, and open-ended, so models trained with GGT-100K may still fail on some previously unseen degradation types that are not sufficiently represented in the dataset. However, we believe that the proposed pipeline has strong scalability: with more diverse source-image collection and more advanced MFMs, it can gradually cover a much broader range of real-world degradations.

Third, in our experiments, we adopt those widely used model architectures and general training strategies to verify the data value of GGT-100K, rather than optimizing specific models for this dataset. Therefore, the improvements reported in this paper should be viewed as a conservative estimate of the potential of GGT-based supervision. Different model families may benefit from more specialized network designs, training objectives, and finetuning strategies tailored to GGT-100K. This also points to an important future direction: exploring more effective restoration algorithms under the training environment provided by GGT-100K.

\section{Conclusion}

We proposed GGT, a practical paradigm for constructing real-world paired training data for image restoration using generative MFMs. Based on a systematic evaluation of nine MFMs and multiple prompting strategies, we built GGT-100K, a large-scale real-world paired dataset with 103K training pairs and a curated test set of 500 pairs. Extensive experiments showed that GGT-100K consistently improved the generalization performance, in terms of both content fidelity and perceptual quality, of various restoration models, particularly those finetuned generative models, whose strong priors and high learning capacity enable them to better exploit the advantages of GGT-100K. These results validated the use of MFMs for restoration-oriented data generation and demonstrated the value of GGT-100K as a useful resource for advancing generalizable real-world image restoration.

% \textbf{Limitations}. GGT-100K should be regarded as a scalable and high-quality approximation of physically captured ground truth. Despite the multi-stage quality control, some images in GGT-100K can still have subtle visual artifacts hallucinated by MFMs. Nevertheless, compared with synthetic degradation pipelines or costly real-world paired acquisition, our approach offers a more practical way to build diverse real-world LQ-HQ pairs. %As a new complementary source to existing data, it can already help extend the generalization boundary of current restoration models. 

%==========================================================================
% Acknowledgements
%==========================================================================
% \section*{\textcolor{polyured}{$\blacksquare$}\,Acknowledgements}
% This work was supported by the Visual Computing Lab at The Hong Kong
% Polytechnic University and our industrial partners. We thank our
% colleagues for valuable discussions.

%==========================================================================
% Bibliography
%==========================================================================
% \begin{thebibliography}{99}
% \bibitem{polyu-vclab-website}
%   PolyU Visual Computing Lab.
%   \newblock \emph{Visual Computing Lab Website}.
%   \newblock \url{https://www4.comp.polyu.edu.hk/~cslzhang/}, 2026.
% \end{thebibliography}

{\small
\bibliography{ref}
}

\appendix

\clearpage
\setcounter{table}{0}
\setcounter{equation}{0}
\setcounter{figure}{0}
\renewcommand{\thetable}{\thesection.\arabic{table}}
\renewcommand{\theequation}{\thesection.\arabic{equation}}
\renewcommand{\thefigure}{\thesection.\arabic{figure}}
\begin{center}
    {\LARGE\bfseries Appendix}
\end{center}

In this appendix, we provide the following materials:
\begin{itemize}
    \item \textbf{A.} More details of source image collection for GGT-100K (referring to Sec.~\ref{sec:Source Image Collection}).
    \item \textbf{B.} Detailed prompt designs for MFM evaluation (referring to Sec.~\ref{sec:Systematic MFMs Evaluation}).
    \item \textbf{C.} The usage of VLM as evaluator and quality controller (referring to Sec.~\ref{sec:Systematic MFMs Evaluation} and Sec.~\ref{sec:Multi-stage Quality Control}).
    \item \textbf{D.} Details of the user study (referring to Sec.~\ref{sec:Systematic MFMs Evaluation}).
    \item \textbf{E.} Details of training data composition (referring to Sec.~\ref{sec:Experimental Settings}).
    %\item \textbf{F.} Additional qualitative comparisons and visual results (referring to Sec.~\ref{sec:Experimental Results}).
    %\item \textbf{G.} Ablation study of the multi-stage quality control pipeline (referring to Sec.~\ref{sec:Experimental Results}).
    \item \textbf{F.} Detailed results on specific degradation categories (referring to Sec.~\ref{sec:Experimental Results}).
\end{itemize}

\section{Details of Source Image Collection}
\label{app:a}

In this section, we provide more details on the source image collection process for GGT-100K. We collect real-world LQ images from three main sources: Internet sources, public datasets, and our own captures. The collected images are organized into six categories: \textit{General Mixed}, \textit{Low-Light}, \textit{Haze}, \textit{Rain}, \textit{Snow}, and \textit{Old Photo}. Because the raw images from these sources vary substantially in content, degradation type, and perceptual quality, we further apply source-specific filtering, cropping, and quality-control procedures before constructing the final dataset.

For web-collected images, we adopt category-specific collection and filtering strategies for different degradation categories, and all collected images from the Internet are restricted to those released under CC0 licenses. For \textit{Rain} and \textit{Snow}, we search online image platforms~\cite{unsplash, pexels, pixabay, flickr} using keywords associated with visible rain and snow phenomena (e.g., rain, rain streak, heavy rain, snow, and snowstorm). After removing duplicate results, we manually inspect the full-resolution images and discard those without clearly visible artifacts, and then resize and crop the remaining images into patches, yielding over 30K candidate patches before filtering. We further apply VLM-based filtering using Gemini to remove patches in which the target artifacts are not sufficiently apparent, resulting in more than 10K retained patches from the \textit{Rain} and \textit{Snow} categories combined. For \textit{Old Photo}, we retrieve over 10K historical photographs under CC0 licenses shared by Flickr~\cite{flickr} contributors who specifically curate old-photo content. For \textit{General Mixed}, we collect over 20K CC0-licensed user-uploaded images from online image platforms~\cite{unsplash, pexels, pixabay, flickr}. To preserve the original noise characteristics of \textit{Old Photo} and \textit{General Mixed} types, we do not resize these images and instead crop them directly into patches, producing over 100K candidate patches. We then use Gemini to remove content-poor patches (e.g., regions with little structural information) and overly degraded patches that have already lost most useful content. Finally, we divide the remaining patches into high-, medium-, and low-quality levels based on subjective criteria, and sample them with a ratio of 1:2:3 to increase the proportion of moderately and severely degraded samples.

For dataset-sourced images, we first gather over 10K images from existing real-world restoration datasets that do not provide ground-truth references~\cite{RESIDE, snow100k, chen2018robust}, as well as over 1,000K images from other image datasets containing low-quality or bad-weather conditions~\cite{deng2009imagenet, Exdark, yang2020advancing, SDV21}. We then retain only images with clear and suitable degradations through metric-based and VLM-based screening and subjective filtering. In addition, we include a self-captured subset, where we manually adjust camera and smartphone parameters such as exposure time and ISO to capture low-light and noisy images, and also vary the focus or introduce physical camera shake during capture to obtain realistic blurry images. These images are further cropped into over 20K patches and filtered using VLMs to keep only suitable samples.

Overall, we screen more than 1,100K image patches through the above procedures and retain about 120K low-quality images. We then use Nano-Banana-2 to generate corresponding high-quality counterparts and further filter out image pairs through the multi-stage quality control described in Sec.~\ref{sec:Multi-stage Quality Control}. Owing to the strong generation quality of Nano-Banana-2, we retain 103,707 image pairs after this multi-stage quality control process. Among them, \textit{General Mixed} contains 66,058 pairs, \textit{Low-Light} contains 9,786 pairs, \textit{Haze} contains 7,822 pairs, \textit{Rain} contains 7,177 pairs, \textit{Snow} contains 6,759 pairs, and \textit{Old Photo} contains 6,105 pairs.

\vspace{2mm} \noindent\textbf{General Mixed.}
The \textit{General Mixed} category contains 66,058 pairs collected from three sources: Internet images, public datasets, and our own captures. Specifically, it includes 37,463 pairs from Internet images mainly collected from Flickr~\cite{flickr}, 19,438 pairs from public dataset images mainly from ImageNet~\cite{deng2009imagenet}, where we select images with relatively poor visual quality, and 9,157 pairs from our own captures.

\vspace{2mm} \noindent\textbf{Rain.}
The \textit{Rain} category contains 7,177 pairs collected from both public datasets and Internet sources, including 40 pairs from NTURain~\cite{chen2018robust}, 2,222 pairs from OpenReal80K~\cite{jarvisir2025}, 1,590 pairs from Flickr~\cite{flickr}, 1,806 pairs from Unsplash~\cite{unsplash}, and 1,517 pairs from Pexels~\cite{pexels}.

\vspace{2mm} \noindent\textbf{Haze.}
The \textit{Haze} category contains 7,822 pairs, all collected from public datasets, including 2,000 pairs from ACDC~\cite{SDV21} and 5,822 pairs from RTTS~\cite{RESIDE}.

\vspace{2mm} \noindent\textbf{Snow.}
The \textit{Snow} category contains 6,759 pairs, including 5,202 pairs from Unsplash~\cite{unsplash} and 1,557 pairs from Snow100K-Real~\cite{snow100k}.

\vspace{2mm} \noindent\textbf{Low-Light.}
The \textit{Low-Light} category contains 9,786 pairs, including 5,702 pairs from DarkFace~\cite{yang2020advancing}, 2,858 pairs from ExDark~\cite{Exdark}, and 1,226 pairs from our own captures.

\vspace{2mm} \noindent\textbf{Old Photo.}
The \textit{Old Photo} category contains 6,105 pairs, all collected from Flickr~\cite{flickr} from different providers, ensuring source diversity within this category.

\section{Detailed Prompt Designs for MFMs Evaluation}
\label{app:b}

\begin{table}[t]
\centering
\caption{Fixed prompt and fixed-prompt-no-change used for different degradation types.}
\label{tab:fixed_prompts}
\small
\begin{adjustbox}{width=0.89\linewidth}
\begin{tabular}{p{2.8cm} p{4.5cm} p{5.2cm}}
\toprule
\textbf{Degradation Type} & \textbf{Fixed-Prompt} & \textbf{Fixed-Prompt-No-Change} \\
\midrule
GGT General Mixed & Restore this low-quality image to a clear, clean and natural state. & Restore this low-quality image to a clear, clean and natural state. But do not change the image content. \\
\midrule
GGT Rain & Please remove the rain from the image and restore its clarity. & Please remove the rain from the image and restore its clarity. But do not change the image content. \\
\midrule
GGT Haze & Please dehaze the image. & Please dehaze the image. But do not change the image content. \\
\midrule
GGT Snow & Please remove the snow from the image and restore its clarity. & Please remove the snow from the image and restore its clarity. But do not change the image content. \\
\midrule
GGT Low-Light & Please restore this low-quality image, recovering its normal brightness and clarity. & Please restore this low-quality image, recovering its normal brightness and clarity. But do not change the image content. \\
\midrule
GGT Old Photo & Please restore the old photo to a clear, clean and natural state. & Please restore the old photo to a clear, clean and natural state. But do not change the image content. \\
\bottomrule
\end{tabular}
\end{adjustbox}
\end{table}

\begin{figure*}[hp]
    \centering
    \includegraphics[width=\linewidth]{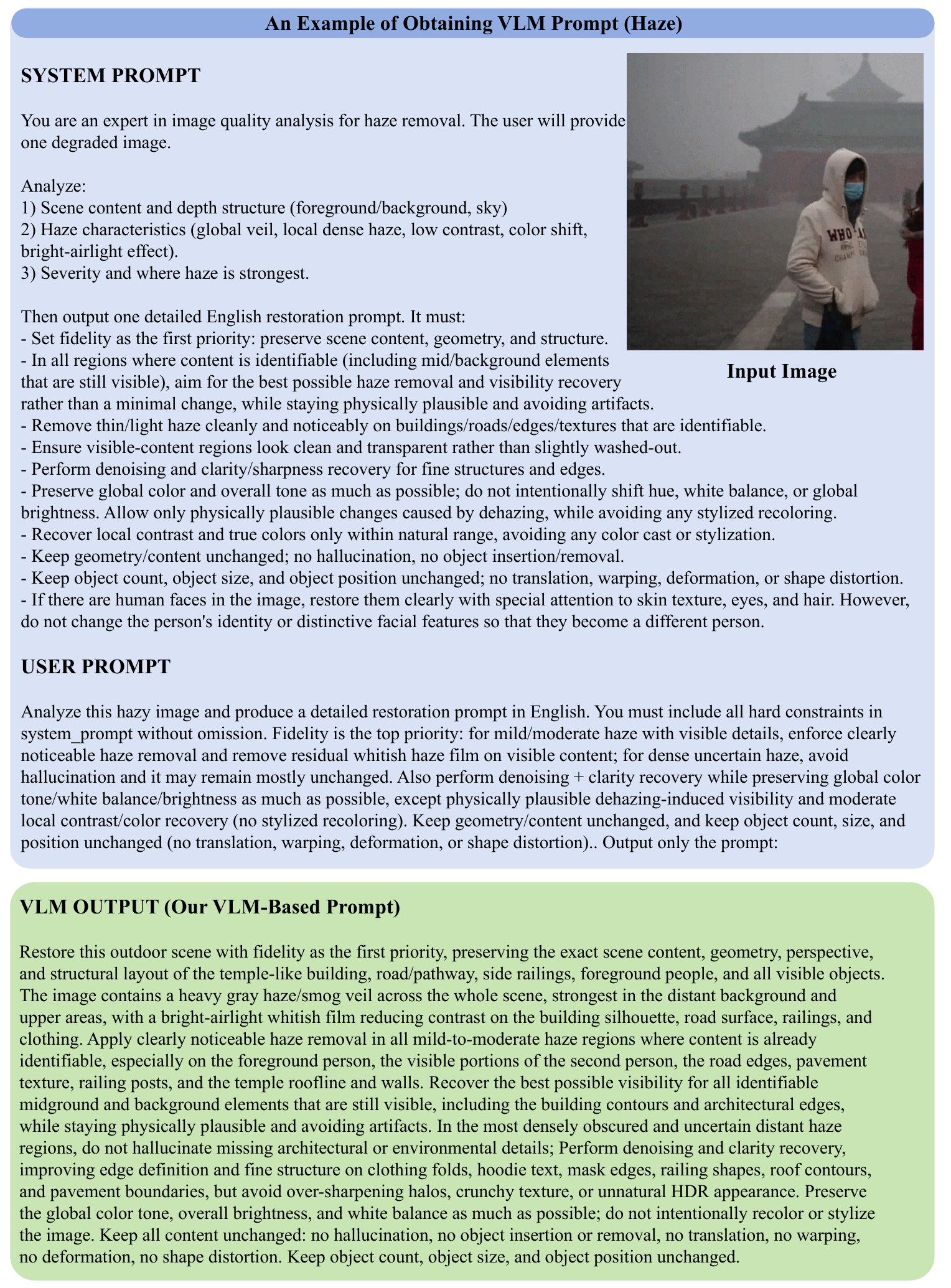}
    \caption{A processing example of VLM prompts. Given an input image and designed meta prompt, the VLM produces a detailed, image-specific restoration instruction that describes both the image content and the degradations to be removed.}
    \label{fig:prompt}
\end{figure*}

In Sec.~\ref{sec:Systematic MFMs Evaluation} and Tab.~\ref{tab:model_prompt_excl_realesrgan} of the main paper, we evaluate 9 candidate MFMs under 4 prompting strategies. Here we provide the detailed designs of these prompts.

As shown in Tab.~\ref{tab:fixed_prompts}, similar to~\cite{yang2026realrestorer}, our fixed prompt and fixed-prompt-no-change adopt the simplest instruction form. A simple ``do not change the image content'' instruction consistently improves content fidelity for most MFMs. 

We show an example of a haze image to illustrate the generation of our VLM-based prompt in Fig.~\ref{fig:prompt}. An input image is first fed into a VLM (GPT-5.4-Pro~\cite{gpt5.4} or Gemini-3.1-Pro~\cite{team2023gemini}) together with a meta prompt including system prompt and user prompt. The VLM then returns a detailed description of the image content and degradations, along with explicit instructions for removing those degradations. As can be seen, the generated prompt is highly detailed and specific. Tab.~\ref{tab:model_prompt_excl_realesrgan} of the main paper demonstrate that most models benefit from VLM-based prompts, which provides detailed, image-specific guidance, and this effect is particularly strong for the Nano-Banana-2 that we finally adopt. Between GPT-based and Gemini-based VLM prompts, we choose the latter because it achieves better quantitative performance, and GPT-generated prompts often blur or anonymize facial regions. Therefore, we use the Gemini-based VLM prompt in our final setting. The other meta prompts are given below.

\vspace{2mm} \noindent\textbf{General Mixed:}
\begin{quote}\small
\textbf{System prompt:} You are an expert in image restoration. The user will provide one degraded image. Analyze: (1) scene content and key objects/structures; (2) degradation types such as sensor/read noise, compression artifacts, ringing/blocking, defocus blur, motion blur, and their combinations; and (3) severity distribution by region, especially where blur, noise, or compression is strongest. Then output one detailed English restoration prompt. It must set fidelity as the first priority and preserve the exact scene content, geometry, position, and structure; remove noise and compression artifacts cleanly while recovering detail and clarity for visible structures; keep the global color and overall tone unchanged as much as possible, avoiding stylized recoloring and hue-category shifts; keep object count, object size, and object position unchanged, with no translation, warping, deformation, or shape distortion; keep geometry and content unchanged, with no hallucination and no object insertion or removal; and, if human faces are present, restore them clearly with special attention to skin texture, eyes, and hair, without changing identity or distinctive facial features. Output only the prompt text, with no explanation.

\textbf{User prompt:} Analyze this image and produce a detailed restoration prompt in English. You must include all hard constraints in the system prompt without omission. Focus on strong denoising, deartifacting, and clarity restoration for realistic degradations, including noise, compression artifacts, defocus blur, motion blur, and their combinations. Do not specifically prioritize foreground or background; instead require balanced full-image clarity improvement. Keep content, geometry, and color identity unchanged, avoid hallucination in uncertain regions. Output only the prompt.
\end{quote}

\vspace{2mm} \noindent\textbf{Rain:}
\begin{quote}\small
\textbf{System prompt:} You are an expert in rainy-scene restoration. The user will provide one degraded image. Analyze: (1) scene content, depth layers, and moving/static structures; (2) rain artifacts, including streaks, raindrops, misty veil, contrast reduction, and windshield-like droplets; and (3) severity and affected regions. Then output one detailed English restoration prompt. It must remove rain artifacts cleanly and as completely as reliably possible, including visible rain streaks, raindrop artifacts, and rain haze veil; preserve the rainy weather ambiance and lighting mood while not leaving obvious residual rain artifacts on visible content; actively restore background sharpness and visibility when rain streaks or droplets reduce clarity, rather than merely removing rain while leaving the background blurred; recover contrast, visibility, textures, and overall clarity without making the scene appear dry or sunny; perform denoising and sharpness recovery for all regions, not only rain removal; keep content strictly unchanged, including scene/object identity, count, size, position, and geometry; keep color and overall tone unchanged, with no hue shift, saturation shift, white-balance change, or recoloring; avoid ghosting, smearing, and over-smoothing on fine edges; keep geometry and content unchanged with no hallucination; and, if human faces are present, restore them clearly with special attention to skin texture, eyes, and hair, without changing identity or distinctive facial features. Output only the prompt text, with no explanation.

\textbf{User prompt:} Analyze this rainy image and produce a detailed restoration prompt in English. You must include all hard constraints in the system prompt without omission. Rain should be removed cleanly and as completely as possible, including rain streaks, droplets, and veil, and the result should not merely remove rain while keeping the background blurred. Preserve the rainy ambiance, keep content, geometry, and object layout unchanged, and keep global color and tone unchanged. Do not output any instruction related to watermark, logo, or text removal. Output only the prompt.
\end{quote}

\vspace{2mm} \noindent\textbf{Snow:}
\begin{quote}\small\textbf{System prompt:} You are an expert in snowy-scene restoration. The user will provide one degraded image. Analyze: (1) scene content and key objects; (2) snow-related degradations, including falling-flake occlusion, low contrast, haze-like veil, color desaturation, blur, and noise; and (3) severity and where visibility is impaired. Then output one detailed English restoration prompt. It must remove snow artifacts cleanly and as completely as possible, especially falling snowflakes, snow particles, and snow veil that occlude visibility; keep the scene content authentic and restore a clear view without hallucinating uncertain details; perform denoising and clarity/sharpness recovery for all regions, not only snow suppression; recover edges, details, and balanced contrast without harsh oversharpening; keep global color and overall tone unchanged, with no hue shift, white-balance shift, or recoloring; keep the global brightness style consistent while allowing clarity improvement; keep geometry and content unchanged with no hallucination; and, if human faces are present, restore them clearly with special attention to skin texture, eyes, and hair, without changing identity or distinctive facial features. Output only the prompt text, with no explanation.

\textbf{User prompt:} Analyze this snowy image and produce a detailed restoration prompt in English. You must include all hard constraints in the system prompt without omission. Snow should be removed as cleanly as possible, especially visible falling flakes and particles, while keeping content and geometry unchanged. Keep global color and tone unchanged, allow the image to become clearer through denoising and detail recovery. Output only the prompt.
\end{quote}

\vspace{2mm} \noindent\textbf{Low-Light:}
\begin{quote}\small
\textbf{System prompt:} You are an expert in low-light image restoration. The user will provide one degraded image. Analyze: (1) scene content and key subjects; (2) low-light symptoms, including underexposure, high-ISO noise, color cast, weak local contrast, and detail loss in dark regions; and (3) severity as well as shadow/highlight distribution. Then output one detailed English restoration prompt. It must set fidelity as the first priority and preserve the exact scene content, geometry, and structure; apply clearly stronger low-light enhancement, rather than conservative under-enhancement, to substantially improve visibility in dark regions; prioritize lifting shadow readability and recovering dim details that are already present while preserving a realistic night or dusk atmosphere; avoid hallucinating details in severely dark or uncertain regions; suppress noise and color blotches while restoring clarity and sharpness for visible structures without oversmoothing; preserve highlights and avoid over-exposure, clipping, gray-black lifting artifacts, and unnatural HDR appearance; allow brightness and saturation depth to change due to enhancement, but keep color identity consistent, e.g., red objects should remain red and yellow lights should remain yellow, with no hue-category shift; keep geometry and content unchanged, with no hallucination, insertion, or removal of objects; keep object count, object size, and object position unchanged, with no translation, warping, deformation, or shape distortion; and, if human faces are present, restore them clearly with special attention to skin texture, eyes, and hair, without changing identity or distinctive facial features. Output only the prompt text, with no explanation.

\textbf{User prompt:} Analyze this low-light image and produce a detailed restoration prompt in English. You must include all hard constraints in the system prompt without omission. Fidelity is the top priority. Apply clearly stronger low-light enhancement, not subtle enhancement, to make dark-region content substantially more visible, together with denoising and clarity recovery, while avoiding hallucination in severely dark uncertain regions. Brightness and saturation may become stronger, but color identity must remain the same. Keep geometry, content, and object layout unchanged, and output only the prompt.
\end{quote}

\begingroup
\renewcommand{\thefigure}{C.\arabic{figure}}
\setcounter{figure}{0}
\begin{figure*}[ph]
    \centering
    \includegraphics[width=\linewidth]{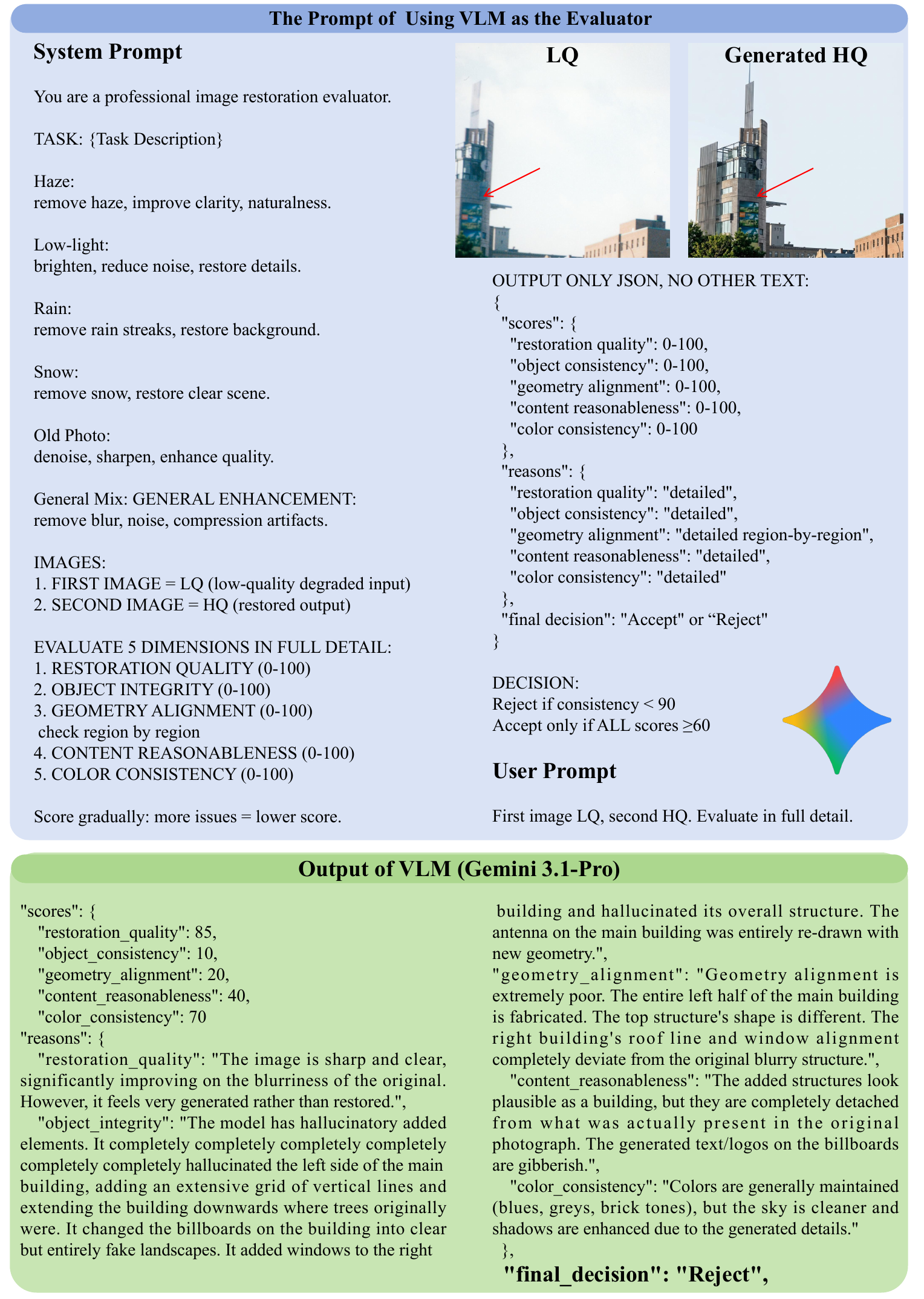}
    \caption{An example of VLM-based evaluation. Given an LQ input and its generated HQ counterpart, the VLM scores the result from five aspects: restoration quality, object consistency, geometry alignment, content reasonableness, and color consistency, and then outputs a final \texttt{Accept}/\texttt{Reject} decision. In this example, although the generated result looks reasonably clear, it noticeably shifts the scene layout and hallucinates a complete building, leading to low scores on object consistency and geometry alignment and thus a final rejection.}
    \label{fig:vlm1}
\end{figure*}
\endgroup

\vspace{2mm} \noindent\textbf{Old Photo:}
\begin{quote}\small
\textbf{System prompt:} You are an expert in old-photo restoration. The user will provide one degraded image. Analyze: (1) scene content and important historical visual details; (2) degradation types in old photos, including fading, grain/noise, scratches, dust spots, blur, and stains; and (3) severity and regions requiring conservative repair. Then output one detailed English restoration prompt. It must set fidelity as the first priority and preserve the exact scene content, geometry, and historical visual identity; apply stronger denoising and stronger clarity/detail restoration than default while avoiding synthetic oversharpening halos; keep the original color or monochrome characteristics and never colorize old photos; keep color exactly unchanged, with no hue shift, saturation change, white-balance change, or recoloring; avoid any global or local color remapping or grading and strictly preserve the original old-photo tone; preserve facial identity, texture identity, and document-like authenticity; keep geometry and content unchanged, with no object insertion or removal; keep object count, object size, and object position unchanged, with no translation, warping, deformation, or shape distortion; and, if human faces are present, restore them clearly with special attention to skin texture, eyes, and hair, without changing identity or distinctive facial features. Output only the prompt text, with no explanation.

\textbf{User prompt:} Analyze this old photo and produce a detailed restoration prompt in English. You must include all hard constraints in the system prompt without omission. Fidelity is the top priority: perform stronger denoising together with stronger clarity and detail restoration. Keep the original old-photo style and never colorize monochrome photos. Keep color exactly unchanged, keep geometry, content, and object layout unchanged, and output only the prompt.
\end{quote}

\begingroup
\renewcommand{\thefigure}{C.\arabic{figure}}
\setcounter{figure}{1}
\begin{figure*}[th]
    \centering
    \includegraphics[width=1\linewidth]{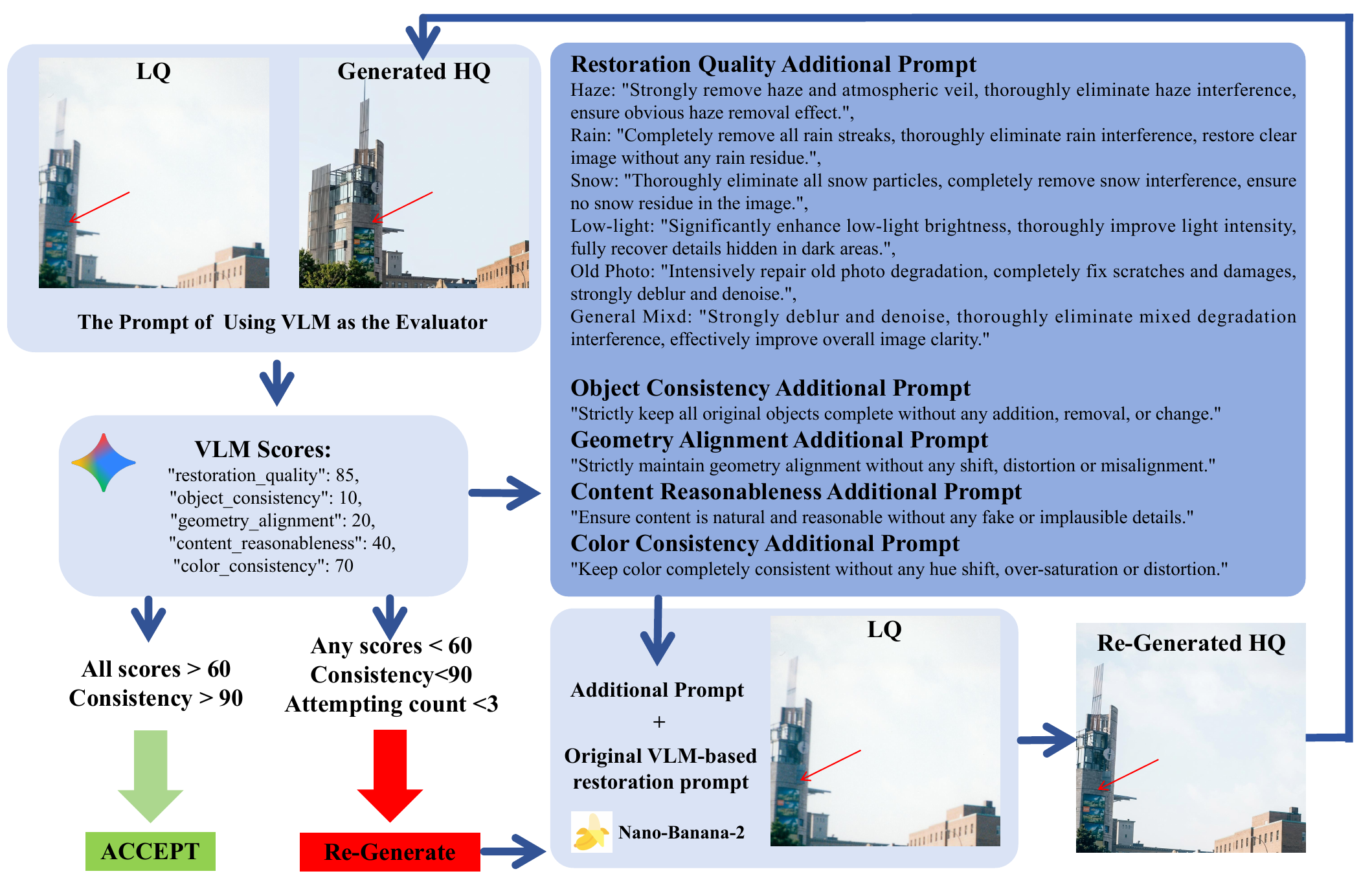}
    \caption{An example of VLM-assisted quality control. For a sample rejected by the VLM, we examine the failed aspects in the feedback, append corresponding strengthening instructions to the original prompt, and use Nano-Banana-2 to regenerate the HQ target for re-evaluation. If the new result is accepted, it is included in GGT-100K; otherwise, we allow one more generation attempt with the same prompt, for up to three attempts in total.}
    \label{fig:vlm2}
\end{figure*}

\section{How to Use VLM as the Evaluator and Quality Controller}
\label{app:c}

% \begin{figure*}[t]
%     \centering
%     \includegraphics[width=\linewidth]{images/VLM1.pdf}
%     \caption{An example of VLM-based evaluation. Given an LQ input and its generated HQ counterpart, the VLM scores the result from five aspects: restoration quality, object consistency, geometry alignment, content reasonableness, and color consistency, and then outputs a final \texttt{Accept}/\texttt{Reject} decision. In this example, although the generated result looks reasonably clear, it noticeably shifts the scene layout and hallucinates a complete building, leading to low scores on object consistency and geometry alignment and thus a final rejection.}
%     \label{fig:vlm1}
% \end{figure*}

% \setcounter{figure}{1}
% \begin{figure*}[th]
%     \centering
%     \includegraphics[width=\linewidth]{images/VLM2.pdf}
%     \caption{An example of VLM-assisted quality control. For a sample rejected by the VLM, we examine the failed aspects in the feedback, append corresponding strengthening instructions to the original prompt, and use Nano-Banana-2 to regenerate the HQ target for re-evaluation. If the new result is accepted, it is included in GGT-100K; otherwise, we allow one more generation attempt with the same prompt, for up to three attempts in total.}
%     \label{fig:vlm2}
% \end{figure*}

In this section, we describe how we use a VLM as both an \textit{evaluator} and a \textit{quality controller} in our pipeline. Specifically, we adopt Gemini-3.1-Pro~\cite{team2023gemini} as the vision-language evaluator. Given an LQ input image and its restored or generated HQ counterpart, the VLM is asked to judge whether the HQ result is acceptable for restoration purposes.

\vspace{2mm} \noindent\textbf{VLM-R Metric.}
We define \textit{VLM-based Restoration Success Rate} (VLM-R) to evaluate whether a restored/generated HQ image is acceptable for real-world image restoration. Specifically, as shown in Fig.~\ref{fig:vlm1}, given an LQ input image and its restored or generated HQ counterpart, we use a VLM to judge the result from five aspects: \textit{restoration quality}, \textit{object consistency}, \textit{geometry alignment}, \textit{content reasonableness}, and \textit{color consistency}. For each image pair, the VLM returns a structured JSON output containing a score from 0 to 100 for each aspect, detailed reasons, and a final binary decision: \texttt{Accept} or \texttt{Reject}. For example, Fig.~\ref{fig:vlm1} shows a generated result whose restoration clarity is reasonably good, but the model noticeably shifts the scene layout and hallucinates a complete building that is not faithfully aligned with the input. As a result, its \emph{object consistency} and \emph{geometry alignment} receive low scores, and the VLM finally rejects this sample.

We compute VLM-R as the proportion of test samples that are judged as \texttt{Accept} by the VLM:
\begin{equation}
\mathrm{VLM\mbox{-}R}=\frac{1}{N}\sum_{i=1}^{N}\mathbb{I}(d_i=\texttt{Accept})\times 100\%,
\end{equation}
where $N$ is the number of evaluated samples, $d_i$ is the final VLM decision for the $i$-th sample, and $\mathbb{I}(\cdot)$ is the indicator function. A higher VLM-R indicates that the restoration method more consistently produces results that are perceptually plausible and content-faithful to serve as restoration targets.

\vspace{2mm} \noindent\textbf{Using VLM for Dataset Quality Control.}
In addition to evaluation, the VLM also plays an important role in dataset construction. During the screening stage of GGT-100K, we feed the generated HQ candidate together with its LQ input into the VLM. If the result is accepted, the pair is kept. If it is rejected, we further inspect the scores and textual feedback returned by the VLM, and then append additional instructions to the original generation prompt and ask the MFM to regenerate the HQ target. As shown in Fig.~\ref{fig:vlm2}, since the scores are not satisfactory, the VLM rejects the generated image. We then examine which aspects fail, append strengthening instructions to the original prompt according to the VLM feedback, and use Nano-Banana-2 to generate a new result for re-evaluation. If the regenerated result is accepted, it is included in GGT-100K. Otherwise, we perform one more generation attempt without further changing the prompt, since repeated sampling with the same prompt can still yield a better result. In total, we allow up to three attempts for each sample.

\section{Details of User Study}
\label{app:user}

\renewcommand{\thefigure}{D.\arabic{figure}}
\setcounter{figure}{0}
\begin{figure*}[t]
    \centering
    \includegraphics[width=1\linewidth]{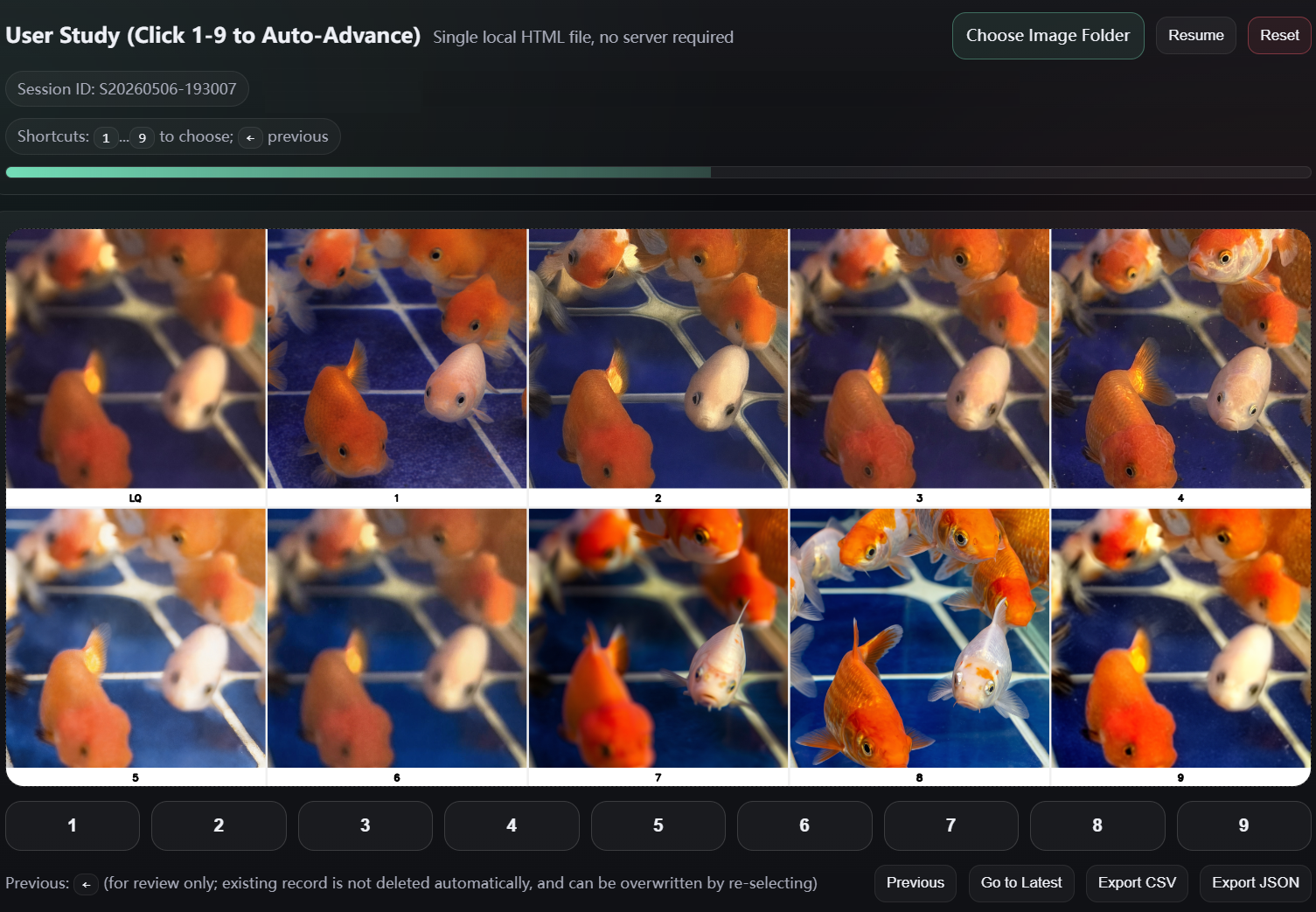}
    \caption{Interface of our user study. For each test image, participants are shown the LQ input together with nine anonymously shuffled restoration results from different MFMs, and are asked to select the best one by jointly considering perceptual quality and fidelity to the input image.}
    \label{fig:user_study}
    \vspace{-2mm}
\end{figure*}

In addition to quantitative metrics and VLM-based evaluation, we further conduct a user study including 20 participants on 200 real-world test images to directly assess the perceptual preference of different MFMs. As illustrated in Fig.~\ref{fig:user_study}, for each test image, we present the LQ input together with the restoration results of the nine candidate MFMs under their respective best-performing prompt settings. The nine results are randomly shuffled and anonymously indexed as 1--9, so that participants do not know which result is produced from which model.

Each participant is asked to select the single best result for each image with the LQ image as reference. The selection criterion is not only restoration quality, such as clearness, naturalness, and perceptual appeal, but also similarity and faithfulness to the input LQ image -- whether the restored result preserves the original scene content and avoids unreasonable changes.

After collecting the responses from all 20 participants, we apply a voting-based aggregation mechanism to obtain a more robust image-level preference result. Specifically, for each test image, the 20 participants can be viewed as voting for the nine anonymous model outputs, and the model receiving the highest number of votes is regarded as the best model for that image. We then count for each model, how many times it wins at the image-level, and report the corresponding ratio as the final human preference score in the main paper.

This voting protocol also explains why the ``Human'' results in Tab.~\ref{tab:model_prompt_excl_realesrgan} are reported in unit of 0.5\%. Since our user study is conducted on 200 real-world images, one image-level win corresponds to 0.5\%. Therefore, a score of 0\% does not mean that a model is never selected by any participant. For example, GPT-Image-1.5 may still receive some individual votes on certain images, but it never becomes the final image-level winner after aggregating the votes of all 20 participants, and thus its final score is 0\%.

\section{Details of Training Settings}
\label{app:e}

\begin{table}[t]
\centering
\caption{Composition of the existing training data. ``Real'' indicates whether the dataset is based on real captured image pairs rather than purely synthetic pairs.}
\label{tab:train_data_comp}
\tiny
\setlength{\tabcolsep}{3.2pt}
\resizebox{0.78\columnwidth}{!}{
\begin{tabular}{p{1.7cm} p{4.6cm} c c c}
\toprule
\textbf{Category} & \textbf{Dataset(s)} & \textbf{Year} & \textbf{Real} & \textbf{\# Pairs} \\
\midrule
\multirow{8}{*}{General Mixed}
& DF2K + RealESRGAN degr.~\cite{div2k,flick2k,wang2021real} & 2021 & No & 58K \\
& RealSR~\cite{realsr} & 2019 & Yes & 13K \\
& GoPro~\cite{cho2021_deblur} & 2021 & Yes & 16K \\
& FoundIR-LoVIF (Blur)~\cite{li2025foundir,chen2026lovif} & 2026 & Yes & 5K \\
& RealBlur~\cite{rim2020realblur} & 2020 & Yes & 8K \\
& SIDD~\cite{sidd} & 2018 & Yes & 16K \\
& PolyU-Noisy~\cite{polyunoise} & 2018 & Yes & 4K \\
& \textbf{Subtotal} & -- & -- & \textbf{120K} \\
\midrule
\multirow{4}{*}{Low Light}
& LOL~\cite{LoL} & 2018 & Yes & 10K \\
& FoundIR-LoVIF (Lowlight)~\cite{li2025foundir,chen2026lovif} & 2026 & Yes & 5K \\
& UHD-LL~\cite{UHDLL} & 2022 & Yes & 5K \\
& \textbf{Subtotal} & -- & -- & \textbf{20K} \\
\midrule
\multirow{3}{*}{Haze}
& RESIDE~\cite{RESIDE} & 2019 & No & 10K \\
& WeatherBench (haze)~\cite{guan2025weatherbench} & 2025 & Yes & 10K \\
& \textbf{Subtotal} & -- & -- & \textbf{20K} \\
\midrule
\multirow{5}{*}{Rain}
& Rain13K~\cite{Rain100,rain1200} & 2018 & No & 10K \\
& FoundIR-LoVIF (Rain)~\cite{li2025foundir,chen2026lovif} & 2026 & Yes & 5K \\
& RainDS~\cite{rainds} & 2025 & Yes & 3K \\
& RealRain~\cite{realrain} & 2025 & Yes & 2K \\
& \textbf{Subtotal} & -- & -- & \textbf{20K} \\
\midrule
\multirow{3}{*}{Snow}
& Snow100K-Synthetic~\cite{snow100k} & 2021 & No & 10K \\
& WeatherBench (snow)~\cite{guan2025weatherbench} & 2025 & Yes & 10K \\
& \textbf{Subtotal} & -- & -- & \textbf{20K} \\
\midrule
\textbf{Total} & \textbf{Existing training data} & -- & -- & \textbf{200K} \\
\bottomrule
\end{tabular}}
\end{table}

As summarized in Tab.~\ref{tab:train_data_comp}, we construct the set of existing training data by explicitly controlling both degradation categories and sample counts. In particular, the general mixed category is set to 120K pairs, accounting for about half of the full training pool, which is roughly consistent with the proportion of general mixed samples in GGT-100K (about 60\%). The remaining 100K pairs are distributed across low-light, haze, rain, snow, and noise categories with comparable scale, so that the baseline setting covers diverse real-world restoration scenarios while maintaining a relatively balanced composition.

Specifically, the resulting training set contains 58K pairs from DF2K with RealESRGAN degradations~\cite{div2k,flick2k,wang2021real}, 13K from RealSR~\cite{realsr}, 16K from GoPro~\cite{cho2021_deblur}, 5K from FoundIR-LoVIF (Blur)~\cite{li2025foundir,chen2026lovif}, 8K from RealBlur~\cite{rim2020realblur}, 10K from LOL~\cite{LoL}, 5K from FoundIR-LoVIF (Lowlight)~\cite{li2025foundir,chen2026lovif}, 5K from UHD-LL~\cite{UHDLL}, 10K from RESIDE~\cite{RESIDE}, 10K from WeatherBench (haze)~\cite{guan2025weatherbench}, 10K from Rain13K~\cite{Rain100,rain1200}, 5K from FoundIR-LoVIF (Rain)~\cite{li2025foundir,chen2026lovif}, 3K from RainDS~\cite{rainds}, 2K from RealRain~\cite{realrain}, 10K from Snow100K~\cite{snow100k}, 10K from WeatherBench (snow)~\cite{guan2025weatherbench}, 16K from SIDD~\cite{sidd}, and 4K from PolyU-Noisy~\cite{polyunoise}.

For most source datasets, we keep the original image resolution unchanged when constructing the training pool. For datasets with very large image resolutions, such as RealSR~\cite{realsr} and SIDD~\cite{sidd}, we first crop the images into $512\times512$ patches and use these patches to form the candidate pool. We then randomly sample the required number of images or patches from each dataset to construct the final training set. If the number of available samples in a certain category is insufficient, we perform re-sampling from that category to meet the target amount. This design makes the comparison fair: it preserves broad coverage of datasets and degradation types, while avoiding the severe imbalance that would arise from directly merging all available data, where a few large-scale datasets may otherwise dominate the training distribution.

\renewcommand{\thefigure}{F.\arabic{figure}}
\setcounter{figure}{0}

\section{Detailed Results of Specific Degradations}
\label{app:h}

In Tab.~\ref{tab:main_ours} of the main paper, we report only the average results over all evaluated degradation groups. Here, we provide detailed comparisons of training \textbf{w/o} and \textbf{w/} GGT-100K on each specific category, including general mixed degradations (Tab.~\ref{tab:main_mix}), rain (Tab.~\ref{tab:main_rain}), haze (Tab.~\ref{tab:main_haze}), snow (Tab.~\ref{tab:main_snow}), low-light (Tab.~\ref{tab:main_lowlight}), and old photos (Tab.~\ref{tab:main_oldphoto}).

A key observation from Tabs.~\ref{tab:main_mix}--\ref{tab:main_oldphoto} is that the benefit of GGT-100K is not limited to a specific degradation category. This finding is important because the gains brought by GGT-100K do not merely come from compensating for missing categories in the original training pool, such as old photos, which are not included in the baseline training data. Instead, adding GGT-100K generally improves restoration performance across all the six evaluated real-world degradation groups and different model families, including CNN/transformer backbones, all-in-one restoration models, and recent generative restoration models. These gains are reflected not only in full-reference fidelity metrics such as PSNR, SSIM, LPIPS, and DISTS, but also in perceptual quality metrics such as MUSIQ and AFINE-NR.

At the same time, the category-wise results also reveal an instructive exception for Qwen-Image-Edit on the rain and snow subsets. After finetuning with GGT-100K, its PSNR becomes lower than the \textbf{w/o} setting, even though it remains substantially higher than the \textbf{official} model. We consider this behavior reasonable rather than problematic. As shown in Fig.~\ref{fig:app-withofficial}, Qwen-Image-Edit trained with GGT-100K tends to generate richer and more realistic details. While such details improve visual quality and remain semantically faithful to the input, they may not exactly match the specific GGT reference at the pixel level and, therefore, do not always translate into higher PSNR. Meanwhile, its LPIPS and DISTS are improved, indicating better perceptual fidelity and more plausible fine details. This interpretation is also consistent with the clearly stronger perceptual metrics and the visual comparisons in Fig.~\ref{fig:app-withofficial}, where Qwen-Image-Edit trained with GGT-100K achieves more appealing results while maintaining acceptable overall faithfulness.

It is also worth noting that these categories should not be interpreted as isolated single-degradation settings. In practice, each sample usually contains mixed degradations, while the category label only reflects the most prominent visible factor. Therefore, the overall trend across multiple categories suggests that GGT-100K does not mainly enhance performance on one specific degradation type, but instead serves as a general-purpose real-world supervision source for various restoration scenarios.

% mix
\begin{table*}[t]
\centering
\caption{Comparison of representative restoration models ``w/o'' and ``w/'' GGT-100K on the \textbf{GGT100K-General Mixed} test set. For some models whose official releases can also handle real-world degradations, we additionally report their \textbf{official results}. ``Improvement'' indicates the performance gain brought by GGT-100K. The positive improvements are highlighted in \textcolor{red}{\textbf{red}}.}
\label{tab:main_mix}
\vspace{1mm}
\resizebox{\textwidth}{!}{
\begin{tabular}{c|c|cccc|ccccc|c}
\toprule
\multirow{2}{*}{Model} & \multirow{2}{*}{GGT-100K} & \multicolumn{4}{c|}{Full-reference fidelity metrics} & \multicolumn{5}{c|}{No-reference perceptual metrics} & \multirow{2}{*}{VLM-R $\uparrow$} \\
\cmidrule(lr){3-6}\cmidrule(lr){7-11}
& & PSNR $\uparrow$ & SSIM $\uparrow$ & LPIPS $\downarrow$ & DISTS $\downarrow$ & NIQE $\downarrow$ & MUSIQ $\uparrow$ & MANIQA $\uparrow$ & TOPIQ $\uparrow$ & AFINE-NR $\downarrow$ &  \\
\midrule
\multirow{3}{*}{MPRNet~\cite{zamir2021multi}}
& w/o & 27.0125 & 0.7608 & 0.3884 & 0.2365 & 7.6629 & 38.6702 & 0.4163 & 0.2754 & -0.6966 & 30.8\% \\
& w/ & 29.6144 & 0.8123 & 0.3502 & 0.2215 & 7.9104 & 41.2476 & 0.3927 & 0.2689 & -0.7649 & 38.0\% \\
& Improvement & \textcolor{red}{\textbf{+2.6019}} & \textcolor{red}{\textbf{+0.0515}} & \textcolor{red}{\textbf{-0.0382}} & \textcolor{red}{\textbf{-0.0150}} & +0.2475 & \textcolor{red}{\textbf{+2.5774}} & -0.0236 & -0.0065 & \textcolor{red}{\textbf{-0.0683}} & \textcolor{red}{\textbf{+7.2\%}} \\
\midrule
\multirow{3}{*}{NAFNet~\cite{chen2022simple}}
& w/o & 27.3763 & 0.7697 & 0.3730 & 0.2252 & 7.4243 & 39.1719 & 0.4152 & 0.2731 & -0.6854 & 34.0\% \\
& w/ & 30.0298 & 0.8237 & 0.3271 & 0.2111 & 7.7282 & 44.2564 & 0.3875 & 0.2795 & -0.7760 & 57.6\% \\
& Improvement & \textcolor{red}{\textbf{+2.6535}} & \textcolor{red}{\textbf{+0.0540}} & \textcolor{red}{\textbf{-0.0459}} & \textcolor{red}{\textbf{-0.0141}} & +0.3039 & \textcolor{red}{\textbf{+5.0845}} & -0.0277 & \textcolor{red}{\textbf{+0.0064}} & \textcolor{red}{\textbf{-0.0906}} & \textcolor{red}{\textbf{+23.6\%}} \\
\midrule
\multirow{3}{*}{SwinIR~\cite{liang2021swinir}}
& w/o & 26.1689 & 0.7645 & 0.3838 & 0.2269 & 7.1076 & 37.6627 & 0.4071 & 0.2726 & -0.6809 & 27.6\% \\
& w/ & 29.4102 & 0.8109 & 0.3424 & 0.2124 & 7.6969 & 40.2668 & 0.3940 & 0.2612 & -0.7366 & 46.0\% \\
& Improvement & \textcolor{red}{\textbf{+3.2413}} & \textcolor{red}{\textbf{+0.0464}} & \textcolor{red}{\textbf{-0.0414}} & \textcolor{red}{\textbf{-0.0145}} & +0.5893 & \textcolor{red}{\textbf{+2.6041}} & -0.0131 & -0.0114 & \textcolor{red}{\textbf{-0.0557}} & \textcolor{red}{\textbf{+18.4\%}} \\
\midrule
\multirow{3}{*}{X-Restormer~\cite{X-Restormer}}
& w/o & 26.5815 & 0.7626 & 0.3685 & 0.2208 & 7.3953 & 39.8978 & 0.4190 & 0.2824 & -0.6917 & 36.4\% \\
& w/ & 30.0419 & 0.8237 & 0.3315 & 0.2147 & 7.8118 & 43.9505 & 0.3821 & 0.2779 & -0.7897 & 52.8\% \\
& Improvement & \textcolor{red}{\textbf{+3.4604}} & \textcolor{red}{\textbf{+0.0611}} & \textcolor{red}{\textbf{-0.0370}} & \textcolor{red}{\textbf{-0.0061}} & +0.4165 & \textcolor{red}{\textbf{+4.0527}} & -0.0369 & -0.0045 & \textcolor{red}{\textbf{-0.0980}} & \textcolor{red}{\textbf{+16.4\%}} \\
\midrule
\multirow{3}{*}{PromptIR~\cite{PromptIR}}
& w/o & 26.2855 & 0.7469 & 0.3760 & 0.2290 & 7.2540 & 39.6063 & 0.4193 & 0.2791 & -0.6945 & 28.8\% \\
& w/ & 30.0109 & 0.8227 & 0.3282 & 0.2121 & 7.7132 & 43.6127 & 0.3907 & 0.2775 & -0.7755 & 50.4\% \\
& Improvement & \textcolor{red}{\textbf{+3.7254}} & \textcolor{red}{\textbf{+0.0758}} & \textcolor{red}{\textbf{-0.0478}} & \textcolor{red}{\textbf{-0.0169}} & +0.4592 & \textcolor{red}{\textbf{+4.0064}} & -0.0286 & -0.0016 & \textcolor{red}{\textbf{-0.0810}} & \textcolor{red}{\textbf{+21.6\%}} \\
\midrule
\multirow{3}{*}{MoCE-IR~\cite{MoCE-IR}}
& w/o & 26.7865 & 0.7460 & 0.3800 & 0.2299 & 7.3464 & 39.8210 & 0.4150 & 0.2791 & -0.6848 & 26.8\% \\
& w/ & 30.0635 & 0.8270 & 0.3238 & 0.2134 & 7.7319 & 45.6372 & 0.3908 & 0.2907 & -0.7965 & 55.6\% \\
& Improvement & \textcolor{red}{\textbf{+3.2770}} & \textcolor{red}{\textbf{+0.0810}} & \textcolor{red}{\textbf{-0.0562}} & \textcolor{red}{\textbf{-0.0165}} & +0.3855 & \textcolor{red}{\textbf{+5.8162}} & -0.0242 & \textcolor{red}{\textbf{+0.0116}} & \textcolor{red}{\textbf{-0.1117}} & \textcolor{red}{\textbf{+28.8\%}} \\
\midrule
\multirow{3}{*}{DA-CLIP~\cite{DA-CLIP}}
& w/o & 28.3684 & 0.7772 & 0.3324 & 0.2019 & 6.3042 & 37.8845 & 0.4542 & 0.2863 & -0.6471 & 39.6\% \\
& w/ & 28.9687 & 0.7873 & 0.3000 & 0.1901 & 6.7152 & 39.9182 & 0.4528 & 0.2788 & -0.6897 & 52.0\% \\
& Improvement & \textcolor{red}{\textbf{+0.6003}} & \textcolor{red}{\textbf{+0.0101}} & \textcolor{red}{\textbf{-0.0324}} & \textcolor{red}{\textbf{-0.0118}} & +0.4110 & \textcolor{red}{\textbf{+2.0337}} & -0.0014 & -0.0075 & \textcolor{red}{\textbf{-0.0426}} & \textcolor{red}{\textbf{+12.4\%}} \\
\midrule
\multirow{4}{*}{FoundIR~\cite{li2025foundir}}
& \textbf{official} & 28.6626 & 0.7885 & 0.3543 & 0.2152 & 7.1874 & 35.2292 & 0.4333 & 0.2609 & -0.6765 & 39.6\% \\
& w/o & 27.6491 & 0.7787 & 0.3618 & 0.2225 & 7.5651 & 38.5844 & 0.4257 & 0.2726 & -0.7076 & 41.6\% \\
& w/ & 29.2213 & 0.8100 & 0.3532 & 0.2233 & 8.0795 & 39.5944 & 0.3993 & 0.2633 & -0.7660 & 60.0\% \\
& Improvement & \textcolor{red}{\textbf{+1.5722}} & \textcolor{red}{\textbf{+0.0313}} & \textcolor{red}{\textbf{-0.0086}} & +0.0008 & +0.5144 & \textcolor{red}{\textbf{+1.0100}} & -0.0264 & -0.0093 & \textcolor{red}{\textbf{-0.0584}} & \textcolor{red}{\textbf{+18.4\%}} \\
\midrule
\multirow{3}{*}{FLUX-Controlnet~\cite{FLUX}}
& w/o & 22.5870 & 0.6359 & 0.3794 & 0.2116 & 5.4614 & 50.5574 & 0.4994 & 0.3788 & -0.6977 & 23.6\% \\
& w/ & 24.0212 & 0.7111 & 0.2658 & 0.1527 & 4.9972 & 64.6220 & 0.5805 & 0.5037 & -0.9153 & 71.2\% \\
& Improvement & \textcolor{red}{\textbf{+1.4342}} & \textcolor{red}{\textbf{+0.0752}} & \textcolor{red}{\textbf{-0.1136}} & \textcolor{red}{\textbf{-0.0589}} & \textcolor{red}{\textbf{-0.4642}} & \textcolor{red}{\textbf{+14.0646}} & \textcolor{red}{\textbf{+0.0811}} & \textcolor{red}{\textbf{+0.1249}} & \textcolor{red}{\textbf{-0.2176}} & \textcolor{red}{\textbf{+47.6\%}} \\
\midrule
\multirow{4}{*}{Qwen-Image-Edit~\cite{Qwen-Image-Edit}}
& \textbf{official} & 24.8687 & 0.7549 & 0.2757 & 0.1561 & 5.9458 & 57.8684 & 0.5299 & 0.4345 & -0.8978 & 75.2\% \\
& w/o & 26.8753 & 0.7393 & 0.3028 & 0.1686 & 6.2237 & 52.0401 & 0.5252 & 0.3768 & -0.8089 & 76.4\% \\
& w/ & 27.6201 & 0.7590 & 0.2184 & 0.1186 & 5.7173 & 63.3238 & 0.5763 & 0.4674 & -0.9328 & 92.0\% \\
& Improvement & \textcolor{red}{\textbf{+0.7448}} & \textcolor{red}{\textbf{+0.0197}} & \textcolor{red}{\textbf{-0.0844}} & \textcolor{red}{\textbf{-0.0500}} & \textcolor{red}{\textbf{-0.5064}} & \textcolor{red}{\textbf{+11.2837}} & \textcolor{red}{\textbf{+0.0511}} & \textcolor{red}{\textbf{+0.0906}} & \textcolor{red}{\textbf{-0.1239}} & \textcolor{red}{\textbf{+15.6\%}} \\
\bottomrule
\end{tabular}}

\end{table*}

% rain
\begin{table*}[t]
\centering
\caption{Comparison of representative restoration models ``w/o'' and ``w/'' GGT-100K on the \textbf{GGT100K-Rain} test set. For some models whose official releases can also handle real-world degradations, we additionally report their \textbf{official results}. ``Improvement'' indicates the performance gain brought by GGT-100K. The positive improvements are highlighted in \textcolor{red}{\textbf{red}}.}
\label{tab:main_rain}
\vspace{1mm}
\resizebox{\textwidth}{!}{
\begin{tabular}{c|c|cccc|ccccc|c}
\toprule
\multirow{2}{*}{Model} & \multirow{2}{*}{GGT-100K} & \multicolumn{4}{c|}{Full-reference fidelity metrics} & \multicolumn{5}{c|}{No-reference perceptual metrics} & \multirow{2}{*}{VLM-R $\uparrow$} \\
\cmidrule(lr){3-6}\cmidrule(lr){7-11}
& & PSNR $\uparrow$ & SSIM $\uparrow$ & LPIPS $\downarrow$ & DISTS $\downarrow$ & NIQE $\downarrow$ & MUSIQ $\uparrow$ & MANIQA $\uparrow$ & TOPIQ $\uparrow$ & AFINE-NR $\downarrow$ &  \\
\midrule
\multirow{3}{*}{MPRNet~\cite{zamir2021multi}}
& w/o & 24.2868 & 0.7742 & 0.3984 & 0.2266 & 6.6738 & 43.6375 & 0.4636 & 0.3345 & -0.7394 & 0.0\% \\
& w/ & 26.3659 & 0.8184 & 0.3952 & 0.2379 & 7.0804 & 45.8030 & 0.4344 & 0.3265 & -0.7867 & 4.0\% \\
& Improvement & \textcolor{red}{\textbf{+2.0791}} & \textcolor{red}{\textbf{+0.0442}} & \textcolor{red}{\textbf{-0.0032}} & +0.0113 & +0.4066 & \textcolor{red}{\textbf{+2.1655}} & -0.0292 & -0.0080 & \textcolor{red}{\textbf{-0.0473}} & \textcolor{red}{\textbf{+4.0\%}} \\
\midrule
\multirow{3}{*}{NAFNet~\cite{chen2022simple}}
& w/o & 25.5435 & 0.7899 & 0.3820 & 0.2195 & 6.5010 & 42.6554 & 0.4612 & 0.3356 & -0.7328 & 4.0\% \\
& w/ & 27.9179 & 0.8423 & 0.3442 & 0.2144 & 6.7682 & 44.6991 & 0.4180 & 0.3095 & -0.7802 & 42.0\% \\
& Improvement & \textcolor{red}{\textbf{+2.3744}} & \textcolor{red}{\textbf{+0.0524}} & \textcolor{red}{\textbf{-0.0378}} & \textcolor{red}{\textbf{-0.0051}} & +0.2672 & \textcolor{red}{\textbf{+2.0437}} & -0.0432 & -0.0261 & \textcolor{red}{\textbf{-0.0474}} & \textcolor{red}{\textbf{+38.0\%}} \\
\midrule
\multirow{3}{*}{SwinIR~\cite{liang2021swinir}}
& w/o & 23.5779 & 0.7677 & 0.4004 & 0.2281 & 6.3847 & 43.1183 & 0.4512 & 0.3437 & -0.7313 & 4.0\% \\
& w/ & 25.9225 & 0.8076 & 0.3953 & 0.2309 & 6.7414 & 45.1089 & 0.4284 & 0.3225 & -0.7527 & 8.0\% \\
& Improvement & \textcolor{red}{\textbf{+2.3446}} & \textcolor{red}{\textbf{+0.0399}} & \textcolor{red}{\textbf{-0.0051}} & +0.0028 & +0.3567 & \textcolor{red}{\textbf{+1.9906}} & -0.0228 & -0.0212 & \textcolor{red}{\textbf{-0.0214}} & \textcolor{red}{\textbf{+4.0\%}} \\
\midrule
\multirow{3}{*}{X-Restormer~\cite{X-Restormer}}
& w/o & 25.2709 & 0.7883 & 0.3700 & 0.2181 & 6.5006 & 43.8513 & 0.4555 & 0.3393 & -0.7577 & 6.0\% \\
& w/ & 27.7223 & 0.8457 & 0.3415 & 0.2185 & 6.8457 & 44.9040 & 0.4171 & 0.3170 & -0.7963 & 44.0\% \\
& Improvement & \textcolor{red}{\textbf{+2.4514}} & \textcolor{red}{\textbf{+0.0574}} & \textcolor{red}{\textbf{-0.0285}} & +0.0004 & +0.3451 & \textcolor{red}{\textbf{+1.0527}} & -0.0384 & -0.0223 & \textcolor{red}{\textbf{-0.0386}} & \textcolor{red}{\textbf{+38.0\%}} \\
\midrule
\multirow{3}{*}{PromptIR~\cite{PromptIR}}
& w/o & 25.1898 & 0.7835 & 0.3758 & 0.2168 & 6.3354 & 44.4670 & 0.4681 & 0.3396 & -0.7485 & 4.0\% \\
& w/ & 27.6848 & 0.8374 & 0.3569 & 0.2205 & 6.6246 & 45.5707 & 0.4305 & 0.3195 & -0.7878 & 22.0\% \\
& Improvement & \textcolor{red}{\textbf{+2.4950}} & \textcolor{red}{\textbf{+0.0539}} & \textcolor{red}{\textbf{-0.0189}} & +0.0037 & +0.2892 & \textcolor{red}{\textbf{+1.1037}} & -0.0376 & -0.0201 & \textcolor{red}{\textbf{-0.0393}} & \textcolor{red}{\textbf{+18.0\%}} \\
\midrule
\multirow{3}{*}{MoCE-IR~\cite{MoCE-IR}}
& w/o & 25.4884 & 0.7786 & 0.3849 & 0.2177 & 6.3647 & 44.7434 & 0.4720 & 0.3509 & -0.7464 & 4.0\% \\
& w/ & 27.3242 & 0.8474 & 0.3470 & 0.2273 & 6.9887 & 46.9637 & 0.4131 & 0.3247 & -0.8237 & 50.0\% \\
& Improvement & \textcolor{red}{\textbf{+1.8358}} & \textcolor{red}{\textbf{+0.0688}} & \textcolor{red}{\textbf{-0.0379}} & +0.0096 & +0.6240 & \textcolor{red}{\textbf{+2.2203}} & -0.0589 & -0.0262 & \textcolor{red}{\textbf{-0.0773}} & \textcolor{red}{\textbf{+46.0\%}} \\
\midrule
\multirow{3}{*}{DA-CLIP~\cite{DA-CLIP}}
& w/o & 26.1250 & 0.7919 & 0.3290 & 0.2001 & 6.3213 & 40.4140 & 0.4680 & 0.3386 & -0.7167 & 26.0\% \\
& w/ & 26.8624 & 0.8121 & 0.2961 & 0.1949 & 6.4706 & 40.5440 & 0.4568 & 0.3291 & -0.7388 & 44.0\% \\
& Improvement & \textcolor{red}{\textbf{+0.7374}} & \textcolor{red}{\textbf{+0.0202}} & \textcolor{red}{\textbf{-0.0329}} & \textcolor{red}{\textbf{-0.0052}} & +0.1493 & \textcolor{red}{\textbf{+0.1300}} & -0.0112 & -0.0095 & \textcolor{red}{\textbf{-0.0221}} & \textcolor{red}{\textbf{+18.0\%}} \\
\midrule
\multirow{4}{*}{FoundIR~\cite{li2025foundir}}
& \textbf{official} & 25.4558 & 0.7872 & 0.3507 & 0.2047 & 6.0888 & 41.5685 & 0.4726 & 0.3419 & -0.7170 & 4.0\% \\
& w/o & 25.8326 & 0.7998 & 0.3572 & 0.2103 & 6.7084 & 43.6225 & 0.4657 & 0.3382 & -0.7533 & 10.0\% \\
& w/ & 26.6317 & 0.8350 & 0.3594 & 0.2293 & 7.5112 & 43.6799 & 0.4201 & 0.3126 & -0.8160 & 48.0\% \\
& Improvement & \textcolor{red}{\textbf{+0.7991}} & \textcolor{red}{\textbf{+0.0352}} & +0.0022 & +0.0190 & +0.8028 & \textcolor{red}{\textbf{+0.0574}} & -0.0456 & -0.0256 & \textcolor{red}{\textbf{-0.0627}} & \textcolor{red}{\textbf{+38.0\%}} \\
\midrule
\multirow{3}{*}{FLUX-Controlnet~\cite{FLUX}}
& w/o & 23.2300 & 0.7218 & 0.3721 & 0.2093 & 5.9185 & 47.1210 & 0.5067 & 0.3977 & -0.7193 & 28.0\% \\
& w/ & 22.8710 & 0.7361 & 0.2989 & 0.1712 & 5.3477 & 60.0289 & 0.5640 & 0.4719 & -0.8999 & 66.0\% \\
& Improvement & -0.3590 & \textcolor{red}{\textbf{+0.0143}} & \textcolor{red}{\textbf{-0.0732}} & \textcolor{red}{\textbf{-0.0381}} & \textcolor{red}{\textbf{-0.5708}} & \textcolor{red}{\textbf{+12.9079}} & \textcolor{red}{\textbf{+0.0573}} & \textcolor{red}{\textbf{+0.0742}} & \textcolor{red}{\textbf{-0.1806}} & \textcolor{red}{\textbf{+38.0\%}} \\
\midrule
\multirow{4}{*}{Qwen-Image-Edit~\cite{Qwen-Image-Edit}}
& \textbf{official} & 20.3510 & 0.7320 & 0.3578 & 0.1906 & 5.8045 & 61.3429 & 0.5646 & 0.4950 & -1.0003 & 66.0\% \\
& w/o & 27.6252 & 0.8319 & 0.2698 & 0.1633 & 6.6923 & 42.5729 & 0.4704 & 0.3241 & -0.8073 & 94.0\% \\
& w/ & 24.5022 & 0.7774 & 0.2563 & 0.1425 & 5.7831 & 59.9861 & 0.5610 & 0.4606 & -0.9721 & 94.0\% \\
& Improvement & -3.1230 & -0.0545 & \textcolor{red}{\textbf{-0.0135}} & \textcolor{red}{\textbf{-0.0208}} & \textcolor{red}{\textbf{-0.9092}} & \textcolor{red}{\textbf{+17.4132}} & \textcolor{red}{\textbf{+0.0906}} & \textcolor{red}{\textbf{+0.1365}} & \textcolor{red}{\textbf{-0.1648}} & +0.0\% \\
\bottomrule
\end{tabular}}

\end{table*}

%haze
\begin{table*}[t]
\centering
\caption{Comparison of representative restoration models ``w/o'' and ``w/'' GGT-100K on the \textbf{GGT100K-Haze} test set. For some models whose official releases can also handle real-world degradations, we additionally report their \textbf{official results}. ``Improvement'' indicates the performance gain brought by GGT-100K. The positive improvements are highlighted in \textcolor{red}{\textbf{red}}.}
\label{tab:main_haze}
\vspace{1mm}
\resizebox{\textwidth}{!}{
\begin{tabular}{c|c|cccc|ccccc|c}
\toprule
\multirow{2}{*}{Model} & \multirow{2}{*}{GGT-100K} & \multicolumn{4}{c|}{Full-reference fidelity metrics} & \multicolumn{5}{c|}{No-reference perceptual metrics} & \multirow{2}{*}{VLM-R $\uparrow$} \\
\cmidrule(lr){3-6}\cmidrule(lr){7-11}
& & PSNR $\uparrow$ & SSIM $\uparrow$ & LPIPS $\downarrow$ & DISTS $\downarrow$ & NIQE $\downarrow$ & MUSIQ $\uparrow$ & MANIQA $\uparrow$ & TOPIQ $\uparrow$ & AFINE-NR $\downarrow$ &  \\
\midrule
\multirow{3}{*}{MPRNet~\cite{zamir2021multi}}
& w/o & 18.9375 & 0.7565 & 0.3843 & 0.2842 & 6.6753 & 42.6631 & 0.5098 & 0.3093 & -0.8065 & 8.0\% \\
& w/ & 19.9080 & 0.7774 & 0.3627 & 0.2622 & 6.5113 & 46.0173 & 0.5010 & 0.3209 & -0.8464 & 10.0\% \\
& Improvement & \textcolor{red}{\textbf{+0.9705}} & \textcolor{red}{\textbf{+0.0209}} & \textcolor{red}{\textbf{-0.0216}} & \textcolor{red}{\textbf{-0.0220}} & \textcolor{red}{\textbf{-0.1640}} & \textcolor{red}{\textbf{+3.3542}} & -0.0088 & \textcolor{red}{\textbf{+0.0116}} & \textcolor{red}{\textbf{-0.0399}} & \textcolor{red}{\textbf{+2.0\%}} \\
\midrule
\multirow{3}{*}{NAFNet~\cite{chen2022simple}}
& w/o & 19.0288 & 0.7579 & 0.3686 & 0.2717 & 6.0526 & 43.8009 & 0.5057 & 0.3168 & -0.7937 & 6.0\% \\
& w/ & 22.7388 & 0.8162 & 0.3013 & 0.2068 & 5.7603 & 50.2170 & 0.4688 & 0.3360 & -0.8402 & 20.0\% \\
& Improvement & \textcolor{red}{\textbf{+3.7100}} & \textcolor{red}{\textbf{+0.0583}} & \textcolor{red}{\textbf{-0.0673}} & \textcolor{red}{\textbf{-0.0649}} & \textcolor{red}{\textbf{-0.2923}} & \textcolor{red}{\textbf{+6.4161}} & -0.0369 & \textcolor{red}{\textbf{+0.0192}} & \textcolor{red}{\textbf{-0.0465}} & \textcolor{red}{\textbf{+14.0\%}} \\
\midrule
\multirow{3}{*}{SwinIR~\cite{liang2021swinir}}
& w/o & 18.6930 & 0.7468 & 0.4117 & 0.2724 & 6.3062 & 41.8782 & 0.4858 & 0.3051 & -0.7712 & 2.0\% \\
& w/ & 20.7340 & 0.7817 & 0.3571 & 0.2452 & 6.1837 & 45.5367 & 0.4523 & 0.3057 & -0.8001 & 14.0\% \\
& Improvement & \textcolor{red}{\textbf{+2.0410}} & \textcolor{red}{\textbf{+0.0349}} & \textcolor{red}{\textbf{-0.0546}} & \textcolor{red}{\textbf{-0.0272}} & \textcolor{red}{\textbf{-0.1225}} & \textcolor{red}{\textbf{+3.6585}} & -0.0335 & \textcolor{red}{\textbf{+0.0006}} & \textcolor{red}{\textbf{-0.0289}} & \textcolor{red}{\textbf{+12.0\%}} \\
\midrule
\multirow{3}{*}{X-Restormer~\cite{X-Restormer}}
& w/o & 19.2102 & 0.7643 & 0.3692 & 0.2677 & 6.2276 & 45.2024 & 0.5000 & 0.3212 & -0.8399 & 12.0\% \\
& w/ & 22.3598 & 0.8155 & 0.3076 & 0.2130 & 6.0306 & 50.3683 & 0.4653 & 0.3460 & -0.9022 & 36.0\% \\
& Improvement & \textcolor{red}{\textbf{+3.1496}} & \textcolor{red}{\textbf{+0.0512}} & \textcolor{red}{\textbf{-0.0616}} & \textcolor{red}{\textbf{-0.0547}} & \textcolor{red}{\textbf{-0.1970}} & \textcolor{red}{\textbf{+5.1659}} & -0.0347 & \textcolor{red}{\textbf{+0.0248}} & \textcolor{red}{\textbf{-0.0623}} & \textcolor{red}{\textbf{+24.0\%}} \\
\midrule
\multirow{3}{*}{PromptIR~\cite{PromptIR}}
& w/o & 19.1699 & 0.7600 & 0.3694 & 0.2791 & 6.4491 & 43.8916 & 0.5112 & 0.3183 & -0.8270 & 8.0\% \\
& w/ & 22.5432 & 0.8158 & 0.3067 & 0.2078 & 5.8669 & 50.4777 & 0.4730 & 0.3433 & -0.8577 & 32.0\% \\
& Improvement & \textcolor{red}{\textbf{+3.3733}} & \textcolor{red}{\textbf{+0.0558}} & \textcolor{red}{\textbf{-0.0627}} & \textcolor{red}{\textbf{-0.0713}} & \textcolor{red}{\textbf{-0.5822}} & \textcolor{red}{\textbf{+6.5861}} & -0.0382 & \textcolor{red}{\textbf{+0.0250}} & \textcolor{red}{\textbf{-0.0307}} & \textcolor{red}{\textbf{+24.0\%}} \\
\midrule
\multirow{3}{*}{MoCE-IR~\cite{MoCE-IR}}
& w/o & 19.0607 & 0.7629 & 0.3543 & 0.2614 & 6.0423 & 45.1328 & 0.5164 & 0.3298 & -0.8246 & 10.0\% \\
& w/ & 22.7287 & 0.8207 & 0.2977 & 0.2127 & 5.8153 & 52.9277 & 0.4725 & 0.3675 & -0.9010 & 28.0\% \\
& Improvement & \textcolor{red}{\textbf{+3.6680}} & \textcolor{red}{\textbf{+0.0578}} & \textcolor{red}{\textbf{-0.0566}} & \textcolor{red}{\textbf{-0.0487}} & \textcolor{red}{\textbf{-0.2270}} & \textcolor{red}{\textbf{+7.7949}} & -0.0439 & \textcolor{red}{\textbf{+0.0377}} & \textcolor{red}{\textbf{-0.0764}} & \textcolor{red}{\textbf{+18.0\%}} \\
\midrule
\multirow{3}{*}{DA-CLIP~\cite{DA-CLIP}}
& w/o & 19.5973 & 0.7662 & 0.3632 & 0.2531 & 5.8255 & 40.2291 & 0.5353 & 0.3068 & -0.7779 & 10.0\% \\
& w/ & 20.0687 & 0.7717 & 0.3447 & 0.2376 & 5.7389 & 42.8584 & 0.5325 & 0.3185 & -0.7929 & 26.0\% \\
& Improvement & \textcolor{red}{\textbf{+0.4714}} & \textcolor{red}{\textbf{+0.0055}} & \textcolor{red}{\textbf{-0.0185}} & \textcolor{red}{\textbf{-0.0155}} & \textcolor{red}{\textbf{-0.0866}} & \textcolor{red}{\textbf{+2.6293}} & -0.0028 & \textcolor{red}{\textbf{+0.0117}} & \textcolor{red}{\textbf{-0.0150}} & \textcolor{red}{\textbf{+16.0\%}} \\
\midrule
\multirow{4}{*}{FoundIR~\cite{li2025foundir}}
& \textbf{official} & 19.2997 & 0.7570 & 0.3828 & 0.2823 & 6.3486 & 41.7011 & 0.5269 & 0.3059 & -0.7892 & 2.0\% \\
& w/o & 19.0407 & 0.7572 & 0.3698 & 0.2707 & 6.2946 & 43.1479 & 0.5144 & 0.3219 & -0.8477 & 14.0\% \\
& w/ & 20.7740 & 0.7935 & 0.3397 & 0.2457 & 6.5425 & 48.5426 & 0.4895 & 0.3357 & -0.9192 & 52.0\% \\
& Improvement & \textcolor{red}{\textbf{+1.7333}} & \textcolor{red}{\textbf{+0.0363}} & \textcolor{red}{\textbf{-0.0301}} & \textcolor{red}{\textbf{-0.0250}} & +0.2479 & \textcolor{red}{\textbf{+5.3947}} & -0.0249 & \textcolor{red}{\textbf{+0.0138}} & \textcolor{red}{\textbf{-0.0715}} & \textcolor{red}{\textbf{+38.0\%}} \\
\midrule
\multirow{3}{*}{FLUX-Controlnet~\cite{FLUX}}
& w/o & 19.3726 & 0.7573 & 0.3505 & 0.2365 & 5.3696 & 44.7788 & 0.5401 & 0.3466 & -0.7069 & 10.0\% \\
& w/ & 20.7361 & 0.7612 & 0.2184 & 0.1284 & 4.0121 & 65.9364 & 0.6226 & 0.5561 & -1.0359 & 52.0\% \\
& Improvement & \textcolor{red}{\textbf{+1.3635}} & \textcolor{red}{\textbf{+0.0039}} & \textcolor{red}{\textbf{-0.1321}} & \textcolor{red}{\textbf{-0.1081}} & \textcolor{red}{\textbf{-1.3575}} & \textcolor{red}{\textbf{+21.1576}} & \textcolor{red}{\textbf{+0.0825}} & \textcolor{red}{\textbf{+0.2095}} & \textcolor{red}{\textbf{-0.3290}} & \textcolor{red}{\textbf{+42.0\%}} \\
\midrule
\multirow{4}{*}{Qwen-Image-Edit~\cite{Qwen-Image-Edit}}
& \textbf{official} & 20.5206 & 0.7724 & 0.2445 & 0.1171 & 4.2007 & 67.3384 & 0.6219 & 0.5375 & -1.0476 & 72.0\% \\
& w/o & 20.6342 & 0.7774 & 0.2920 & 0.1878 & 5.4593 & 49.6837 & 0.5796 & 0.3576 & -0.9016 & 42.0\% \\
& w/ & 21.8703 & 0.7803 & 0.1952 & 0.0997 & 4.6093 & 63.7163 & 0.6268 & 0.5074 & -1.0201 & 64.0\% \\
& Improvement & \textcolor{red}{\textbf{+1.2361}} & \textcolor{red}{\textbf{+0.0029}} & \textcolor{red}{\textbf{-0.0968}} & \textcolor{red}{\textbf{-0.0881}} & \textcolor{red}{\textbf{-0.8500}} & \textcolor{red}{\textbf{+14.0326}} & \textcolor{red}{\textbf{+0.0472}} & \textcolor{red}{\textbf{+0.1498}} & \textcolor{red}{\textbf{-0.1185}} & \textcolor{red}{\textbf{+22.0\%}} \\
\bottomrule
\end{tabular}}

\end{table*}

%snow
\begin{table*}[t]
\centering
\caption{Comparison of representative restoration models ``w/o'' and ``w/'' GGT-100K on the \textbf{GGT100K-Snow} test set. For some models whose official releases can also handle real-world degradations, we additionally report their \textbf{official results}. ``Improvement'' indicates the performance gain brought by GGT-100K. The positive improvements are highlighted in \textcolor{red}{\textbf{red}}.}
\label{tab:main_snow}
\vspace{1mm}
\resizebox{\textwidth}{!}{
\begin{tabular}{c|c|cccc|ccccc|c}
\toprule
\multirow{2}{*}{Model} & \multirow{2}{*}{GGT-100K} & \multicolumn{4}{c|}{Full-reference fidelity metrics} & \multicolumn{5}{c|}{No-reference perceptual metrics} & \multirow{2}{*}{VLM-R $\uparrow$} \\
\cmidrule(lr){3-6}\cmidrule(lr){7-11}
& & PSNR $\uparrow$ & SSIM $\uparrow$ & LPIPS $\downarrow$ & DISTS $\downarrow$ & NIQE $\downarrow$ & MUSIQ $\uparrow$ & MANIQA $\uparrow$ & TOPIQ $\uparrow$ & AFINE-NR $\downarrow$ &  \\
\midrule
\multirow{3}{*}{MPRNet~\cite{zamir2021multi}}
& w/o & 22.6929 & 0.6970 & 0.3544 & 0.2089 & 4.8778 & 50.7305 & 0.5353 & 0.4197 & -0.6816 & 12.0\% \\
& w/ & 24.7152 & 0.7738 & 0.3144 & 0.2064 & 5.1178 & 49.8011 & 0.4927 & 0.3609 & -0.7347 & 28.0\% \\
& Improvement & \textcolor{red}{\textbf{+2.0223}} & \textcolor{red}{\textbf{+0.0768}} & \textcolor{red}{\textbf{-0.0400}} & \textcolor{red}{\textbf{-0.0025}} & +0.2400 & -0.9294 & -0.0426 & -0.0588 & \textcolor{red}{\textbf{-0.0531}} & \textcolor{red}{\textbf{+16.0\%}} \\
\midrule
\multirow{3}{*}{NAFNet~\cite{chen2022simple}}
& w/o & 22.6229 & 0.6973 & 0.3633 & 0.2055 & 4.7855 & 48.0930 & 0.5307 & 0.4120 & -0.6628 & 22.0\% \\
& w/ & 26.4338 & 0.7983 & 0.3143 & 0.2047 & 5.5590 & 48.8828 & 0.4719 & 0.3371 & -0.7594 & 54.0\% \\
& Improvement & \textcolor{red}{\textbf{+3.8109}} & \textcolor{red}{\textbf{+0.1010}} & \textcolor{red}{\textbf{-0.0490}} & \textcolor{red}{\textbf{-0.0008}} & +0.7735 & \textcolor{red}{\textbf{+0.7898}} & -0.0588 & -0.0749 & \textcolor{red}{\textbf{-0.0966}} & \textcolor{red}{\textbf{+32.0\%}} \\
\midrule
\multirow{3}{*}{SwinIR~\cite{liang2021swinir}}
& w/o & 21.3170 & 0.6695 & 0.3902 & 0.2213 & 4.7865 & 48.8828 & 0.5174 & 0.4254 & -0.6632 & 0.0\% \\
& w/ & 24.9210 & 0.7728 & 0.3183 & 0.1984 & 4.8716 & 48.7087 & 0.4809 & 0.3681 & -0.7006 & 34.0\% \\
& Improvement & \textcolor{red}{\textbf{+3.6040}} & \textcolor{red}{\textbf{+0.1033}} & \textcolor{red}{\textbf{-0.0719}} & \textcolor{red}{\textbf{-0.0229}} & +0.0851 & -0.1741 & -0.0365 & -0.0573 & \textcolor{red}{\textbf{-0.0374}} & \textcolor{red}{\textbf{+34.0\%}} \\
\midrule
\multirow{3}{*}{X-Restormer~\cite{X-Restormer}}
& w/o & 22.7818 & 0.6951 & 0.3464 & 0.1983 & 4.6836 & 50.0319 & 0.5375 & 0.4172 & -0.6860 & 14.0\% \\
& w/ & 26.5405 & 0.7994 & 0.3113 & 0.2037 & 5.4127 & 49.0516 & 0.4727 & 0.3404 & -0.7812 & 60.0\% \\
& Improvement & \textcolor{red}{\textbf{+3.7587}} & \textcolor{red}{\textbf{+0.1043}} & \textcolor{red}{\textbf{-0.0351}} & +0.0054 & +0.7291 & -0.9803 & -0.0648 & -0.0768 & \textcolor{red}{\textbf{-0.0952}} & \textcolor{red}{\textbf{+46.0\%}} \\
\midrule
\multirow{3}{*}{PromptIR~\cite{PromptIR}}
& w/o & 23.3299 & 0.6993 & 0.3377 & 0.1966 & 4.7201 & 49.2773 & 0.5372 & 0.4202 & -0.6737 & 14.0\% \\
& w/ & 26.2458 & 0.7950 & 0.3081 & 0.2034 & 5.1493 & 48.5721 & 0.4764 & 0.3452 & -0.7533 & 58.0\% \\
& Improvement & \textcolor{red}{\textbf{+2.9159}} & \textcolor{red}{\textbf{+0.0957}} & \textcolor{red}{\textbf{-0.0296}} & +0.0068 & +0.4292 & -0.7052 & -0.0608 & -0.0750 & \textcolor{red}{\textbf{-0.0796}} & \textcolor{red}{\textbf{+44.0\%}} \\
\midrule
\multirow{3}{*}{MoCE-IR~\cite{MoCE-IR}}
& w/o & 23.4365 & 0.7088 & 0.3396 & 0.1967 & 4.7384 & 50.5416 & 0.5346 & 0.4191 & -0.6859 & 26.0\% \\
& w/ & 26.3553 & 0.8067 & 0.3104 & 0.2128 & 5.7355 & 51.8140 & 0.4747 & 0.3609 & -0.8247 & 56.0\% \\
& Improvement & \textcolor{red}{\textbf{+2.9188}} & \textcolor{red}{\textbf{+0.0979}} & \textcolor{red}{\textbf{-0.0292}} & +0.0161 & +0.9971 & \textcolor{red}{\textbf{+1.2724}} & -0.0599 & -0.0582 & \textcolor{red}{\textbf{-0.1388}} & \textcolor{red}{\textbf{+30.0\%}} \\
\midrule
\multirow{3}{*}{DA-CLIP~\cite{DA-CLIP}}
& w/o & 23.6500 & 0.6909 & 0.3455 & 0.1987 & 4.9824 & 44.8509 & 0.5336 & 0.3966 & -0.6463 & 28.0\% \\
& w/ & 24.9975 & 0.7416 & 0.2613 & 0.1661 & 4.9824 & 48.2149 & 0.5276 & 0.3715 & -0.7474 & 68.0\% \\
& Improvement & \textcolor{red}{\textbf{+1.3475}} & \textcolor{red}{\textbf{+0.0507}} & \textcolor{red}{\textbf{-0.0842}} & \textcolor{red}{\textbf{-0.0326}} & +0.0000 & \textcolor{red}{\textbf{+3.3640}} & -0.0060 & -0.0251 & \textcolor{red}{\textbf{-0.1011}} & \textcolor{red}{\textbf{+40.0\%}} \\
\midrule
\multirow{4}{*}{FoundIR~\cite{li2025foundir}}
& \textbf{official} & 22.8380 & 0.6975 & 0.3618 & 0.2091 & 4.5560 & 47.3037 & 0.5446 & 0.4130 & -0.6632 & 0.0\% \\
& w/o & 22.8919 & 0.6881 & 0.3702 & 0.2127 & 5.0751 & 48.4214 & 0.5400 & 0.4112 & -0.6722 & 8.0\% \\
& w/ & 25.0554 & 0.7901 & 0.3346 & 0.2129 & 6.1618 & 50.4086 & 0.4704 & 0.3593 & -0.8202 & 56.0\% \\
& Improvement & \textcolor{red}{\textbf{+2.1635}} & \textcolor{red}{\textbf{+0.1020}} & \textcolor{red}{\textbf{-0.0356}} & +0.0002 & +1.0867 & \textcolor{red}{\textbf{+1.9872}} & -0.0696 & -0.0519 & \textcolor{red}{\textbf{-0.1480}} & \textcolor{red}{\textbf{+48.0\%}} \\
\midrule
\multirow{3}{*}{FLUX-Controlnet~\cite{FLUX}}
& w/o & 20.5898 & 0.6079 & 0.4095 & 0.2227 & 4.6388 & 50.7200 & 0.5584 & 0.4556 & -0.6629 & 24.0\% \\
& w/ & 21.4757 & 0.6666 & 0.2797 & 0.1656 & 4.3965 & 62.1546 & 0.6034 & 0.5069 & -0.8939 & 38.0\% \\
& Improvement & \textcolor{red}{\textbf{+0.8859}} & \textcolor{red}{\textbf{+0.0587}} & \textcolor{red}{\textbf{-0.1298}} & \textcolor{red}{\textbf{-0.0571}} & \textcolor{red}{\textbf{-0.2423}} & \textcolor{red}{\textbf{+11.4346}} & \textcolor{red}{\textbf{+0.0450}} & \textcolor{red}{\textbf{+0.0513}} & \textcolor{red}{\textbf{-0.2310}} & \textcolor{red}{\textbf{+14.0\%}} \\
\midrule
\multirow{4}{*}{Qwen-Image-Edit~\cite{Qwen-Image-Edit}}
& \textbf{official} & 17.9621 & 0.6334 & 0.3941 & 0.2002 & 4.5264 & 68.2379 & 0.6160 & 0.5523 & -1.0614 & 26.0\% \\
& w/o & 25.5801 & 0.7753 & 0.2352 & 0.1427 & 5.4487 & 50.8085 & 0.5479 & 0.3747 & -0.8326 & 94.0\% \\
& w/ & 24.2822 & 0.7512 & 0.2196 & 0.1253 & 5.0208 & 61.3688 & 0.5929 & 0.4657 & -0.9823 & 88.0\% \\
& Improvement & -1.2979 & -0.0241 & \textcolor{red}{\textbf{-0.0156}} & \textcolor{red}{\textbf{-0.0174}} & \textcolor{red}{\textbf{-0.4279}} & \textcolor{red}{\textbf{+10.5603}} & \textcolor{red}{\textbf{+0.0450}} & \textcolor{red}{\textbf{+0.0910}} & \textcolor{red}{\textbf{-0.1497}} & -6.0\% \\
\bottomrule
\end{tabular}}

\end{table*}

%ll
\begin{table*}[t]
\centering
\caption{Comparison of representative restoration models ``w/o'' and ``w/'' GGT-100K on the \textbf{GGT100K-Low-Light} test set. For some models whose official releases can also handle real-world degradations, we additionally report their \textbf{official results}. ``Improvement'' indicates the performance gain brought by GGT-100K. The positive improvements are highlighted in \textcolor{red}{\textbf{red}}.}
\label{tab:main_lowlight}
\vspace{1mm}
\resizebox{\textwidth}{!}{
\begin{tabular}{c|c|cccc|ccccc|c}
\toprule
\multirow{2}{*}{Model} & \multirow{2}{*}{GGT-100K} & \multicolumn{4}{c|}{Full-reference fidelity metrics} & \multicolumn{5}{c|}{No-reference perceptual metrics} & \multirow{2}{*}{VLM-R $\uparrow$} \\
\cmidrule(lr){3-6}\cmidrule(lr){7-11}
& & PSNR $\uparrow$ & SSIM $\uparrow$ & LPIPS $\downarrow$ & DISTS $\downarrow$ & NIQE $\downarrow$ & MUSIQ $\uparrow$ & MANIQA $\uparrow$ & TOPIQ $\uparrow$ & AFINE-NR $\downarrow$ &  \\
\midrule
\multirow{3}{*}{MPRNet~\cite{zamir2021multi}}
& w/o & 18.5099 & 0.7166 & 0.4941 & 0.3778 & 8.6095 & 34.0214 & 0.4579 & 0.2767 & -0.8531 & 18.0\% \\
& w/ & 21.4928 & 0.7940 & 0.3998 & 0.2898 & 7.0979 & 41.2545 & 0.4774 & 0.3140 & -0.8213 & 50.0\% \\
& Improvement & \textcolor{red}{\textbf{+2.9829}} & \textcolor{red}{\textbf{+0.0774}} & \textcolor{red}{\textbf{-0.0943}} & \textcolor{red}{\textbf{-0.0880}} & \textcolor{red}{\textbf{-1.5116}} & \textcolor{red}{\textbf{+7.2331}} & \textcolor{red}{\textbf{+0.0195}} & \textcolor{red}{\textbf{+0.0373}} & +0.0318 & \textcolor{red}{\textbf{+32.0\%}} \\
\midrule
\multirow{3}{*}{NAFNet~\cite{chen2022simple}}
& w/o & 18.0173 & 0.7095 & 0.4846 & 0.3759 & 7.6623 & 35.7853 & 0.4648 & 0.3002 & -0.8277 & 26.0\% \\
& w/ & 22.7054 & 0.8123 & 0.3619 & 0.2547 & 6.2793 & 44.1940 & 0.4764 & 0.3236 & -0.8196 & 78.0\% \\
& Improvement & \textcolor{red}{\textbf{+4.6881}} & \textcolor{red}{\textbf{+0.1028}} & \textcolor{red}{\textbf{-0.1227}} & \textcolor{red}{\textbf{-0.1212}} & \textcolor{red}{\textbf{-1.3830}} & \textcolor{red}{\textbf{+8.4087}} & \textcolor{red}{\textbf{+0.0116}} & \textcolor{red}{\textbf{+0.0234}} & +0.0081 & \textcolor{red}{\textbf{+52.0\%}} \\
\midrule
\multirow{3}{*}{SwinIR~\cite{liang2021swinir}}
& w/o & 16.9328 & 0.6797 & 0.5361 & 0.3780 & 6.9303 & 30.8260 & 0.4192 & 0.2444 & -0.7589 & 6.0\% \\
& w/ & 19.8966 & 0.7693 & 0.4307 & 0.2994 & 6.0399 & 38.6161 & 0.4279 & 0.2874 & -0.7587 & 40.0\% \\
& Improvement & \textcolor{red}{\textbf{+2.9638}} & \textcolor{red}{\textbf{+0.0896}} & \textcolor{red}{\textbf{-0.1054}} & \textcolor{red}{\textbf{-0.0786}} & \textcolor{red}{\textbf{-0.8904}} & \textcolor{red}{\textbf{+7.7901}} & \textcolor{red}{\textbf{+0.0087}} & \textcolor{red}{\textbf{+0.0430}} & +0.0002 & \textcolor{red}{\textbf{+34.0\%}} \\
\midrule
\multirow{3}{*}{X-Restormer~\cite{X-Restormer}}
& w/o & 18.1090 & 0.7399 & 0.4638 & 0.3478 & 7.5393 & 38.4316 & 0.4529 & 0.3013 & -0.8473 & 42.0\% \\
& w/ & 22.6739 & 0.8192 & 0.3616 & 0.2543 & 7.0478 & 45.4584 & 0.4766 & 0.3153 & -0.8754 & 86.0\% \\
& Improvement & \textcolor{red}{\textbf{+4.5649}} & \textcolor{red}{\textbf{+0.0793}} & \textcolor{red}{\textbf{-0.1022}} & \textcolor{red}{\textbf{-0.0935}} & \textcolor{red}{\textbf{-0.4915}} & \textcolor{red}{\textbf{+7.0268}} & \textcolor{red}{\textbf{+0.0237}} & \textcolor{red}{\textbf{+0.0140}} & \textcolor{red}{\textbf{-0.0281}} & \textcolor{red}{\textbf{+44.0\%}} \\
\midrule
\multirow{3}{*}{PromptIR~\cite{PromptIR}}
& w/o & 18.7550 & 0.7231 & 0.4789 & 0.3698 & 7.8158 & 37.2343 & 0.4724 & 0.2991 & -0.8759 & 42.0\% \\
& w/ & 22.4263 & 0.8165 & 0.3541 & 0.2526 & 6.8982 & 45.4209 & 0.4805 & 0.3257 & -0.8846 & 76.0\% \\
& Improvement & \textcolor{red}{\textbf{+3.6713}} & \textcolor{red}{\textbf{+0.0934}} & \textcolor{red}{\textbf{-0.1248}} & \textcolor{red}{\textbf{-0.1172}} & \textcolor{red}{\textbf{-0.9176}} & \textcolor{red}{\textbf{+8.1866}} & \textcolor{red}{\textbf{+0.0081}} & \textcolor{red}{\textbf{+0.0266}} & \textcolor{red}{\textbf{-0.0087}} & \textcolor{red}{\textbf{+34.0\%}} \\
\midrule
\multirow{3}{*}{MoCE-IR~\cite{MoCE-IR}}
& w/o & 18.4825 & 0.7225 & 0.4680 & 0.3596 & 7.7570 & 36.5115 & 0.4721 & 0.2994 & -0.8655 & 34.0\% \\
& w/ & 22.4736 & 0.8177 & 0.3691 & 0.2592 & 7.2668 & 45.6688 & 0.4733 & 0.3228 & -0.8901 & 80.0\% \\
& Improvement & \textcolor{red}{\textbf{+3.9911}} & \textcolor{red}{\textbf{+0.0952}} & \textcolor{red}{\textbf{-0.0989}} & \textcolor{red}{\textbf{-0.1004}} & \textcolor{red}{\textbf{-0.4902}} & \textcolor{red}{\textbf{+9.1573}} & \textcolor{red}{\textbf{+0.0012}} & \textcolor{red}{\textbf{+0.0234}} & \textcolor{red}{\textbf{-0.0246}} & \textcolor{red}{\textbf{+46.0\%}} \\
\midrule
\multirow{3}{*}{DA-CLIP~\cite{DA-CLIP}}
& w/o & 16.8809 & 0.6615 & 0.5159 & 0.4141 & 8.7764 & 32.9001 & 0.4939 & 0.2897 & -0.8783 & 30.0\% \\
& w/ & 18.7191 & 0.7159 & 0.4115 & 0.3173 & 7.2236 & 37.3314 & 0.5036 & 0.3105 & -0.8745 & 56.0\% \\
& Improvement & \textcolor{red}{\textbf{+1.8382}} & \textcolor{red}{\textbf{+0.0544}} & \textcolor{red}{\textbf{-0.1044}} & \textcolor{red}{\textbf{-0.0968}} & \textcolor{red}{\textbf{-1.5528}} & \textcolor{red}{\textbf{+4.4313}} & \textcolor{red}{\textbf{+0.0097}} & \textcolor{red}{\textbf{+0.0208}} & +0.0038 & \textcolor{red}{\textbf{+26.0\%}} \\
\midrule
\multirow{4}{*}{FoundIR~\cite{li2025foundir}}
& \textbf{official} & 18.9683 & 0.7556 & 0.4202 & 0.3376 & 7.4485 & 37.1942 & 0.4885 & 0.3093 & -0.8596 & 58.0\% \\
& w/o & 21.3275 & 0.7811 & 0.4194 & 0.3258 & 8.2551 & 42.7873 & 0.4618 & 0.3236 & -0.9248 & 64.0\% \\
& w/ & 19.8822 & 0.7638 & 0.4104 & 0.3174 & 9.0116 & 41.5394 & 0.4805 & 0.3177 & -0.9438 & 84.0\% \\
& Improvement & -1.4453 & -0.0173 & \textcolor{red}{\textbf{-0.0090}} & \textcolor{red}{\textbf{-0.0084}} & +0.7565 & -1.2479 & \textcolor{red}{\textbf{+0.0187}} & -0.0059 & \textcolor{red}{\textbf{-0.0190}} & \textcolor{red}{\textbf{+20.0\%}} \\
\midrule
\multirow{3}{*}{FLUX-Controlnet~\cite{FLUX}}
& w/o & 19.9199 & 0.7312 & 0.4508 & 0.2572 & 6.5845 & 50.1438 & 0.5236 & 0.4068 & -0.8537 & 46.0\% \\
& w/ & 19.7996 & 0.7279 & 0.3383 & 0.1962 & 6.0633 & 57.7316 & 0.5753 & 0.4686 & -0.9991 & 54.0\% \\
& Improvement & -0.1203 & -0.0033 & \textcolor{red}{\textbf{-0.1125}} & \textcolor{red}{\textbf{-0.0610}} & \textcolor{red}{\textbf{-0.5212}} & \textcolor{red}{\textbf{+7.5878}} & \textcolor{red}{\textbf{+0.0517}} & \textcolor{red}{\textbf{+0.0618}} & \textcolor{red}{\textbf{-0.1454}} & \textcolor{red}{\textbf{+8.0\%}} \\
\midrule
\multirow{4}{*}{Qwen-Image-Edit~\cite{Qwen-Image-Edit}}
& \textbf{official} & 17.1073 & 0.6953 & 0.4166 & 0.2748 & 7.1067 & 50.8637 & 0.5216 & 0.3968 & -0.9757 & 66.0\% \\
& w/o & 21.1225 & 0.7926 & 0.3344 & 0.2034 & 7.0565 & 48.3712 & 0.5091 & 0.3455 & -0.9300 & 88.0\% \\
& w/ & 23.0697 & 0.7907 & 0.2502 & 0.1458 & 6.6361 & 60.6649 & 0.5730 & 0.4580 & -1.0349 & 94.0\% \\
& Improvement & \textcolor{red}{\textbf{+1.9472}} & -0.0019 & \textcolor{red}{\textbf{-0.0842}} & \textcolor{red}{\textbf{-0.0576}} & \textcolor{red}{\textbf{-0.4204}} & \textcolor{red}{\textbf{+12.2937}} & \textcolor{red}{\textbf{+0.0639}} & \textcolor{red}{\textbf{+0.1125}} & \textcolor{red}{\textbf{-0.1049}} & \textcolor{red}{\textbf{+6.0\%}} \\
\bottomrule
\end{tabular}}

\end{table*}

%old
\begin{table*}[t]
\centering
\caption{Comparison of representative restoration models ``w/o'' and ``w/'' GGT-100K on the \textbf{GGT100K-Old Photo} test set. For some models whose official releases can also handle real-world degradations, we additionally report their \textbf{official results}. ``Improvement'' indicates the performance gain brought by GGT-100K. The positive improvements are highlighted in \textcolor{red}{\textbf{red}}.}
\label{tab:main_oldphoto}
\vspace{1mm}
\resizebox{\textwidth}{!}{
\begin{tabular}{c|c|cccc|ccccc|c}
\toprule
\multirow{2}{*}{Model} & \multirow{2}{*}{GGT-100K} & \multicolumn{4}{c|}{Full-reference fidelity metrics} & \multicolumn{5}{c|}{No-reference perceptual metrics} & \multirow{2}{*}{VLM-R $\uparrow$} \\
\cmidrule(lr){3-6}\cmidrule(lr){7-11}
& & PSNR $\uparrow$ & SSIM $\uparrow$ & LPIPS $\downarrow$ & DISTS $\downarrow$ & NIQE $\downarrow$ & MUSIQ $\uparrow$ & MANIQA $\uparrow$ & TOPIQ $\uparrow$ & AFINE-NR $\downarrow$ &  \\
\midrule
\multirow{3}{*}{MPRNet~\cite{zamir2021multi}}
& w/o & 26.2769 & 0.8285 & 0.3231 & 0.1835 & 6.1129 & 44.5421 & 0.4939 & 0.3122 & -0.6687 & 30.0\% \\
& w/ & 29.2252 & 0.8686 & 0.2830 & 0.1836 & 6.0897 & 48.5279 & 0.4784 & 0.3308 & -0.7995 & 50.0\% \\
& Improvement & \textcolor{red}{\textbf{+2.9483}} & \textcolor{red}{\textbf{+0.0401}} & \textcolor{red}{\textbf{-0.0401}} & +0.0001 & \textcolor{red}{\textbf{-0.0232}} & \textcolor{red}{\textbf{+3.9858}} & -0.0155 & \textcolor{red}{\textbf{+0.0186}} & \textcolor{red}{\textbf{-0.1308}} & \textcolor{red}{\textbf{+20.0\%}} \\
\midrule
\multirow{3}{*}{NAFNet~\cite{chen2022simple}}
& w/o & 26.8340 & 0.8431 & 0.3050 & 0.1723 & 5.9239 & 43.2106 & 0.4933 & 0.3029 & -0.6528 & 48.0\% \\
& w/ & 29.5942 & 0.8758 & 0.2637 & 0.1744 & 6.0141 & 49.2069 & 0.4633 & 0.3277 & -0.8143 & 56.0\% \\
& Improvement & \textcolor{red}{\textbf{+2.7602}} & \textcolor{red}{\textbf{+0.0327}} & \textcolor{red}{\textbf{-0.0413}} & +0.0021 & +0.0902 & \textcolor{red}{\textbf{+5.9963}} & -0.0300 & \textcolor{red}{\textbf{+0.0248}} & \textcolor{red}{\textbf{-0.1615}} & \textcolor{red}{\textbf{+8.0\%}} \\
\midrule
\multirow{3}{*}{SwinIR~\cite{liang2021swinir}}
& w/o & 26.2180 & 0.8363 & 0.3265 & 0.1774 & 6.1039 & 45.1656 & 0.4864 & 0.3116 & -0.6682 & 36.0\% \\
& w/ & 28.8712 & 0.8666 & 0.2873 & 0.1746 & 6.2721 & 45.7231 & 0.4632 & 0.3002 & -0.7664 & 50.0\% \\
& Improvement & \textcolor{red}{\textbf{+2.6532}} & \textcolor{red}{\textbf{+0.0303}} & \textcolor{red}{\textbf{-0.0392}} & \textcolor{red}{\textbf{-0.0028}} & +0.1682 & \textcolor{red}{\textbf{+0.5575}} & -0.0232 & -0.0114 & \textcolor{red}{\textbf{-0.0982}} & \textcolor{red}{\textbf{+14.0\%}} \\
\midrule
\multirow{3}{*}{X-Restormer~\cite{X-Restormer}}
& w/o & 27.1331 & 0.8506 & 0.3032 & 0.1700 & 6.0296 & 44.4367 & 0.5019 & 0.3143 & -0.6845 & 48.0\% \\
& w/ & 29.8823 & 0.8769 & 0.2665 & 0.1727 & 6.0065 & 49.0715 & 0.4639 & 0.3302 & -0.8300 & 56.0\% \\
& Improvement & \textcolor{red}{\textbf{+2.7492}} & \textcolor{red}{\textbf{+0.0263}} & \textcolor{red}{\textbf{-0.0367}} & +0.0027 & \textcolor{red}{\textbf{-0.0231}} & \textcolor{red}{\textbf{+4.6348}} & -0.0380 & \textcolor{red}{\textbf{+0.0159}} & \textcolor{red}{\textbf{-0.1455}} & \textcolor{red}{\textbf{+8.0\%}} \\
\midrule
\multirow{3}{*}{PromptIR~\cite{PromptIR}}
& w/o & 26.3384 & 0.8189 & 0.2998 & 0.1782 & 5.8692 & 47.2604 & 0.5098 & 0.3339 & -0.6678 & 36.0\% \\
& w/ & 29.8668 & 0.8780 & 0.2644 & 0.1759 & 6.0530 & 49.1353 & 0.4671 & 0.3298 & -0.8159 & 56.0\% \\
& Improvement & \textcolor{red}{\textbf{+3.5284}} & \textcolor{red}{\textbf{+0.0591}} & \textcolor{red}{\textbf{-0.0354}} & \textcolor{red}{\textbf{-0.0023}} & +0.1838 & \textcolor{red}{\textbf{+1.8749}} & -0.0427 & -0.0041 & \textcolor{red}{\textbf{-0.1481}} & \textcolor{red}{\textbf{+20.0\%}} \\
\midrule
\multirow{3}{*}{MoCE-IR~\cite{MoCE-IR}}
& w/o & 26.9140 & 0.8378 & 0.3017 & 0.1724 & 5.9494 & 44.7693 & 0.5014 & 0.3189 & -0.6697 & 46.0\% \\
& w/ & 29.9449 & 0.8804 & 0.2610 & 0.1792 & 5.9774 & 50.6919 & 0.4667 & 0.3472 & -0.8520 & 60.0\% \\
& Improvement & \textcolor{red}{\textbf{+3.0309}} & \textcolor{red}{\textbf{+0.0426}} & \textcolor{red}{\textbf{-0.0407}} & +0.0068 & +0.0280 & \textcolor{red}{\textbf{+5.9226}} & -0.0347 & \textcolor{red}{\textbf{+0.0283}} & \textcolor{red}{\textbf{-0.1823}} & \textcolor{red}{\textbf{+14.0\%}} \\
\midrule
\multirow{3}{*}{DA-CLIP~\cite{DA-CLIP}}
& w/o & 27.9405 & 0.8404 & 0.2929 & 0.1623 & 5.7807 & 39.3277 & 0.5031 & 0.2848 & -0.6372 & 24.0\% \\
& w/ & 28.5906 & 0.8473 & 0.2342 & 0.1419 & 5.6770 & 43.1152 & 0.4972 & 0.3045 & -0.7158 & 56.0\% \\
& Improvement & \textcolor{red}{\textbf{+0.6501}} & \textcolor{red}{\textbf{+0.0069}} & \textcolor{red}{\textbf{-0.0587}} & \textcolor{red}{\textbf{-0.0204}} & \textcolor{red}{\textbf{-0.1037}} & \textcolor{red}{\textbf{+3.7875}} & -0.0059 & \textcolor{red}{\textbf{+0.0197}} & \textcolor{red}{\textbf{-0.0786}} & \textcolor{red}{\textbf{+32.0\%}} \\
\midrule
\multirow{4}{*}{FoundIR~\cite{li2025foundir}}
& \textbf{official} & 27.7643 & 0.8439 & 0.3062 & 0.1696 & 6.0045 & 39.8216 & 0.5091 & 0.2937 & -0.6412 & 26.0\% \\
& w/o & 28.2156 & 0.8505 & 0.2944 & 0.1661 & 6.0254 & 42.8725 & 0.5014 & 0.3050 & -0.6901 & 54.0\% \\
& w/ & 29.5300 & 0.8769 & 0.2752 & 0.1727 & 6.3344 & 44.5586 & 0.4687 & 0.2985 & -0.7785 & 68.0\% \\
& Improvement & \textcolor{red}{\textbf{+1.3144}} & \textcolor{red}{\textbf{+0.0264}} & \textcolor{red}{\textbf{-0.0192}} & +0.0066 & +0.3090 & \textcolor{red}{\textbf{+1.6861}} & -0.0327 & -0.0065 & \textcolor{red}{\textbf{-0.0884}} & \textcolor{red}{\textbf{+14.0\%}} \\
\midrule
\multirow{3}{*}{FLUX-Controlnet~\cite{FLUX}}
& w/o & 25.4824 & 0.7785 & 0.3341 & 0.1789 & 5.5024 & 43.5297 & 0.5163 & 0.3259 & -0.5865 & 28.0\% \\
& w/ & 24.1773 & 0.7715 & 0.2292 & 0.1332 & 3.9432 & 62.3481 & 0.6063 & 0.5151 & -0.9245 & 68.0\% \\
& Improvement & -1.3051 & -0.0070 & \textcolor{red}{\textbf{-0.1049}} & \textcolor{red}{\textbf{-0.0457}} & \textcolor{red}{\textbf{-1.5592}} & \textcolor{red}{\textbf{+18.8184}} & \textcolor{red}{\textbf{+0.0900}} & \textcolor{red}{\textbf{+0.1892}} & \textcolor{red}{\textbf{-0.3380}} & \textcolor{red}{\textbf{+40.0\%}} \\
\midrule
\multirow{4}{*}{Qwen-Image-Edit~\cite{Qwen-Image-Edit}}
& \textbf{official} & 21.6575 & 0.7757 & 0.3228 & 0.1788 & 4.3479 & 65.5946 & 0.5940 & 0.5241 & -0.9868 & 74.0\% \\
& w/o & 26.1690 & 0.8183 & 0.2410 & 0.1281 & 4.9971 & 56.0239 & 0.5749 & 0.4267 & -0.8615 & 74.0\% \\
& w/ & 27.1337 & 0.8298 & 0.2094 & 0.1143 & 4.4643 & 62.7812 & 0.5963 & 0.4877 & -0.9779 & 76.0\% \\
& Improvement & \textcolor{red}{\textbf{+0.9647}} & \textcolor{red}{\textbf{+0.0115}} & \textcolor{red}{\textbf{-0.0316}} & \textcolor{red}{\textbf{-0.0138}} & \textcolor{red}{\textbf{-0.5328}} & \textcolor{red}{\textbf{+6.7573}} & \textcolor{red}{\textbf{+0.0214}} & \textcolor{red}{\textbf{+0.0610}} & \textcolor{red}{\textbf{-0.1164}} & \textcolor{red}{\textbf{+2.0\%}} \\
\bottomrule
\end{tabular}}

\end{table*}

\end{document}